\pdfoutput=1
\documentclass[11pt]{article}

\usepackage{acl}

\usepackage{times}
\usepackage{latexsym}
\usepackage{multicol}

\usepackage[T1]{fontenc}
\usepackage[utf8]{inputenc}

\usepackage{microtype}

\usepackage{hyperref}
\usepackage{url}
\usepackage{graphicx}
\usepackage{multirow}
\usepackage{breqn}

\usepackage{caption}
\usepackage{subcaption}

\usepackage{amsfonts}

\usepackage[textwidth=90]{todonotes}

\usepackage{balance}

\title{Your fairness may vary:\\Pretrained language model fairness in toxic text classification}

\author{Ioana Baldini \qquad Dennis Wei \qquad Karthikeyan Natesan Ramamurthy\\
\textbf{Mikhail Yurochkin \qquad Moninder Singh}\\
    IBM Research\\
  \texttt{\{ioana,dwei,knatesa,moninder\}@us.ibm.com}\\ \texttt{mikhail.yurochkin@ibm.com}
}

\begin{document}
\maketitle
\begin{abstract}
The popularity of pretrained language models in natural language processing systems calls for a careful evaluation of such models in down-stream tasks, which have a higher potential for societal impact. The evaluation of such systems usually focuses on \textit{accuracy measures}. Our findings in this paper call for attention to be paid to \textit{fairness measures} as well. Through the analysis of more than a dozen pretrained language models of varying sizes on two toxic text classification tasks (English), we demonstrate that focusing on accuracy measures alone
can lead to models with wide variation in fairness characteristics. Specifically, we observe that fairness can vary even more than accuracy with increasing training data size and different random initializations. At the same time, we find that little of the fairness variation is explained by model size, despite claims in the literature. To improve model fairness without retraining, we show that two post-processing methods developed for structured, tabular data can be successfully applied to a range of pretrained language models. \textit{Warning: This paper contains samples of offensive text.}\\
\end{abstract}

\section{Introduction}

% importance of pre-trained LMs
Pre-trained, bidirectional language models~\citep{Devlin2019BERT,liu2019roberta, Radford2019Language, Clark2020Electra,He2021Deberta}\footnote{We use the acronym LM(s) to refer to language model(s) throughout the paper.} have revolutionized natural language processing (NLP) research.
% , trained from language data.
LMs have provided a route to significant performance increases in several NLP tasks as demonstrated by NLP leaderboards~\citep{Rajpurkar2018Know,wang2019superglue,wang2019glue,AI2Leaderboards}. More importantly, LMs have been applied to practical problems, leading to improved results for web search~\citep{Nayak2019Understanding} and have become an asset in fields such as medical evidence inference~\citep{lehman2019inferring,Subramanian2020Natural} and chemistry~\citep{Schwaller2021Mapping}. 
While the progress in NLP tasks due to LMs is clear, the reasons behind this success are not as well understood~\citep{Rogers2021Primer,mccoy2019right}, and there are also important downsides. In particular, several studies have documented the \textit{bias} of LMs~\citep{Bolukbasi2016Man,hutchinson2020social,Webster2020Measuring,Borkan2019Nuanced,DeVassimon2021Stereotype} and others discuss potential societal harms~\citep{Blodgett2020Language,Bender2021Dangers} for individuals or groups. We use the term \emph{bias} to refer to systematic disparity in representation or outcomes for individuals based on their membership in certain protected groups such as gender, race, and ethnicity. 

In this work, we focus on one important application of fine-tuned LMs, toxic text classification. Text toxicity predictors are already used in deployed systems~\citep{Perspective2021} and they are a crucial component for content moderation since online harassment is on the rise~\citep{vogels2021state}. 
In downstream applications such as toxic text classification, it is important to examine the behavior of LMs in terms of measures other than task-specific accuracy. This provides a more holistic understanding of model performance and %, leading to improved insights into 
appropriate uses of LMs for these tasks. As a first step toward this goal, we provide herein an empirical characterization of LMs for the task of toxic text classification using a combination of accuracy and bias measures, and study two post-processing methods for bias mitigation that have proved successful for structured, tabular data. For assessing bias, in this paper, we focus on \textit{group fairness}, which we explain in Section~\ref{sec:background} as it applies in general in machine learning, and discuss what it means in the context of NLP tasks in the same section. The implications of measuring group fairness for the toxicity classification task studied in this paper are described in Section~\ref{sec:methodology}.

One aspect of LMs that is hard to ignore is the increase in their size, as measured by the number of parameters in their architectures. In general, larger LMs seem to perform better on NLP tasks as they have the capacity to capture more complex correlations present in the training data. \citet{Bender2021Dangers} claim that this same property may also lead to more pronounced biases in their predictions, as the large data that LMs are trained on is not curated. On the other hand, for \emph{image} classification models that use large neural networks,~\citet{Hooker2020Characterising} discuss how model pruning can lead to more biased predictions. In this work, we consider a wide variety of model architectures and sizes. We acknowledge that size is relative and what we consider large in this paper may not be considered as such in a different context.

We address the following questions regarding the effect of various factors on model performance:
\begin{enumerate}
% reduce spacing between items
  \setlength{\itemsep}{1pt}
  \setlength{\parskip}{0pt}
  \setlength{\parsep}{0pt}
    \item \textit{Model size}:
    How do the accuracy and group fairness of fine-tuned LM-based classifiers vary with their size? 
    \item \textit{Random seeds}: LMs that start from different random initializations can behave differently in classification. What is the effect of random seeds on the accuracy-fairness relationship?
    \item \textit{Data size}: The size of fine-tuning data is also an important dimension alongside model size. What happens to accuracy and fairness when more/less data is used for fine-tuning? 
    \item \textit{Bias mitigation via post-processing}: Given the expense of training and fine-tuning large LMs, to what extent can we mitigate %extrinsic 
    bias by only post-processing LM outputs?
\end{enumerate}

% importance of empirical studies of fairness
We study the accuracy-fairness relationship in more than a dozen fine-tuned LMs for two different datasets that deal with prediction of text toxicity. The key contributions of our analysis are:
\begin{enumerate}
% reduce spacing between items
  \setlength{\itemsep}{1pt}
  \setlength{\parskip}{0pt}
  \setlength{\parsep}{0pt}
\item We empirically show that no blanket statement can be made regarding the fairness characteristics of fine-tuned LMs with respect to their size. It really depends on the combination of LM, task, and dataset.
\item We find that optimizing for accuracy measures alone can lead to models with wide variation in fairness characteristics. Specifically:
\begin{enumerate}
    \item While increasing data size for fine-tuning does not improve accuracy much beyond a point, the improvement in fairness is more significant and may continue after the improvement in accuracy has stopped for certain datasets and tasks. This suggests that choosing data sizes based on accuracy alone could lead to suboptimal performance with respect to fairness.
    \item While accuracy measures are known to vary with different random initializations~\citep{dodge2020Finetuning}, the variation in fairness measures can be even greater.
\end{enumerate}
\item We demonstrate that post-processing bias mitigation is an effective, computationally affordable solution to enhance fairness in fine-tuned LMs. In particular, one of the methods we experimented with allows for a large accuracy-fairness tradeoff space, leading to relative improvements of 50\% for fairness, as measured by equalized odds, while reducing accuracy only by 2\% (see Figure~\ref{fig:jigsaw_bias_mitigation} religion group).
\end{enumerate}

Our observations strengthen the chorus of recent work addressing bias mitigation in NLP in calling for a careful empirical analysis of fairness with fine-tuned LMs in the context of their application. To allow group fairness analysis, annotations of group membership are preferred and sometimes required, and, thus, we urge the research community to include protected group annotations in datasets to enable extrinsic fairness evaluations that are as close as possible to the point of deployment. 

%\section{Background, terminology and related work}
\section{Background and related work}
\label{sec:background}
\subsection{Fairness in machine learning}
As machine learning models have become routinely deployed in practice, many studies noticed their tendency to perform unfairly in various contexts \citep{angwin2016Machine,angwin2017Minority,buolamwini2018gender,park2021comparison}. To understand and measure model bias, researchers have proposed many definitions of algorithmic fairness. Broadly speaking, they fall into two categories: \emph{group fairness}~\citep{chouldechova2018frontiers} and~\emph{individual fairness}~\citep{dwork2012fairness}. At a high level, group fairness requires similar average outcomes on different groups of individuals considered, for example comparable university acceptance rates across ethnicities. Individual fairness requires similar outcomes for similar individuals, e.g. two university applicants with similar credentials, but different ethnicity, gender, family background, etc., should either be both accepted or both rejected. In this paper we consider group fairness, noting that both have their pros and cons~\citep{chouldechova2018frontiers,dwork2012fairness}.

There are many definitions of group fairness and we refer to \citet{Verma2018Fairness} for a comprehensive overview and to ~\citet{czarnowska2021quantifying} for a discussion of metrics in the context of measuring social biases in NLP. Statistical parity (SP) is one of the earlier definitions which requires the output of a model to be independent of the \emph{sensitive attribute}, such as race or gender. In other words, the average outcome (e.g. prediction) across groups defined by the sensitive attribute needs to be similar. 
An alternative measure is equalized odds (EO)~\citep{hardt2016equality}, which requires the model output \emph{conditioned} on the true label to be independent of the sensitive attribute. 
The violation of conditional independence for a given label (positive or negative) can be measured by the difference in accuracy across sensitive groups conditioned on that label. Taking the maximum or an average (average EO) of these label-specific differences quantifies the overall EO violation.

Many methods for achieving group fairness have been proposed. These methods are typically categorized as follows: (a) modifying the training data (pre-processing), (b) incorporating fairness constraints while training the model (in-processing), and (c) transforming the model output to enhance fairness (post-processing). A summary and implementation of group bias mitigation approaches are discussed in~\citet{bellamy2019ai}. In this study, we investigate the use of post-processing methods to enhance fairness in classification tasks. We chose post-processing approaches since they do not require modification of training data or model training procedures, and, hence, can be efficiently applied to all LMs we consider.
In addition, post-processing approaches could minimize the environmental impact of re-training/fine-tuning LMs~\citep{patterson2021carbon, Strubell2019Energy}. We consider two post-processing approaches proposed by~\citet{Wei2020Optimized} and~\citet{hardt2016equality}, which have shown considerable success in mitigating bias for tabular data. \citet{Wei2020Optimized} optimize a score (predicted probability) transformation function to satisfy fairness constraints that are linear in conditional means of scores while minimizing a cross-entropy objective. \citet{hardt2016equality} propose to solve a linear program to find probabilities with which to change the predicted output labels such that the equalized odds violation is minimized. 

\subsection{Fairness in Natural Language Processing}
In NLP systems, bias is broadly understood in two categories, intrinsic and extrinsic. Intrinsic bias refers to bias inherent in the representations, e.g. word embeddings used in NLP~\citep{Bolukbasi2016Man}.
Extrinsic bias refers to bias in downstream tasks, such as disparity in false positive rates across groups defined by sensitive attributes in a specified application/task. The concepts of intrinsic and extrinsic bias also correlate well with the notions of \textit{representational} and \textit{allocative} harms. While allocative harms arise from disparities across different groups in terms of decisions that lead to allocation of benefits/harms, representational harms are those perpetuated by representation of individuals in the feature space~\citep{Crawford2013}. \citet{abbasi2019fairness} discuss how harms from stereotypical representations manifest as allocative harms later in the ML pipeline. However, probably because of the complexity of LMs, measuring intrinsic bias in the representations created by LMs may not necessarily reflect the behavior of models built by fine-tuning LMs. \citet{GoldfarbTarrant2021Intrinsic} discuss how intrinsic measures of bias do not correlate with extrinsic, application-specific, bias measures. Since we are concerned with the application of LMs to the specific task of toxic text classification, we restrict our focus to group fairness measures, which fall under the category of extrinsic bias.
Previous work on bias mitigation in NLP has been focused on pre- and in-processing methods~\citep{sun2019mitigating, ball2021differential} and to the best of our knowledge, we are the first to use post-processing methods with NLP tasks.

\section{Methodology}
\label{sec:methodology}
We are interested in studying how group fairness varies across different fine-tuned LMs for binary classification. We choose to focus on text toxicity as the prediction task. Due to an increase in online harassment~\citep{vogels2021state} and the potential of both propagating harmful stereotypes of minority groups and/or inadvertently reducing their voices, the task of predicting toxicity in text has received increased attention in recent years~\citep{kiritchenko2021confronting}. While we acknowledge that text toxicity presents different complex nuances (e.g., offensive text, harassment, hate speech), we focus on a binary task formulation. We adopt the definition of toxicity described in~\citet{Borkan2019Nuanced} as \textit{``anything that is rude, disrespectful, or unreasonable that would make someone want to leave a conversation''}.
%quote from paper: "rude, disrespectful, or unreasonable that would make someone want to leave a conversation"

\subsection{Datasets} 
We used two datasets that deal with toxic text classification: 1) Jigsaw, a large dataset released for the ``Unintended Bias in Toxicity Classification'' Kaggle competition~\citep{kaggle2019jigsaw} that contains online comments on news articles, and 2) HateXplain, a dataset recently introduced with the intent of studying explanations for offensive and hate speech in Twitter and Twitter-like data (i.e., \url{gab.com}). Both datasets have fine-grained annotations for religion, race and gender. We used as sensitive groups the coarse-grained groups (e.g., mention of any religion, see Section~\ref{sec:methodology:fairness}) as opposed to the finer-grained annotations (e.g., Muslim). Details about the sizes of the datasets, the splits we used and text samples can be found in Appendix~\ref{appendix:datasets}.

\begin{table*}[t]
\small
\caption{The size (number of parameters, size on disk) for the language models considered in this study.}
\label{table:lms}
\begin{center}
\begin{tabular}{l|lcc}
\hline
Size Group & Language Model & \# of parameters & Size on disk \\
\hline
\multirow{4}{*}{Small} & ALBERT ~\citep{Lan2020Albert} & 12M & 45MB\\
& MobileBERT~\citep{Sun2020MobileBERT} & 25.3M & 95MB\\
& SqueezeBERT~\citep{iandola2020squeezebert} & 51M$^*$ & 196MB\\
& DistilBERT~\citep{sanh2020distilbert} & 66M & 256MB\\
\hline
\multirow{7}{*}{Regular} & BERT~\citep{Devlin2019BERT} & 110M & 418MB\\
% & BERTweet~\citep{Nguyen2020BERTweet} & 110M & 418MB\\
& ELECTRA~\citep{Clark2020Electra} & 110M & 418MB\\
& Funnel (small)~\citep{Dai2020Funnel} & 117M$^*$ & 444MB\\
& RoBERTa~\citep{liu2019roberta} & 125M & 476MB\\
& GPT2~\citep{Radford2019Language} & 117M & 487MB\\
& DeBERTa~\citep{He2021Deberta} & 140M & 532MB\\
\hline
\multirow{4}{*}{Large} & ELECTRA-large & 335M & 1.3GB\\
& BERT-large & 340M & 1.3GB\\
& RoBERTa-large & 355M & 1.4GB\\
& DeBERTa-large & 400M & 1.6GB\\
\hline
\multicolumn{4}{l}{$^*$Approximate number of parameters.}
\\
\end{tabular}
\end{center}
\end{table*}

\subsection{Language models, fine-tuning and computation infrastructure}
\label{section:lms}
We consider more than a dozen LMs that cover a large spectrum of sizes. We selected the models to not only represent various sizes but also different styles of architecture and training. The models in our study are shown in Table~\ref{table:lms} along with the number of parameters and the size of the \textit{PyTorch}~\citep{NEURIPS2019_9015} model on disk. If not specified, the version of the model used is \textit{base}. For all our experiments, we used the Hugging Face implementation of Transformers~\citep{wolf2020transformers} and the corresponding implementations for all LMs in our study. In particular, we use the \textit{text sequence classifier} without any modifications to increase reproducibility.

We run model fine-tuning for 1-3 epochs and choose the best model based on the highest accuracy obtained on the dev split. When presenting experimental results, we focus primarily on balanced accuracy as the Jigsaw dataset is highly imbalanced and reporting only accuracy may be misleading. In general, higher accuracy leads to higher balanced accuracy, with the exception of two LMs -- GPT2 and SqueezeBERT. For these two, the best balanced accuracy is less than 2 percentage points higher than the balanced accuracy resulting from choosing the highest overall accuracy across the various hyper-parameter runs. We experiment with two learning rates ($2e-6$ and $2e-5$) and observe that the large models tend to prefer smaller learning rate, degenerating for higher learning rates. For large LMs with Jigsaw we fine-tune for one epoch to keep the compute time under 24 hours. The model accuracy we obtained are in line with state-of-the-art results for these types of tasks. The large LMs are fine-tuned on A100 Nvidia GPUs, while the rest of the models are fine-tuned on V100 Nvidia GPUs. The experiments for HateXplain run from 10 minutes to under an hour, while the experiments for the large models with Jigsaw can take up to 24 hours.

\subsection{Sensitive groups and fairness measures}
\label{sec:methodology:fairness}
In all our measurements, we considered the following topics as sensitive: religion, race and gender. We categorize a text sample as belonging to a \emph{sensitive group} if it mentions one of these topics (e.g., religion), and otherwise to the complementary group (no religion). Except in Section~\ref{sec:subgroups}, we do not analyze finer-grained subgroups (e.g., Jewish), but consider larger groups (any reference to religion, such as Muslim, Jewish, atheist). There are several reasons that justify this choice. First, unlike tabular data where each sample corresponds to an individual belonging to one identity (e.g., either female or male), we do not have information on the demographics of the person producing the text. Our categorization is based on the \emph{content}. In addition, for the datasets we used, most subgroups account for significantly less than 1\% of the data. Moreover, there is considerable overlap between subgroups. For example, in the test split for Jigsaw, 40\% of the text belonging to the male subgroup also belongs to the female subgroup. To summarize, we analyze the bias/fairness of toxic text prediction in the presence or absence of information that refers to religion, race or gender, respectively. The intent is to not have the performance of the predictor be influenced by these sensitive topics.

We use equalized odds as the group fairness measure. Equalized odds is defined as the maximum of the absolute true positive rate difference and false positive rate difference, where these differences are between a sensitive group and its complementary group. In toxic text classification, a true positive means that a toxic text is correctly identified as such, while a false positive means that a benign piece of text is marked as toxic. In terms of harms, a false negative (toxic text that is missed) may cause individuals to feel threatened or disrespected, while a false positive may be seen as censoring, which is particularly problematic if it reduces the voices of minority protected groups from online conversations. By using the sensitive groups of religion/race/gender mentioned above, we aim to analyze and reduce the effect of the presence or absence of religion/race/gender terms on the false negative and false positive rates. By taking the maximum, we are emphasizing the larger discrepancy as opposed to other studies that take the average of the two rate differences (average equalized odds). Note that unlike statistical parity, equalized odds does allow the sensitive (e.g., mention of religion) and complementary (no religion) groups to have different toxicity (positive prediction) rates.

%\section{Post-processing methods for bias mitigation}
\section{Bias mitigation post-processing}
\label{sec:method:post}

We investigated the use of post-processing methods to mitigate violations of equalized odds. By post-processing, we mean methods that operate only on the outputs of the fine-tuned LMs and do not modify the models themselves\footnote{This is not to be confused with the post-processing of LM \emph{embeddings}, before they are passed to classification layers. In this case, the classification layers must be retrained to account for the modified embeddings.}. The ability to avoid retraining models is a major advantage of post-processing due to the large computational cost of fine-tuning LMs. Post-processing also targets unfairness at a point closest to deployment and hence can have a direct impact on downstream operations that use the model predictions.

\textbf{Hardt, Price, Srebro (2016) (HPS):} The first post-processing method that we consider is by \citet{hardt2016equality} (abbreviated HPS, using the last names of the authors), who were the original proposers of the equalized odds criterion for fairness. We used the open-source implementation of their method from~\citet{bellamy2019ai}, which post-processes binary predictions to satisfy EO while minimizing classification loss. While this method is effective in enforcing EO, one limitation is that it does not offer a trade-off between minimizing the deviation from EO and reducing the loss in accuracy.

\textbf{Fair Score Transformer (FST):} We study the FST method of \citet{Wei2020Optimized}, in part to provide the above-mentioned trade-off, and in part because it is a recent post-processing method shown to be competitive with several other methods (including in-processing). 
FST takes predicted probabilities (referred to as scores) as input and post-processes them to satisfy a fairness criterion. 
We choose \emph{generalized equalized odds} (GEO), a score-based variant of EO, as the fairness criterion and then threshold the output score to produce a binary prediction. The application of FST required attention to three issues: 1) its ability to work with input scores that may not be calibrated probabilities; 2) the choice of fairness parameter $\epsilon$, which bounds the allowed GEO on the data used to fit FST; 3) the choice of binary classification threshold $t$. We consider a range of $\epsilon$ and $t$ values to explore the trade-off between EO and accuracy.
% We fine tune both $\epsilon$ and $t$ which lead sot a trade-off between EO and accuracy.
Due to numerical instability of the FST implementation in the original paper (occasional non-convergence in reasonable time for the Jigsaw dataset), we obtained a closed-form solution for one step in the optimization that leads to a more efficient implementation, running in minutes for all models and all datasets considered. More details on this implementation and the tuning of the parameters can be found in Appendix~\ref{appendix:fst}.

\textbf{Threshold post-processing (TPP):} We also tested the effect of thresholding alone, without fairness-enhancing transformations. We refer to this as \emph{threshold post-processing} (TPP). This simple method corresponds to FST without calibrating the LM outputs, choosing $\epsilon$ large enough so that FST yields an identity transformation, and thresholding at level $t$.

% Lastly, due to the observation above regarding issue 3), we also tested the effect of thresholding alone, without fairness-enhancing transformation. We refer to this as \emph{threshold post-processing} (TPP). In terms of the three issues discussed above, this simple methods corresponds to an identity transformation for FST that can be obtained by 1) not calibrating the LM outputs, 2) choosing $\epsilon$ large enough so that FST yields an identity transformation, and 3) thresholding at level $t$.

\begin{figure}[htb]
\footnotesize
\centering
\begin{subfigure}{0.48\columnwidth}
\centering
Jigsaw
\end{subfigure}
\begin{subfigure}{0.48\columnwidth}
\centering
HateXplain
\end{subfigure}
\newline
\begin{subfigure}{\columnwidth}
\includegraphics[width=\columnwidth]{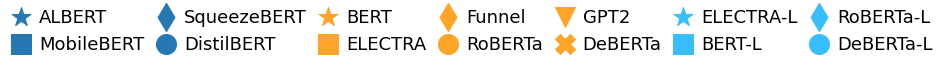}
\end{subfigure}
\begin{subfigure}{0.48\columnwidth}
\includegraphics[width=\columnwidth]{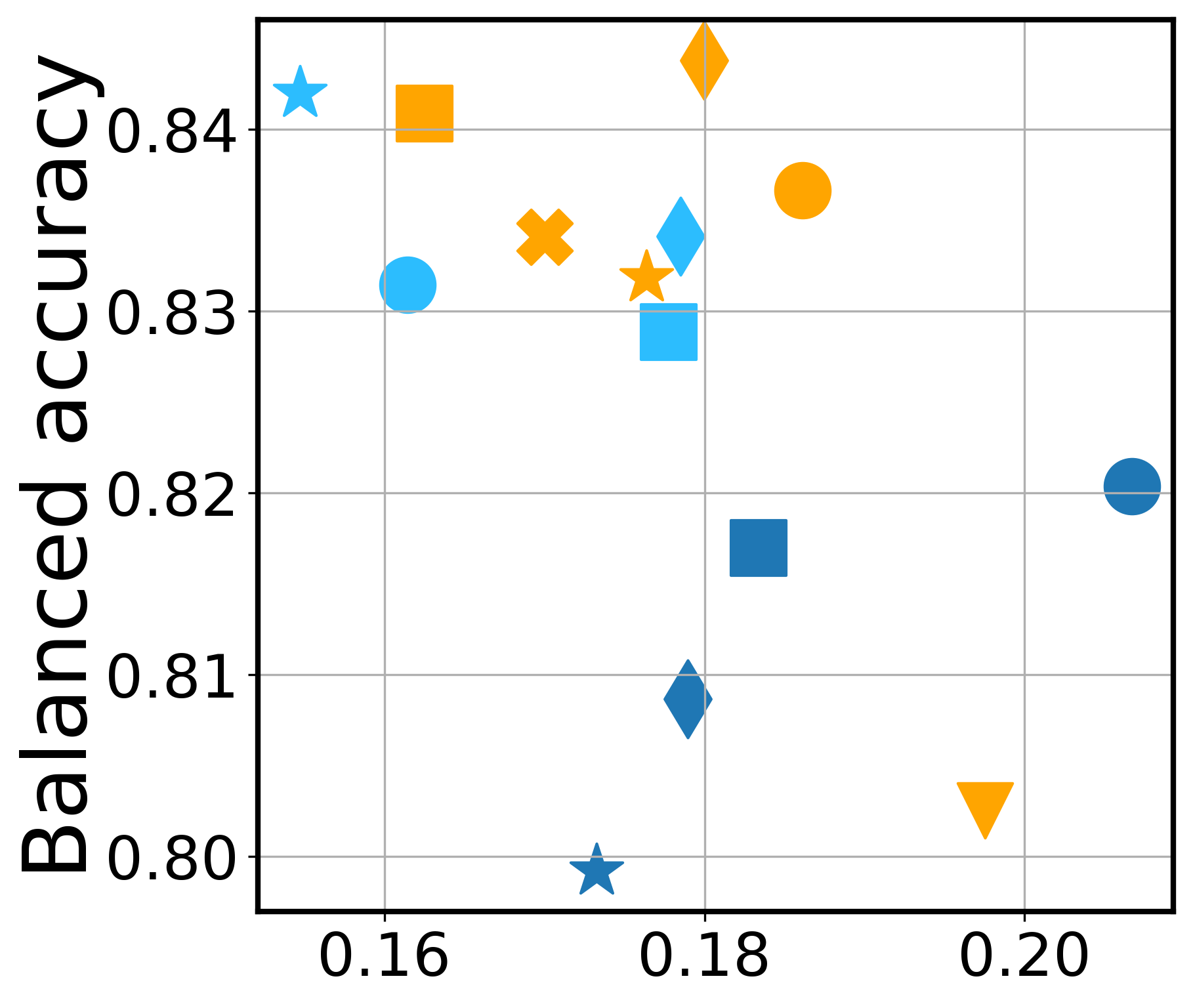}
\end{subfigure}
\begin{subfigure}{0.48\columnwidth}
\includegraphics[width=\columnwidth]{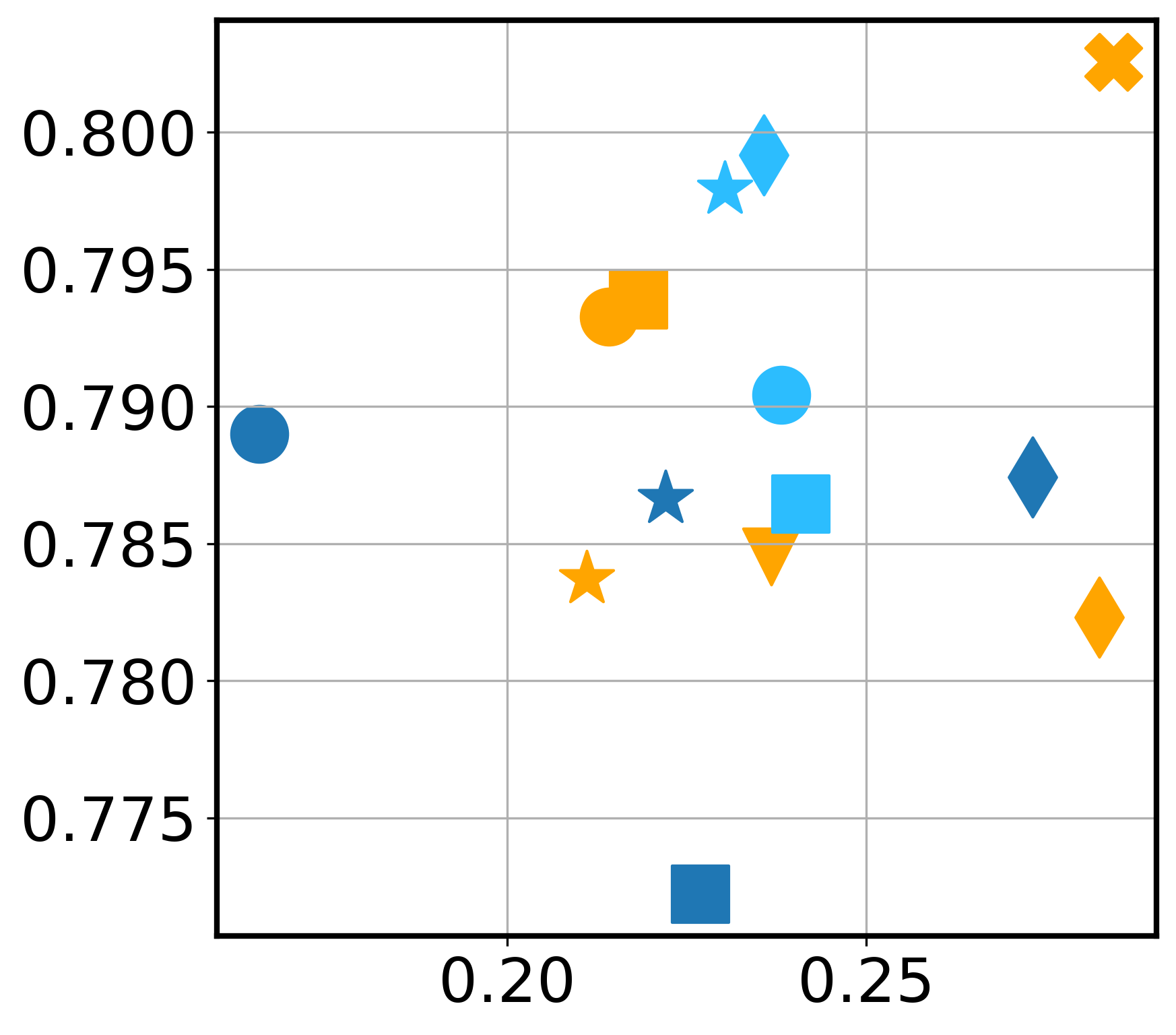}
\end{subfigure}
\newline
\begin{subfigure}{\columnwidth}
\centering
religion
\end{subfigure}
\newline
\begin{subfigure}{0.48\columnwidth}
\includegraphics[width=\columnwidth]{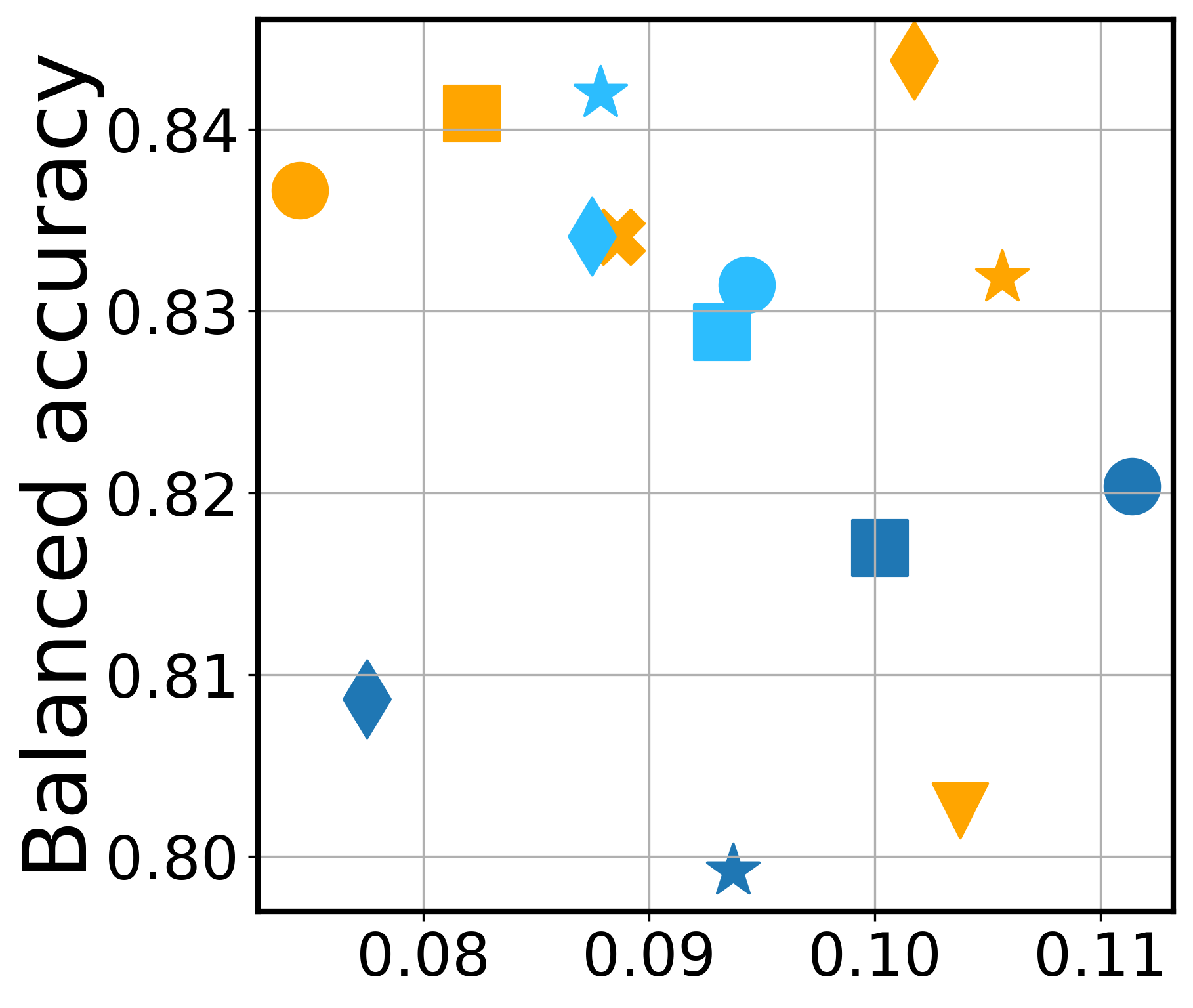}
\end{subfigure}
\begin{subfigure}{0.48\columnwidth}
\includegraphics[width=\columnwidth]{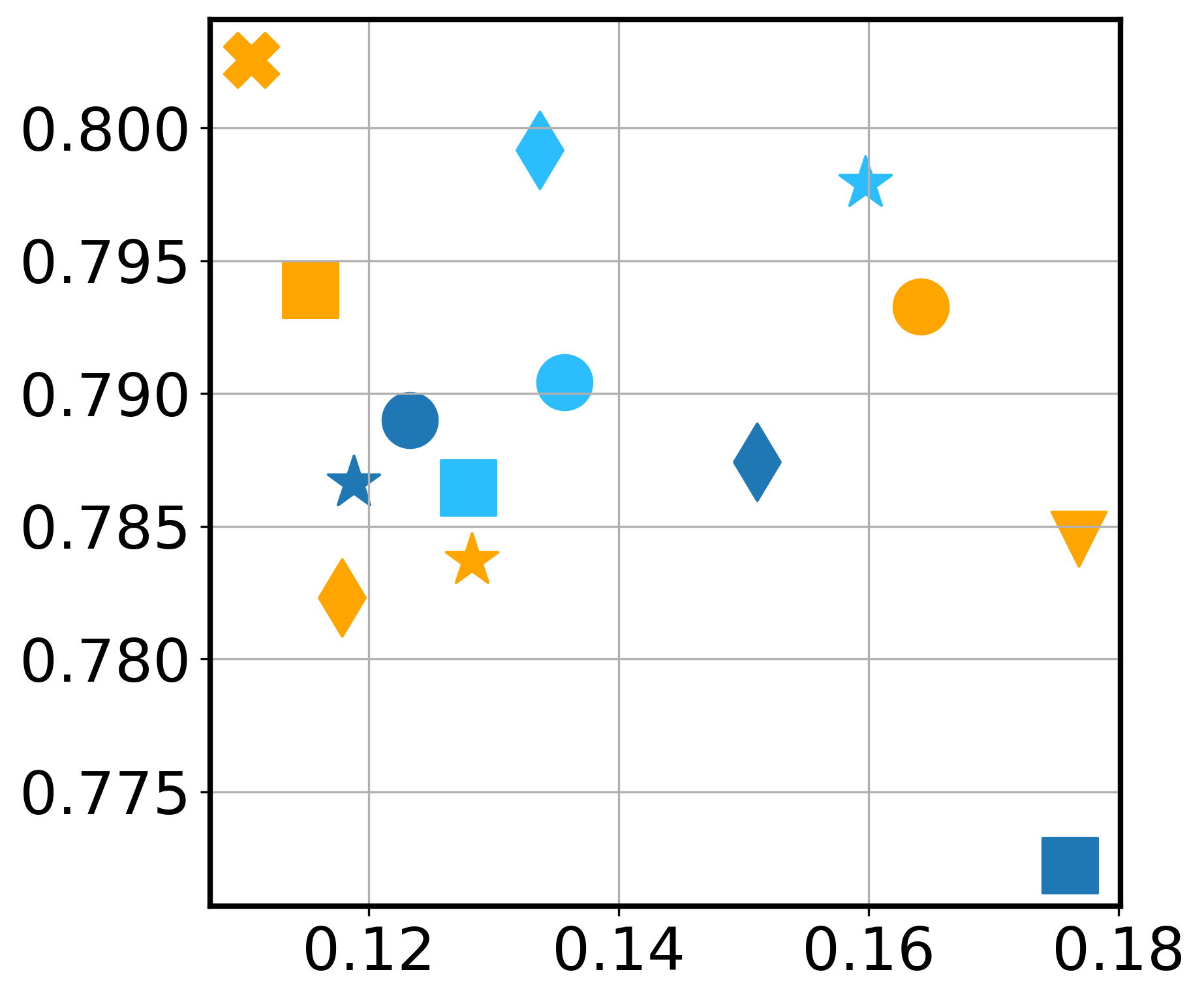}
\end{subfigure}
\newline
\begin{subfigure}{\columnwidth}
\centering
race
\end{subfigure}
\newline
\begin{subfigure}{0.48\columnwidth}
\includegraphics[width=\columnwidth]{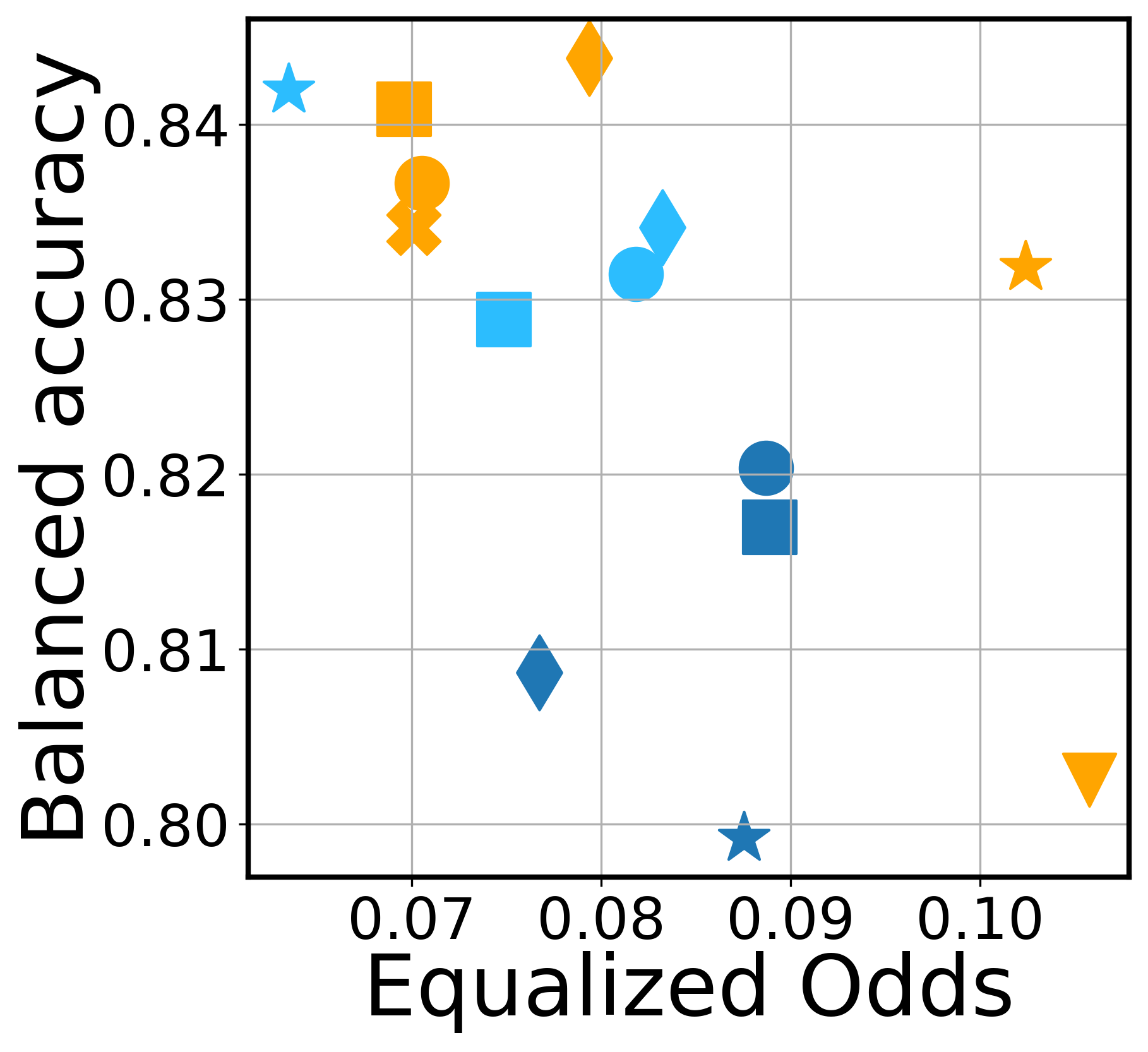}
\end{subfigure}
\begin{subfigure}{0.48\columnwidth}
\includegraphics[width=\columnwidth]{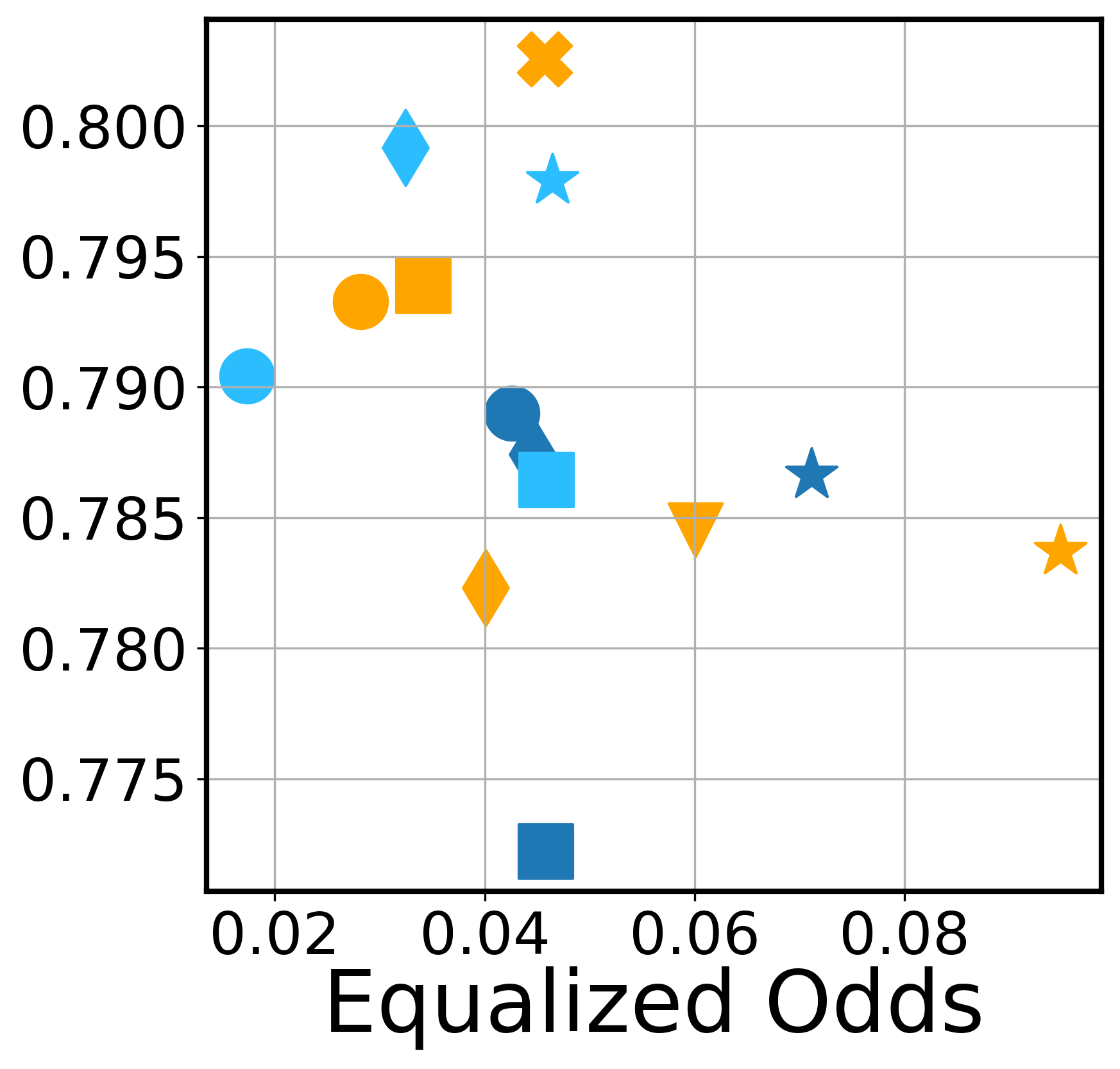}
\end{subfigure}
\begin{subfigure}{\columnwidth}
\centering
gender
\end{subfigure}
\caption{Balanced accuracy versus equalized odds for %several 
fine-tuned LMs on the Jigsaw and HateXplain datasets.}
\label{fig:perf_fair}
\end{figure}

\section{The accuracy-fairness relationship in toxic text classification}

We report on the performance and fairness characteristics of several LMs while varying parameters such as random seeds and training data size. We also experiment with post-processing methods for group bias mitigation and show that it is possible to reduce some of the bias presented by these models.

\subsection{Characterization of language models of varied sizes}

The first set of experiments present how performance and fairness measures vary across models. In Figure~\ref{fig:perf_fair} we show the performance as measured by balanced accuracy\footnote{We use balanced accuracy as a measure for performance as it is more informative, especially for the imbalanced Jigsaw dataset where a trivial predictor that always outputs the label ``normal'' would achieve $\sim$92\% accuracy.} and the group fairness as measured by equalized odds on the $x$-axis (lower EO is better). The models are color-coded by their size - dark blue for small models, orange for regular size models and light blue for large models. The variation in balanced accuracy is not as wide as the variation in equalized odds. For the HateXplain dataset, the gap between balanced accuracy and fairness variability is more prominent. In terms of accuracy (not balanced), the models perform even closer as shown in the plots in Appendix~\ref{appendix:perf_plots}. For EO, the spread is significant, with gaps of $0.10$ between the largest and smallest values for Jigsaw, and $0.15$ for HateXplain. Depending on the dataset and sensitive group, some larger models seem to lead to lower EO; for example, ELECTRA-large achieves the best accuracy-EO results for religion as the sensitive group (Jigsaw). For race, SqueezeBERT, which is one of the small models in the study, achieves one of the best balanced accuracy-EO operating points for Jigsaw (considering it is half the size of RoBERTa which has better balanced accuracy but similar EO), hinting that size is not well correlated with the fairness of the model. Similarly, for HateXplain (religion), DistilBERT, again a small model, obtains the best balanced accuracy-EO operating point. In the next section, we analyze models trained using various random seeds and find a low correlation between EO and model size.

These results strongly suggest that fairness measures should be included in the evaluation of LMs. In the next sections, we demonstrate that, if fairness is not carefully considered, we can end up with models with widely varying fairness characteristics depending on the training conditions.

\subsection{The influence of random seeds}% on accuracy and fairness}
\begin{figure}[htb]
\footnotesize
\centering
\begin{subfigure}{0.48\columnwidth}
\centering
Jigsaw
\end{subfigure}
\begin{subfigure}{0.48\columnwidth}
\centering
HateXplain
\end{subfigure}
\newline
\begin{subfigure}{\columnwidth}
\includegraphics[width=\columnwidth]{acl-2022-plots-CR/lms-perf-fairness-legend.png}
\end{subfigure}
\begin{subfigure}{0.48\columnwidth}
\includegraphics[width=\columnwidth]{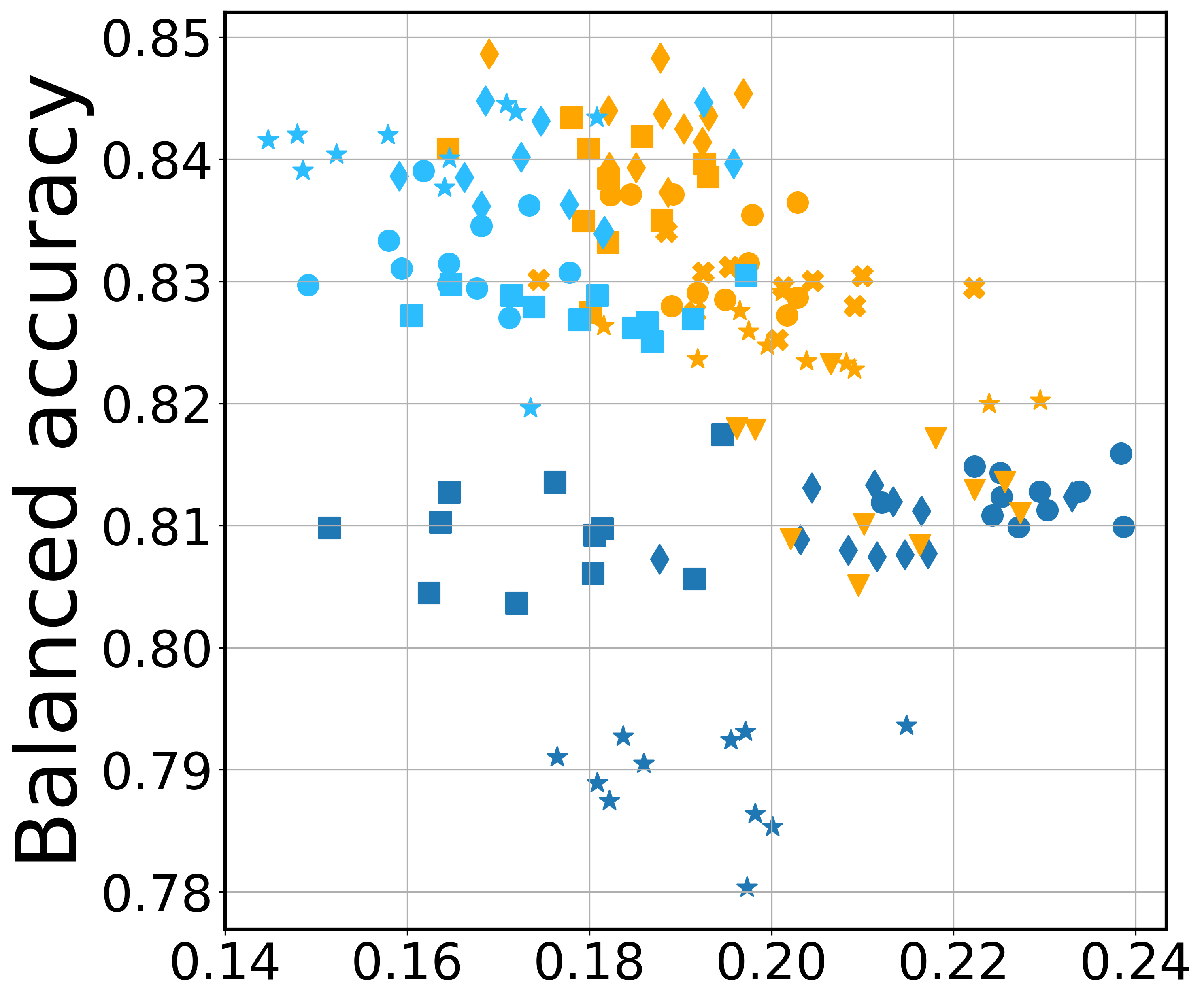}
\end{subfigure}
\begin{subfigure}{0.46\columnwidth}
\includegraphics[width=\columnwidth]{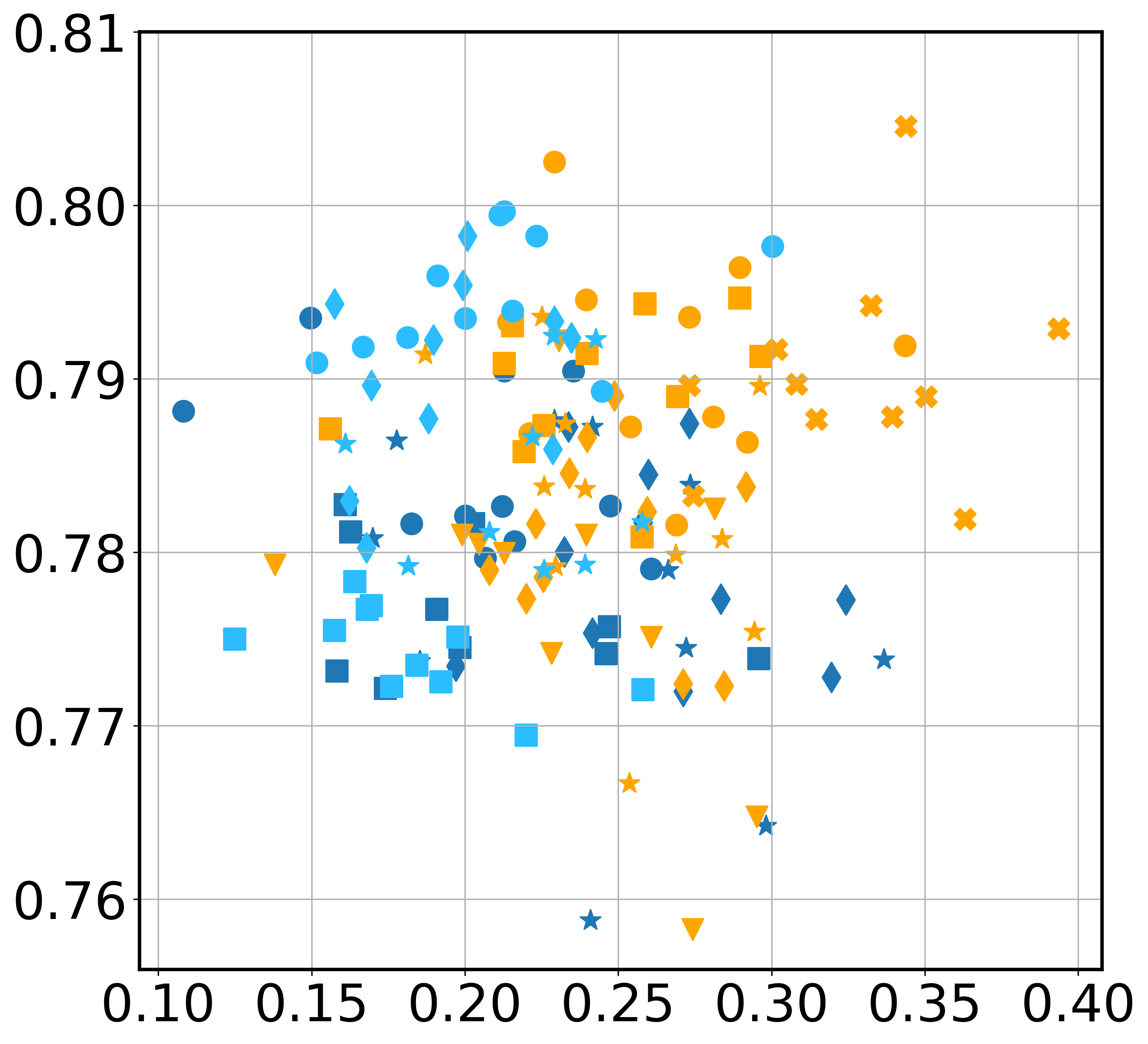}
\end{subfigure}
\newline
\begin{subfigure}{\columnwidth}
\centering
religion
\end{subfigure}
\newline
\begin{subfigure}{0.48\columnwidth}
\includegraphics[width=\columnwidth]{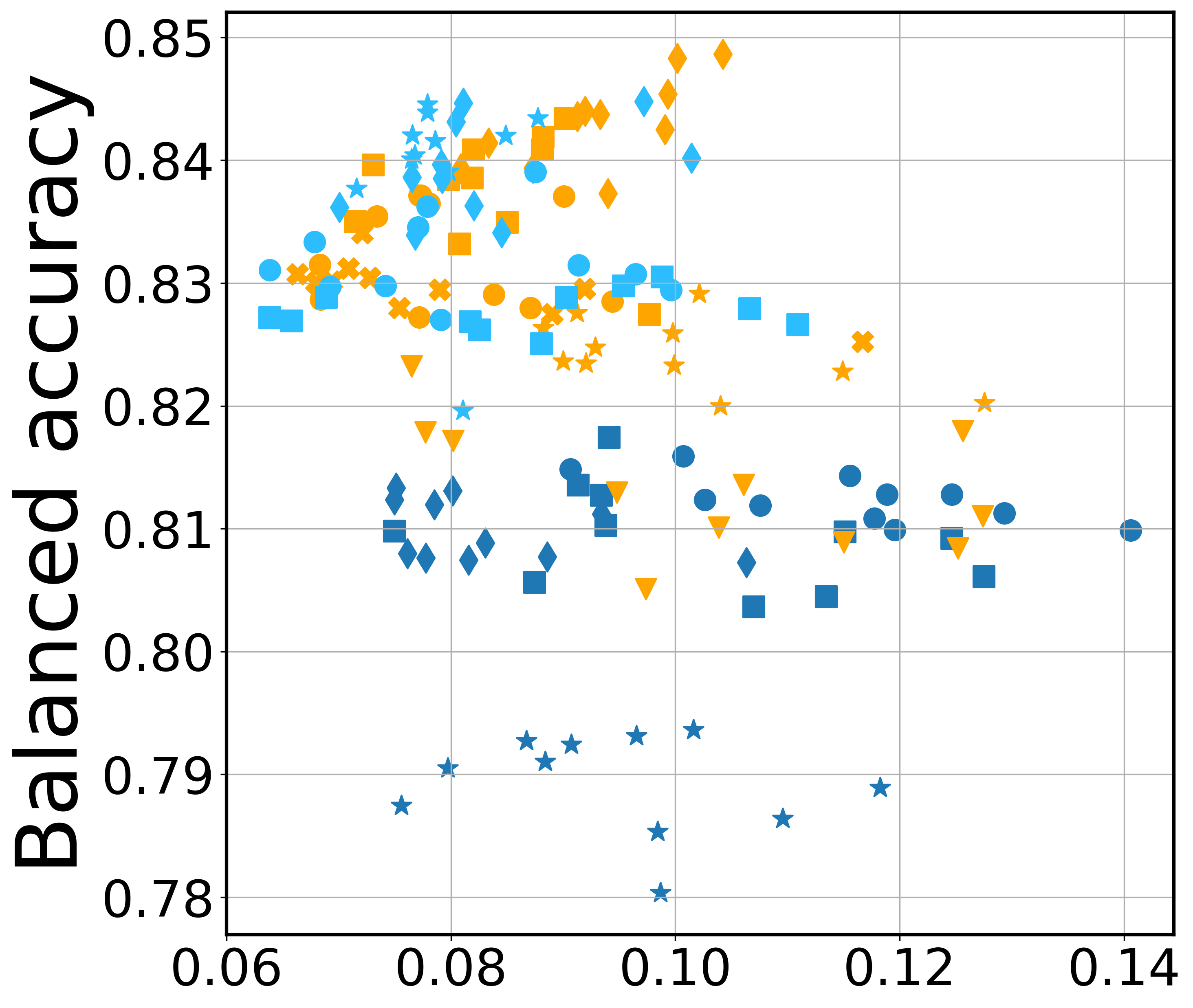}
\end{subfigure}
\begin{subfigure}{0.46\columnwidth}
\includegraphics[width=\columnwidth]{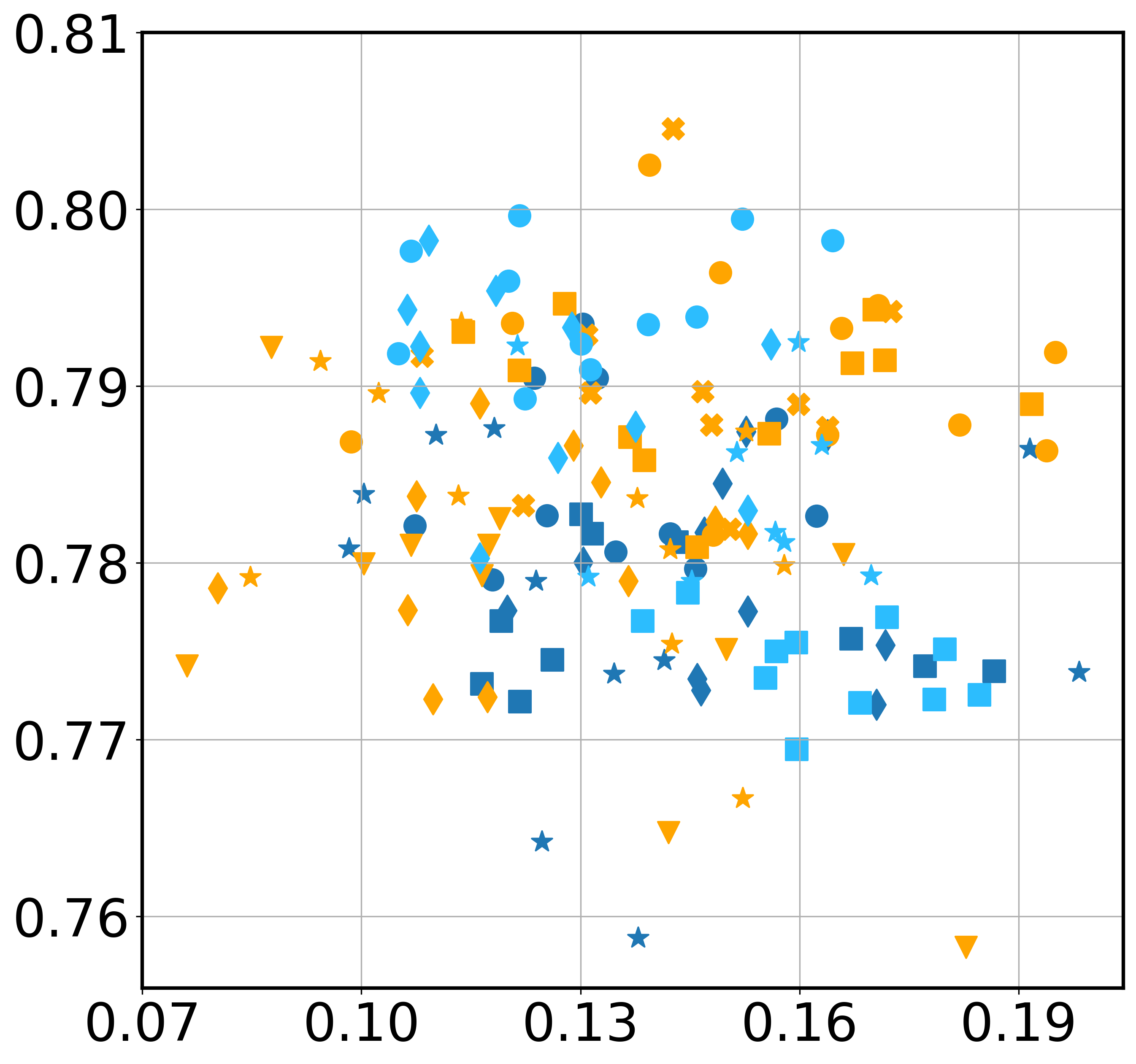}
\end{subfigure}
\newline
\begin{subfigure}{\columnwidth}
\centering
race
\end{subfigure}
\newline
\begin{subfigure}{0.48\columnwidth}
\includegraphics[width=\columnwidth]{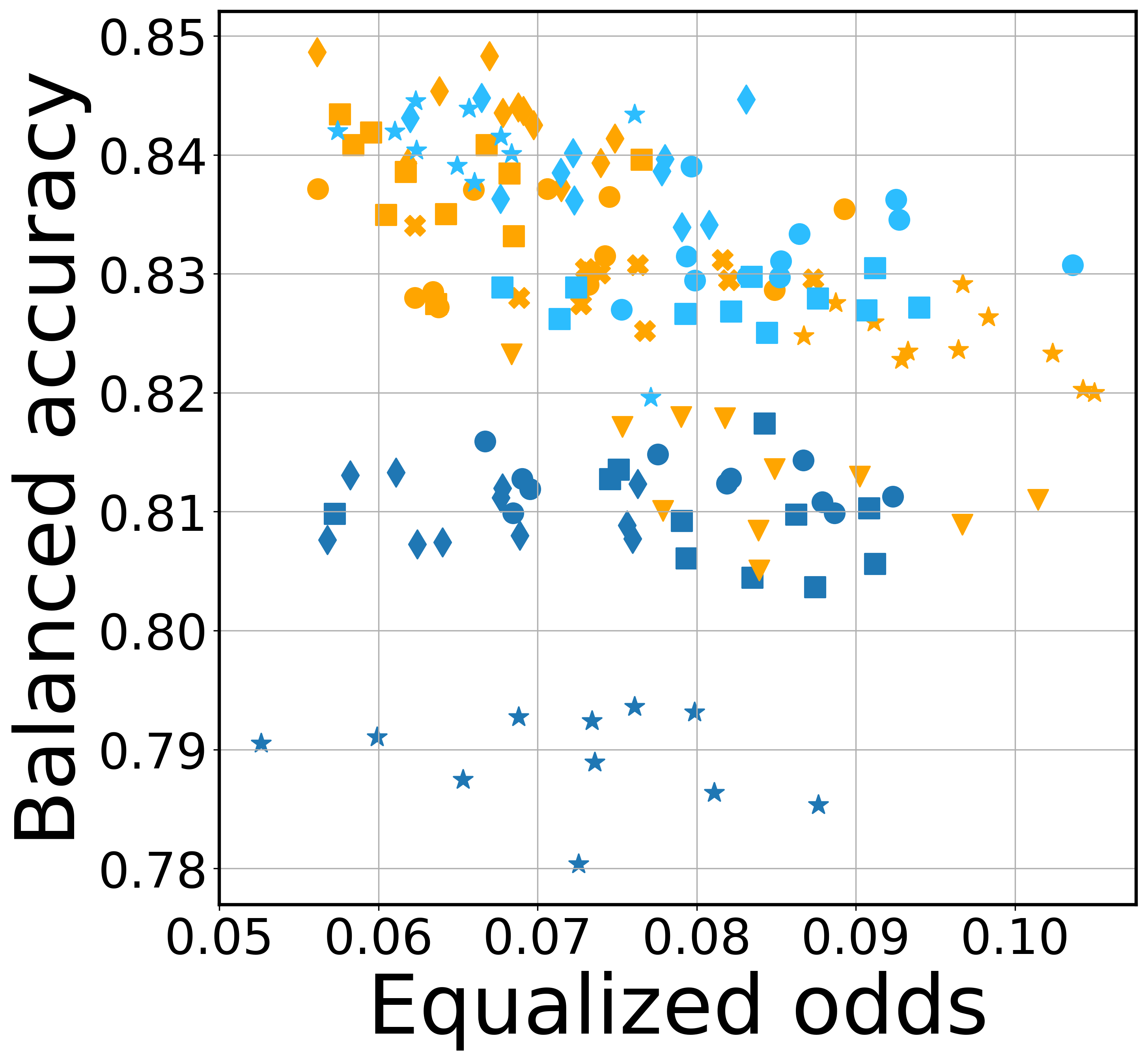}
\end{subfigure}
\begin{subfigure}{0.46\columnwidth}
\includegraphics[width=\columnwidth]{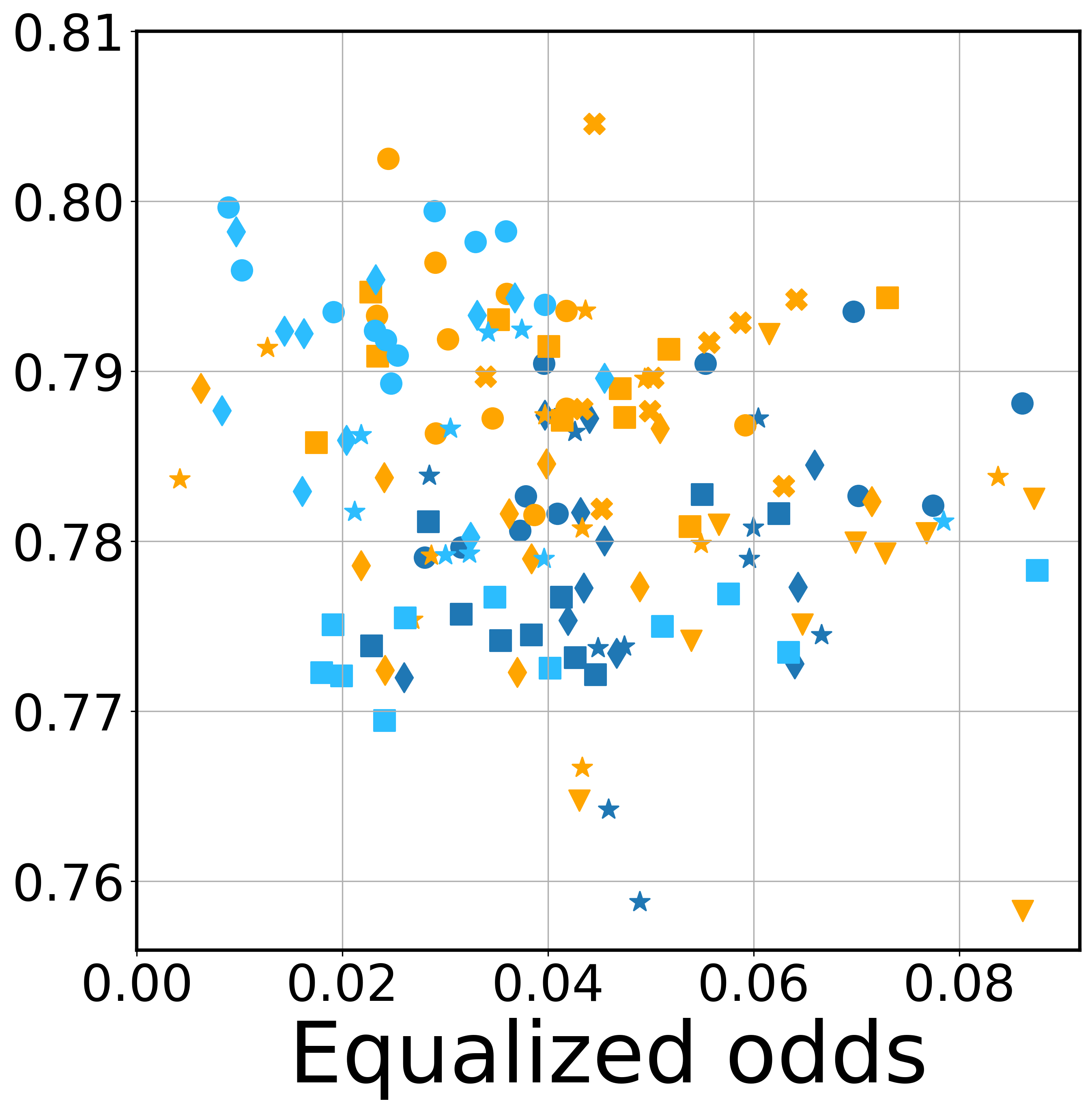}
\end{subfigure}
\begin{subfigure}{\columnwidth}
\centering
gender
\end{subfigure}
\caption{Balanced accuracy versus equalized odds for fine-tuned LMs when varying the random seed used in fine-tuning.}
\label{fig:rand_seeds}
\end{figure}

\label{section:seeds}
Fine-tuning LMs depends on a random seed used for mini-batch sampling and for initializing the weights in the last layers of the network responsible for the binary classification. It is well documented in the literature that this random seed may influence the accuracy of the resulting model~\citep{dodge2020Finetuning}. In Figure~\ref{fig:rand_seeds} we show that while balanced accuracy is somewhat stable, fairness can vary widely by only changing the random seed.
% In the set of experiments we ran, we discover that, for the datasets and tasks considered, the balanced accuracy is somewhat stable while model fairness can vary widely by only changing the random seed. The results are shown in Figure~\ref{fig:rand_seeds}. 
In fact, if we were to plot the accuracy instead of the balanced accuracy, all points would be virtually on a horizontal line for Jigsaw, as shown in Figure~\ref{appendix:perf_plots}. The variations for EO are larger. For Jigsaw, we observe a variation of up to 0.05 in equalized odds for some cases.
For HateXplain, the variation is considerably larger, with several models presenting a spread of 0.15 or more for the sensitive group of religion. For example, for DeBERTa-L, depending on the random seed, one could get one of the best models with respect to performance-fairness trade-offs, or one of the worst (balanced accuracy varies within 0.79-0.80, while EO varies over 0.11-0.30). The results in our experiments align with the ones discussed in a recent study on underspecification in machine learning~\citep{Amour2020Underspecification}, where different random seeds lead to small variations in accuracy, but considerable variations in intrinsic bias as measured by gendered correlations.

% To further probe whether there is a correlation between fairness and model size, we used the results for random seeds to fit a linear regression model for the equalized odds measure as a function of the log of model size on disk. We fit one linear model per protected group. The $R^2$ measures are $0.128$, $0.115$, and $0.007$ for the religion, race, and gender groups in Jigsaw; for HateXplain these measures are $0.035$, $0.0$, and  $0.078$ for the same groups. This clearly shows very low correlation between fairness and model size. Similar conclusions can be drawn from Pearson's coefficient of correlation with values of -0.357 for Jigsaw and -0.188 for HateXplain, with p-values of .000005 and 0.017, respectively.

To further probe whether there is a correlation between fairness and model size, we used the results for multiple random seeds to compute Pearson's coefficient of correlation. These values are -0.357 for Jigsaw and -0.188 for HateXplain, with p-values of 5e-6 and 0.017, respectively. These results show a low correlation between fairness as measured by EO and model size.

% \begin{figure*}[h]
% \footnotesize
% \begin{center}
% \begin{tabular}{ccc}
% & Jigsaw Dataset &\\
%  \includegraphics[scale=0.12]{img/jigsaw_seeds_religion.png}&
%  \hspace{-.4cm}\includegraphics[scale=0.12]{img/jigsaw_seeds_race.png}&
%  \hspace{-.4cm}\includegraphics[scale=0.12]{img/jigsaw_seeds_gender_and_sexual_orientation_legend.png}
%  \\
%  & HateXplain Dataset &\\
%  \includegraphics[scale=0.12]{img/hatex_random_seeds_religion.png}&
%  \hspace{-.4cm}\includegraphics[scale=0.12]{img/hatex_random_seeds_race.png}&
%  \hspace{-.4cm}\includegraphics[scale=0.12]{img/hatex_random_seeds_gender_legend.png}
%  \\
%  a) religion & b) race & c) gender\\
% \end{tabular}
% \end{center}
% \vspace{-.5cm}
% \caption{Balanced accuracy versus equalized odds for several fine-tuned LMs when varying only the random seed used in fine-tuning.}
% \label{fig:rand_seeds}
% \end{figure*}

\subsection{Low data regime}

\begin{figure}[htb]
\footnotesize
\centering
\begin{subfigure}{0.48\columnwidth}
\centering
Jigsaw
\end{subfigure}
\begin{subfigure}{0.48\columnwidth}
\centering
HateXplain
\end{subfigure}

\begin{subfigure}{0.48\columnwidth}
\includegraphics[width=\columnwidth]{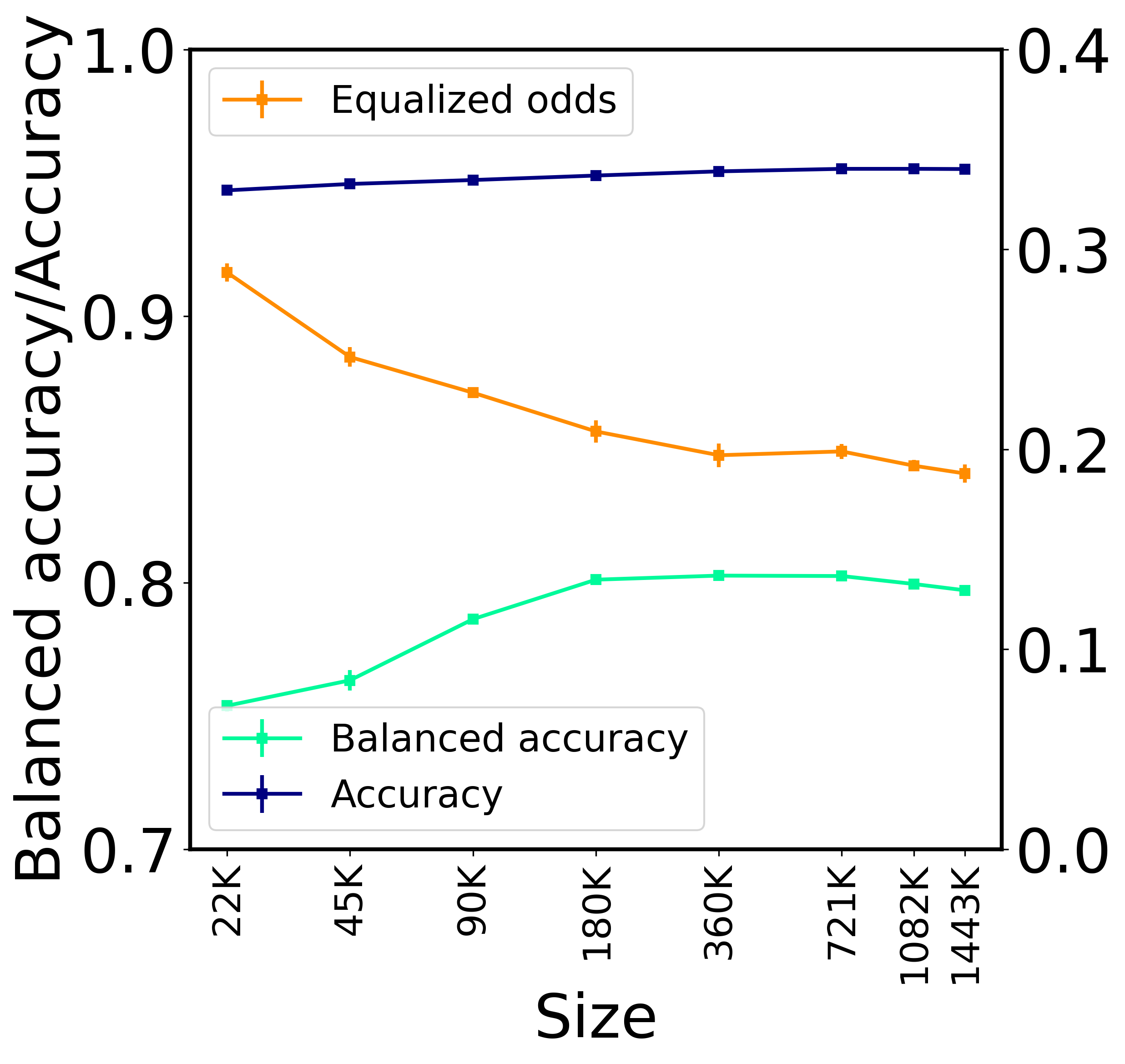}
\end{subfigure}
\begin{subfigure}{0.48\columnwidth}
\includegraphics[width=\columnwidth]{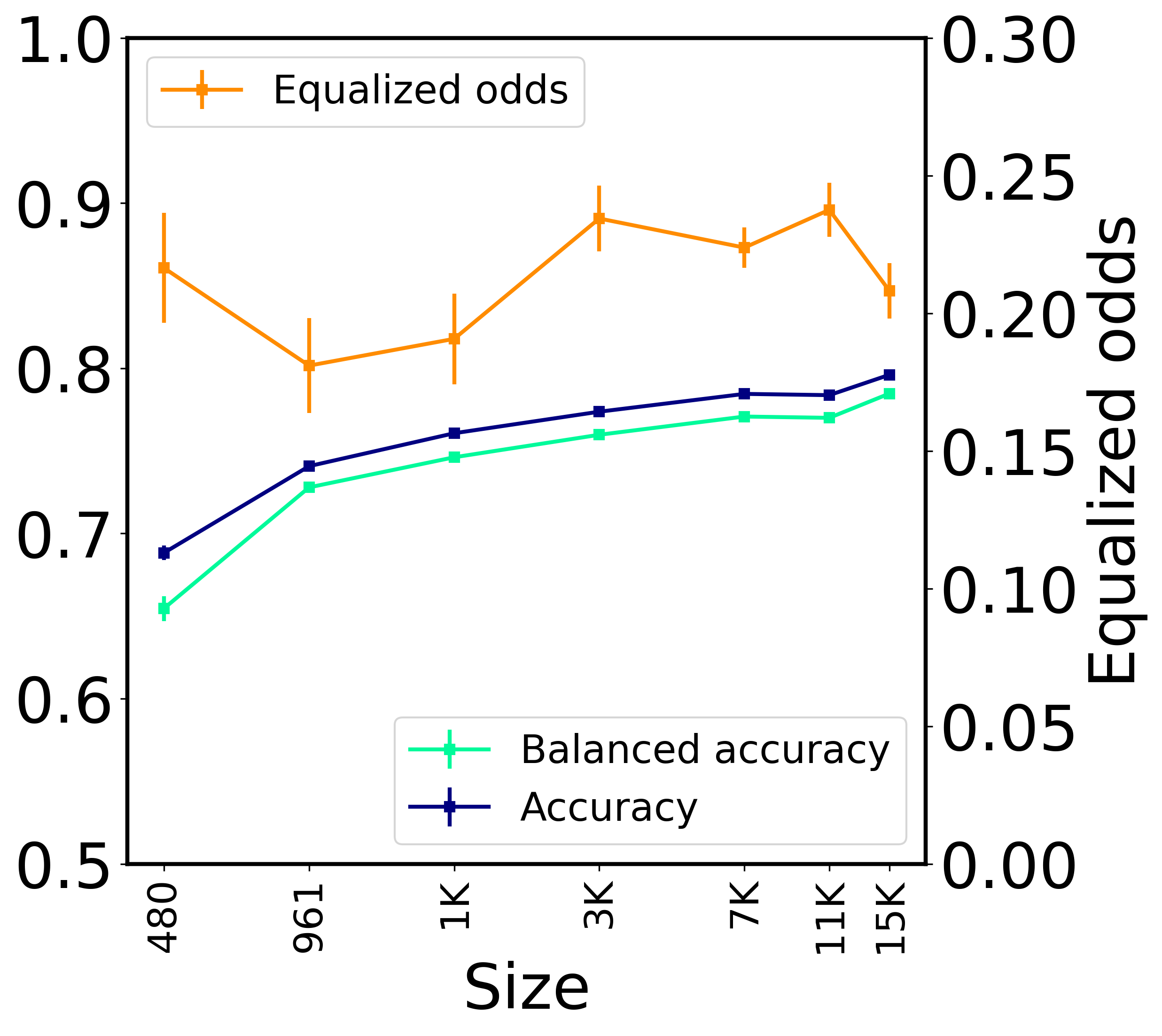}
\end{subfigure}
\newline
\begin{subfigure}{\columnwidth}
\centering
DistilBERT
\end{subfigure}
\newline
\begin{subfigure}{0.48\columnwidth}
\includegraphics[width=\columnwidth]{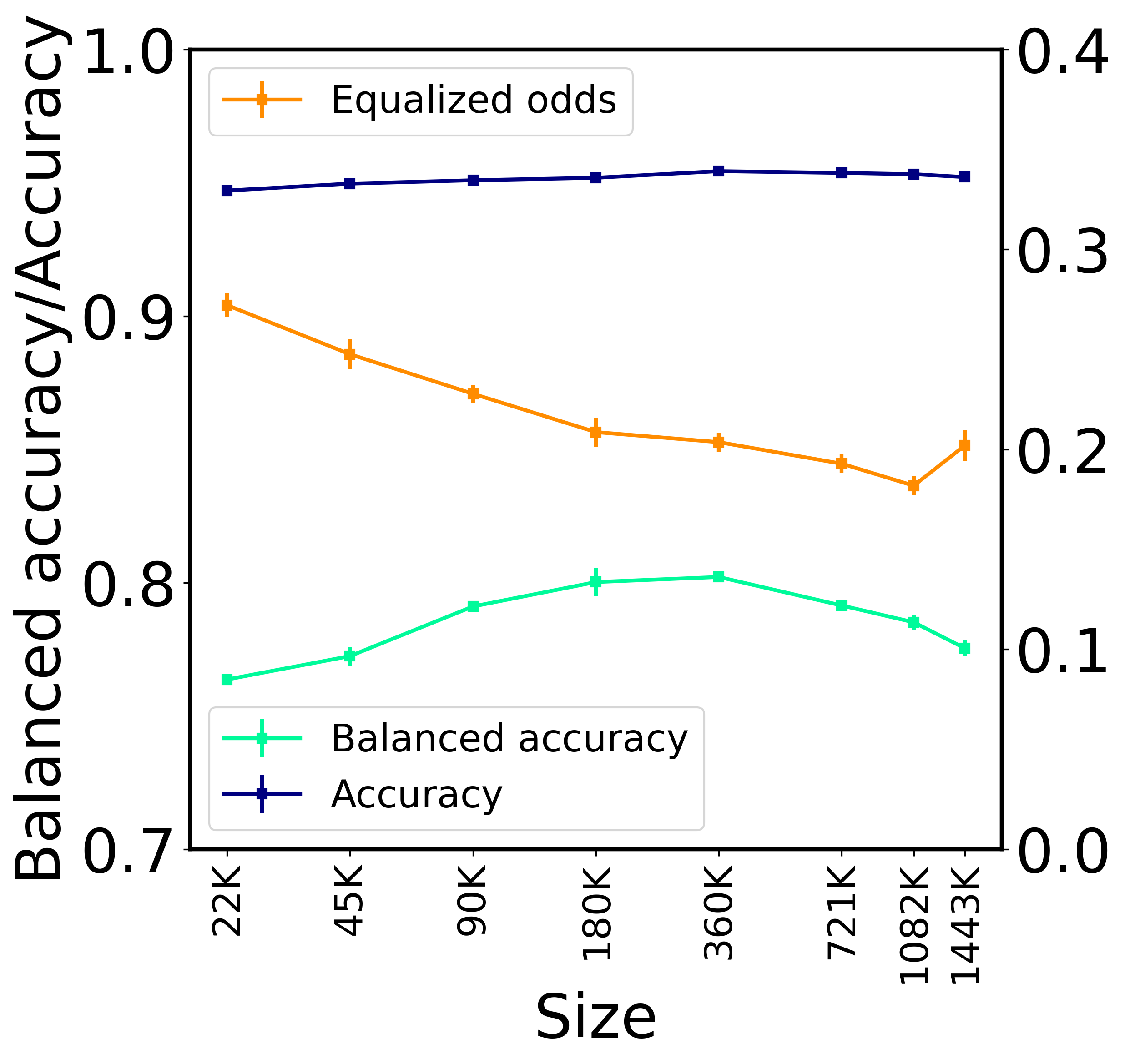}
\end{subfigure}
\begin{subfigure}{0.48\columnwidth}
\includegraphics[width=\columnwidth]{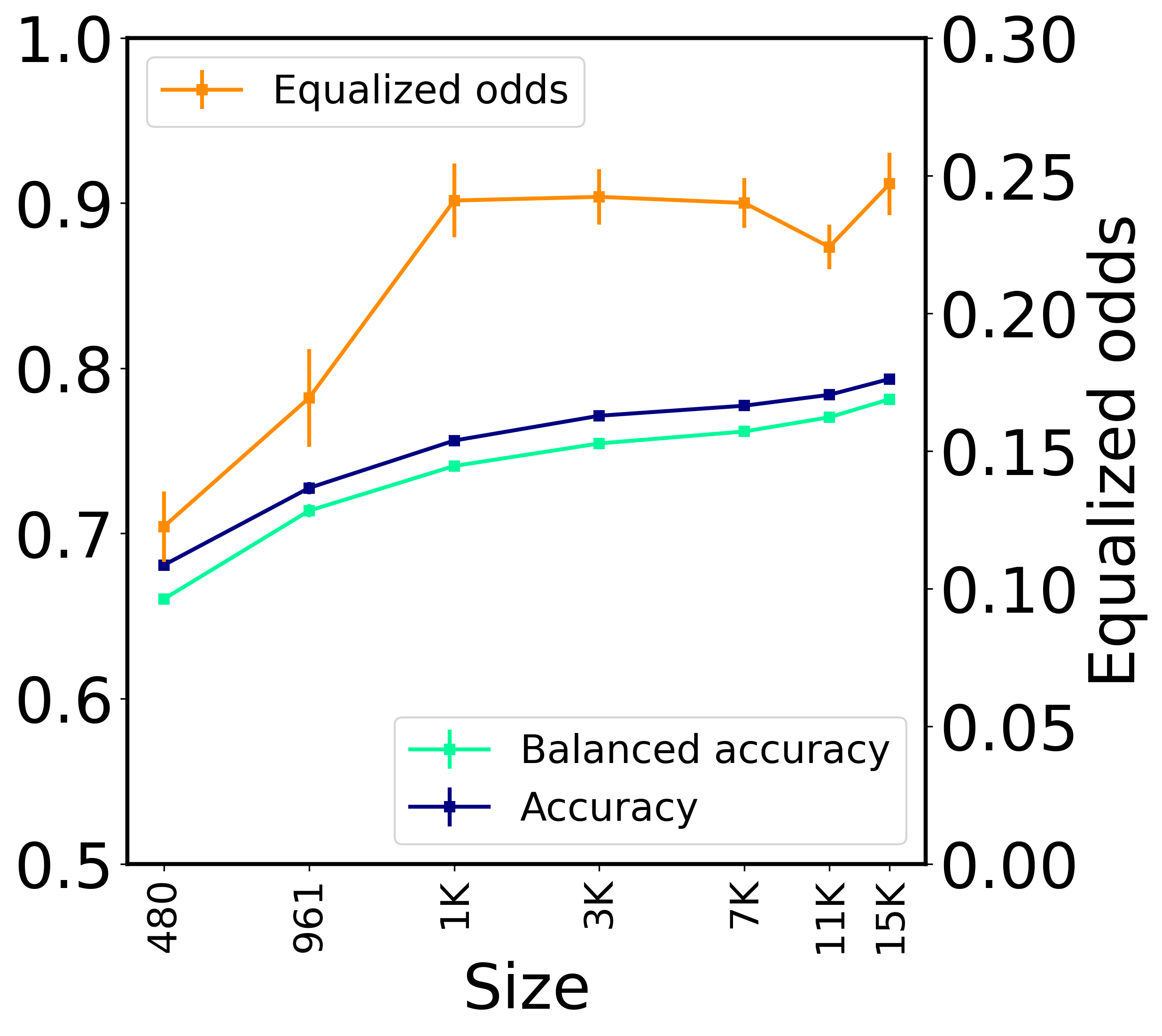}
\end{subfigure}
\newline
\begin{subfigure}{\columnwidth}
\centering
BERT
\end{subfigure}
\newline
\begin{subfigure}{0.48\columnwidth}
\includegraphics[width=\columnwidth]{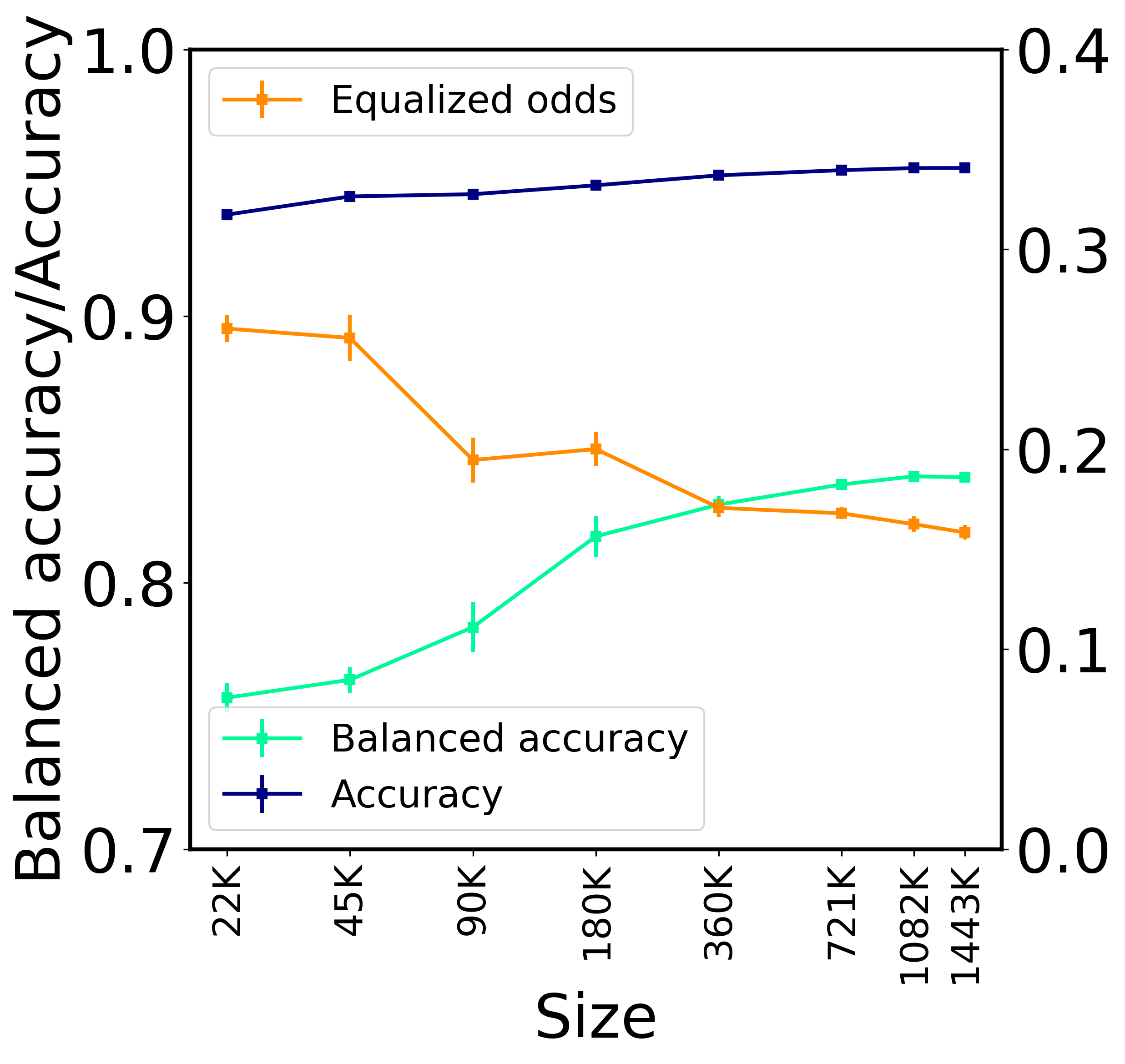}
\end{subfigure}
\begin{subfigure}{0.48\columnwidth}
\includegraphics[width=\columnwidth]{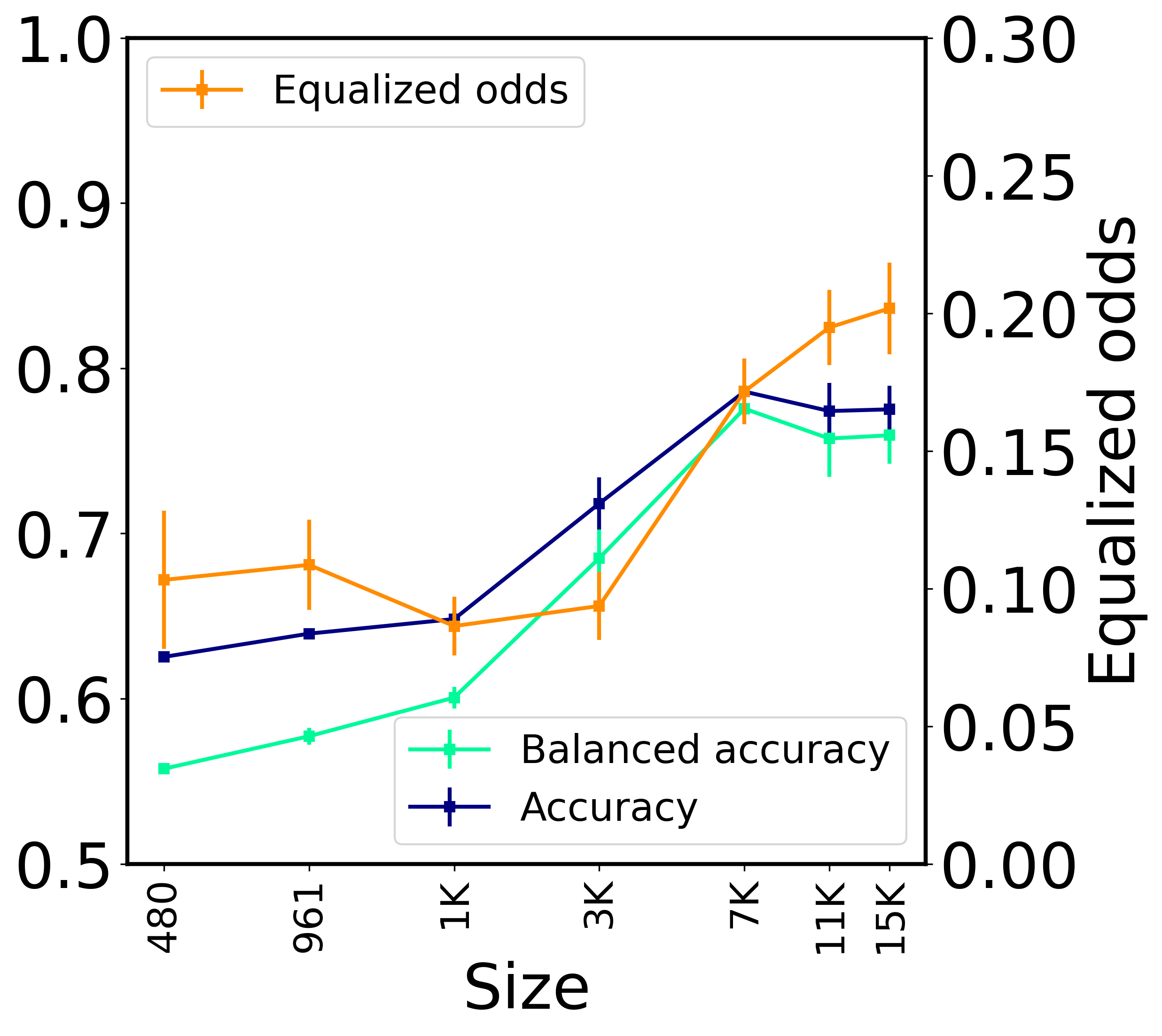}
\end{subfigure}
\begin{subfigure}{\columnwidth}
\centering
ELECTRA-large
\end{subfigure}
\caption{Accuracy, balanced accuracy and equalized odds (religion) for fine-tuned LMs when varying the fine-tuning data size and the random seeds. Error bars denote $\pm 1$ SE (standard error) of the mean.}
\label{fig:low_data}
\end{figure}

In general, it is well known that more training data improves model accuracy.
We experiment with fine-tuning the models using a fraction of the training dataset, while keeping the test set the same. When the smaller datasets are subsampled from the original dataset, we ensure that the larger datasets include the smaller ones to simulate situations when more data is collected and used for training. The results are shown for one small/regular/large model in Figure~\ref{fig:low_data}. Each data point in the graph represents the average of eleven runs performed with different random seeds, one for each run. In very few cases, the random seed led to a degenerate model and we did not include these runs in the averaged results. Overall, there were up to five degenerate runs for each dataset (across all 14 models in this study, not only the ones presented in the figure).

We observe that in the case of Jigsaw, equalized odds generally keeps improving even when the accuracy plateaus, suggesting that, from a fairness point of view, it may be beneficial to collect more data for fine-tuning. This does not seem to be the case for the HateXplain dataset, where the accuracy does not plateau and the fairness measure oscillates. A reason could be that HateXplain is much smaller in size than Jigsaw and hence Jigsaw's training is more stable. Similar trends are observed for the rest of the models in our study.

\begin{table*}[tb]
\footnotesize
\begin{tabular}{l|cccc||ccc||cccc}
& Religion & Christian & Jewish & Muslim & Race & White & Black & Gender & Female & Male & LGBT\\
\hline
Baseline & 0.18 & 0.10 & 0.06 & 0.20 & 0.10 & 0.12 & 0.13 & 0.10 & 0.12 & 0.13 & 0.15 \\
FST & 0.08 & 0.03 & 0.06 & 0.11 & 0.09 & 0.11 & 0.11 & 0.05 & 0.07 & 0.07 & 0.15\\
\end{tabular}
\caption{BERT (Jigsaw): Equalized odds before and after applying FST for all sensitive groups and their subgroups.}
\label{table:subgroups}
\end{table*}

\subsection{Bias mitigation through post-processing}% methods}

\begin{figure}[htb]
\centering
\footnotesize
\includegraphics[width=.8\columnwidth]{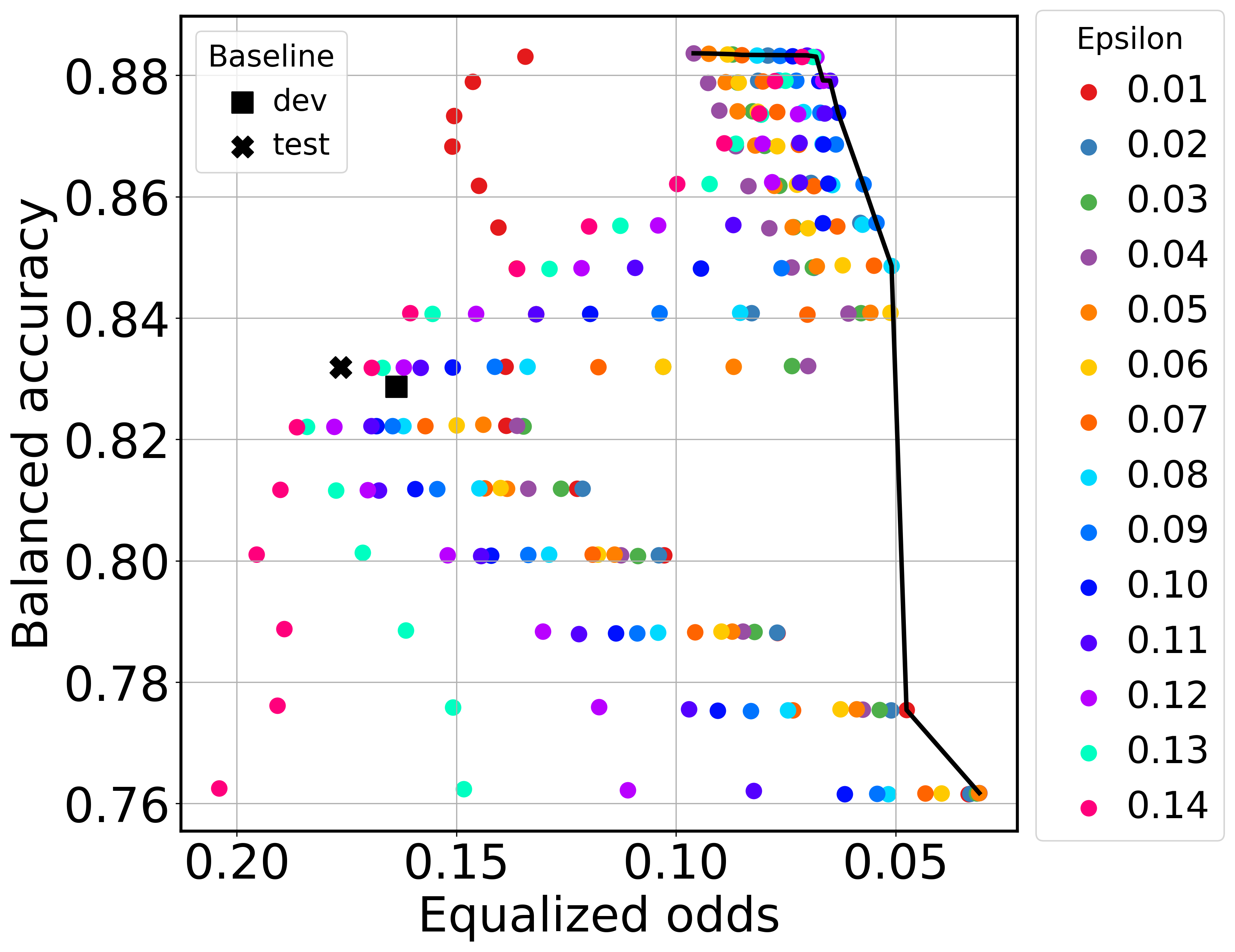}
\caption{FST tuning for BERT: Balanced accuracy versus equalized odds on the Jigsaw dataset when varying fairness parameter $\epsilon$ and classification threshold $t$ for the FST method for group bias mitigation (religion).}
\label{fig:jigsaw_disco_religion}
\end{figure}

In this section, we experiment with applying post-processing methods for group bias mitigation. We first discuss the results of parameter tuning for Fair Score Transformer (FST)~\citep{Wei2020Optimized}. More details can be found in Appendix~\ref{appendix:fst}. The FST method has one tunable parameter, $\epsilon$. 
Using the transformed scores from FST, we also investigate tuning the threshold used in the binary classifier, instead of using the default value of 0.5, as explained in Section~\ref{sec:method:post}. Figure~\ref{fig:jigsaw_disco_religion} depicts the data points obtained by varying $\epsilon$ and the classification threshold~\footnote{All points are shown for the dev set as this plot illustrates the tuning of FST parameters.}. Note that we plot EO decreasingly on the x-axis, and overall better operating points are closer to the top right corner. When choosing an operating point, the points on the black Pareto frontier are the most interesting points: highest balanced accuracy and lowest equalized odds. For reference, we also show the baseline points without bias mitigation for the dev and test sets. All data points are plotted for fine-tuned BERT. Similar trends are observed for the rest of the models considered in this study and for the HateXplain dataset.

\begin{figure}[htb]
\includegraphics[width=\columnwidth]{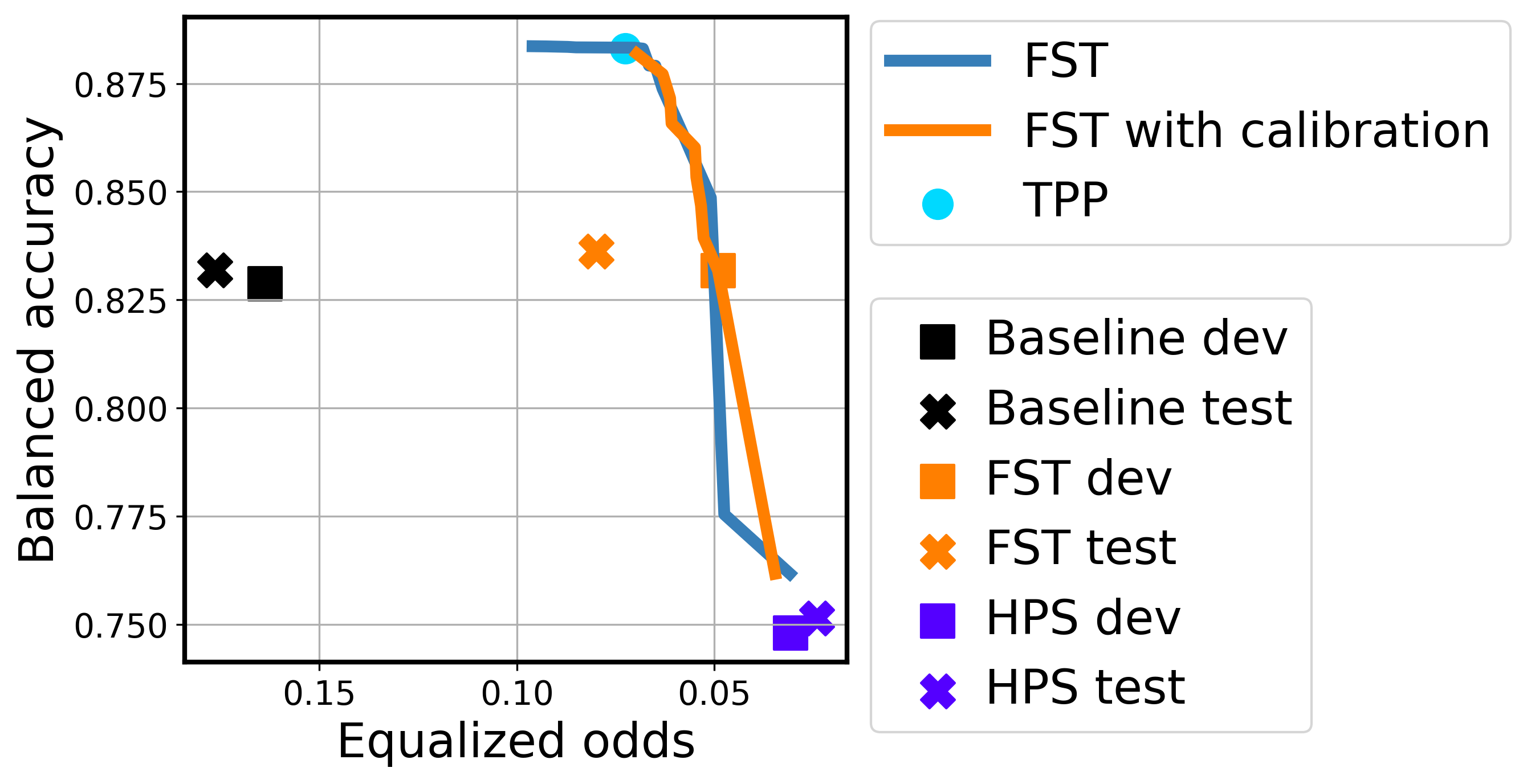}
\caption{BERT: Balanced accuracy versus equalized odds on the Jigsaw dataset when applying the FST and HPS methods for group bias mitigation and threshold post-processing (TPP) alone (religion).}
\label{fig:jigsaw_bias_mitigation_religion}
\end{figure}

We also experimented with calibrating the scores using logistic regression before post-processing. In Figure~\ref{fig:jigsaw_bias_mitigation_religion}, we plot the Pareto frontiers of bias mitigation when applying FST, with and without calibration, along with the threshold post-processing (TPP) method. We also show the result of HPS, which yields a single operating point, as well as the baselines without bias mitigation. In general on the Jigsaw dataset, FST is successful in reducing EO with different degrees of success depending on the model/group (see Appendix~\ref{appendix:bias_mitigation} for additional plots), offering an interesting set of points with different accuracy-EO trade-offs. For reference, we show the corresponding point for the test set (orange \texttt{x}) for the operating point in dev that achieves an equalized odds of at most 0.05 (orange square). In certain cases, FST manages to lower the equalized odds with minimal or no decrease in accuracy, as seen for religion in Figure~\ref{fig:jigsaw_bias_mitigation_religion}. 
Note that all points in the plots except for the \texttt{x} points are plotted using the dev split.

In comparison, HPS seems particularly effective in lowering the equalized odds and thus improving the fairness of the model, with some penalty on the accuracy. For Jigsaw, applying only TPP (i.e., tuning the threshold used in the binary classification) also offers some interesting operating points. TPP has a small search space compared to FST and sometimes the Pareto frontier is reduced to one point, as is the case in Figure~\ref{fig:jigsaw_bias_mitigation_religion}. In general, FST has superior Pareto frontiers compared to TPP alone. In addition, as we discuss in Appendix~\ref{appendix:bias_mitigation}, TPP proved inefficient for the HateXplain dataset. Last, using score calibration before feeding the scores to FST does not seem to offer significant improvements. Similar trends can be observed for the rest of the models. 

Overall, we find the post-processing methods for bias mitigation worth considering. They are straightforward to apply, run in the order of seconds or minutes on the CPU of a regular laptop and they offer interesting operating points. On the other hand, pre-processing or in-processing techniques for bias elimination would incur significant computational cost. Obtaining the Pareto frontiers is instantaneous as the search space for FST is not that large. For more results and discussion of bias mitigation, we refer the reader to Appendix~\ref{appendix:bias_mitigation}.

\subsection{Sensitive groups and subgroups}
\label{sec:subgroups}

In our analysis so far, we looked at sensitive groups that refer to religion, race and gender. In this section we use the Jigsaw dataset to zoom in and analyze the equalized odds for a sensitive group and its constituent subgroups. We select all subgroups that have at least 100 samples in the test split. We continue to apply FST only at the larger group level (e.g., religion) and examine its effect on subgroups. In Table~\ref{table:subgroups}, we show the EO measure for BERT before and after applying FST for all sensitive groups and subgroups. FST consistently manages to lower EO for individual subgroups, without overly favoring one subgroup over another. There are a few instances that do not observe any change, mostly the smallest subgroups. Note that subgroups can be overlapping since they do not represent identities of individuals, instead they derive from the text which may mention multiple subgroups. One notable example is that male and female subgroups have similar EO, both baseline and after FST. This justifies using larger sensitive groups for fitting FST since it seems the discussion of gender overall is problematic as opposed to one gender in particular. 

\section{Limitations}
\label{section:limitations}
%We were also limited by our computation resources. The addition of A100 GPUs to our computing cluster enabled us to have shorter turnaround for our experiments. Most models finished fine-tuning in a couple of hours.
In our study, we covered a series of different models that varied in network architecture, size as number of parameters, training procedures, and pretraining data. As we did not keep any of the elements constant (e.g., architecture) while varying the rest (e.g., pretraining data, size, training procedure), it is hard to draw insights on how each individual element affects the fairness of the resulting prediction outcomes. 
%Our intent was to show that no blanket statement can be made with respect to model/size/training and a careful analysis is required for all models we consider in research and/or deployment.
We would like to emphasize that identifying toxic text is not an easy task, not even for humans. As such, we expect the datasets to be noisy and contain samples that are not annotated correctly. Upon manual inspection, we could identify some samples for which we did not agree with their labels. Motivated by this observation, we started looking into understanding the quality of datasets used in toxic text prediction~\citep{arhin2021ground}. As a consequence, while we expect the trends shown in this paper to hold, the actual absolute numbers may vary with datasets/tasks.
More observations and limitations can be found in Section~\ref{section:ethics}.

\section{Conclusions}
In this work, we addressed the following research questions for language models: how do model size, training size, random seeds affect the relationship between performance and fairness (as measured by equalized odds)? Can post-processing methods for bias mitigation lead to better operating points for both accuracy and fairness? We find these questions important in the context of the ethics of using language models in text toxicity prediction, in particular, and in NLP research, in general.
We presented a comprehensive study of language models and their performance/fairness relationship. We chose several models to cover different sizes and different architectures. While we did not consider some of the largest recent models available, we believe we have experimented with a wide variety of models that have been discussed well in the literature. 
%Using A100 GPUs, we were able to finish fine-tuning for our largest models in at most 24 hours.
We hope that this study can drive the following point across: 
we cannot make a blanket statement on the fairness of language models with respect to their size or architecture, while training factors such as data size and random seeds can make a large difference. This makes it all the more important for researchers/practitioners to make fairness an integral part of the performance evaluation of language models. 
%before deployment.

\section{Ethics Statement}
\label{section:ethics}
 
This research used a considerable amount of computational resources and this is our main ethics concern for conducting this work. We did try to keep the number and the size of models we experimented with limited, to reduce the carbon footprint of the experiments. We hope the results we show in this paper are worth the computational resources used. 

In this study, we looked at coarse-grained groups defined by the text content mentioning religion/race/gender, which may obfuscate the behavior of the models with respect to finer-grained groups, such as females and males. Similarly, we did not consider intersectionality. 

Bias mitigation can lead to undesirable outcomes. For example, one aspect we did not look into is what happens with other groups when the mitigation is applied only for one of the groups. In addition, we focused only on group fairness and do not provide any insights into individual fairness. We also recognize that abstract metrics have limitations and the societal impacts resulting from bias mitigation are not well understood~\citep{olteanu2017limits}. These issues are universal to bias mitigation techniques and not particular to our use case.

Last, but not least, the datasets we used are English only. We acknowledge the importance of performing similar studies on multi-lingual datasets.

\bibliography{fairnlp}

\begin{thebibliography}{71}
\expandafter\ifx\csname natexlab\endcsname\relax\def\natexlab#1{#1}\fi

\bibitem[{Abbasi et~al.(2019)Abbasi, Friedler, Scheidegger, and
  Venkatasubramanian}]{abbasi2019fairness}
Mohsen Abbasi, Sorelle~A Friedler, Carlos Scheidegger, and Suresh
  Venkatasubramanian. 2019.
\newblock Fairness in representation: quantifying stereotyping as a
  representational harm.
\newblock In \emph{Proceedings of the 2019 SIAM International Conference on
  Data Mining}.

\bibitem[{AI2(2021)}]{AI2Leaderboards}
Allen Institute for~AI AI2. 2021.
\newblock \href {https://leaderboard.allenai.org/} {Leaderboards}.

\bibitem[{Angwin et~al.(2017)Angwin, Larson, Kirchner, and
  Mattu}]{angwin2017Minority}
Julia Angwin, Jeff Larson, Lauren Kirchner, and Surya Mattu. 2017.
\newblock Minority {{Neighborhoods Pay Higher Car Insurance Premiums Than White
  Areas With}} the {{Same Risk}}.
\newblock
  https://www.propublica.org/article/minority-neighborhoods-higher-car-insurance-premiums-white-areas-same-risk.

\bibitem[{Angwin et~al.(2016)Angwin, Larson, Mattu, and
  Kirchner}]{angwin2016Machine}
Julia Angwin, Jeff Larson, Surya Mattu, and Lauren Kirchner. 2016.
\newblock Machine {{Bias}}.
\newblock
  www.propublica.org/article/machine-bias-risk-assessments-in-criminal-sentencing.

\bibitem[{Arhin et~al.(2021)Arhin, Baldini, Wei, Ramamurthy, and
  Singh}]{arhin2021ground}
Kofi Arhin, Ioana Baldini, Dennis Wei, Karthikeyan~Natesan Ramamurthy, and
  Moninder Singh. 2021.
\newblock Ground-truth, whose truth? examing the difficulties with annotating
  toxic text datasets.
\newblock In \emph{Data-Centric AI Workshop colocated with NeurIPS 2021}.

\bibitem[{Awasthi et~al.(2020)Awasthi, Kleindessner, and
  Morgenstern}]{Awasthi2020Equalized}
Pranjal Awasthi, Matth{\"{a}}us Kleindessner, and Jamie Morgenstern. 2020.
\newblock \href {http://proceedings.mlr.press/v108/awasthi20a.html} {Equalized
  odds postprocessing under imperfect group information}.
\newblock In \emph{The 23rd International Conference on Artificial Intelligence
  and Statistics, {AISTATS} 2020}.

\bibitem[{Ball-Burack et~al.(2021)Ball-Burack, Lee, Cobbe, and
  Singh}]{ball2021differential}
Ari Ball-Burack, Michelle Seng~Ah Lee, Jennifer Cobbe, and Jatinder Singh.
  2021.
\newblock Differential tweetment: Mitigating racial dialect bias in harmful
  tweet detection.
\newblock In \emph{Proceedings of the 2021 ACM Conference on Fairness,
  Accountability, and Transparency}.

\bibitem[{Bellamy et~al.(2019)Bellamy, Dey, Hind, Hoffman, Houde, Kannan,
  Lohia, Martino, Mehta, Mojsilovi{\'c} et~al.}]{bellamy2019ai}
Rachel~KE Bellamy, Kuntal Dey, Michael Hind, Samuel~C Hoffman, Stephanie Houde,
  Kalapriya Kannan, Pranay Lohia, Jacquelyn Martino, Sameep Mehta, Aleksandra
  Mojsilovi{\'c}, et~al. 2019.
\newblock {AI Fairness 360}: An extensible toolkit for detecting and mitigating
  algorithmic bias.
\newblock \emph{IBM Journal of Research and Development}.

\bibitem[{Bender et~al.(2021)Bender, Gebru, McMillan-Major, and
  Shmitchell}]{Bender2021Dangers}
Emily~M. Bender, Timnit Gebru, Angelina McMillan-Major, and Shmargaret
  Shmitchell. 2021.
\newblock \href {https://doi.org/10.1145/3442188.3445922} {On the dangers of
  stochastic parrots: Can language models be too
  big?\raisebox{-5pt}{\includegraphics[scale=0.1]{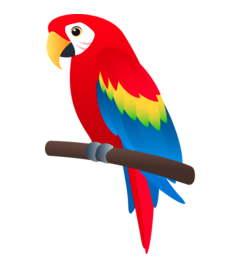}}}.
\newblock In \emph{Proceedings of the 2021 ACM Conference on Fairness,
  Accountability, and Transparency}.

\bibitem[{Blodgett et~al.(2020)Blodgett, Barocas, Daum{\'e}~III, and
  Wallach}]{Blodgett2020Language}
Su~Lin Blodgett, Solon Barocas, Hal Daum{\'e}~III, and Hanna Wallach. 2020.
\newblock \href {https://doi.org/10.18653/v1/2020.acl-main.485} {Language
  (technology) is power: A critical survey of {``}bias{''} in {NLP}}.
\newblock In \emph{Proceedings of the 58th Annual Meeting of the Association
  for Computational Linguistics}.

\bibitem[{Bogdanoff(2017)}]{bogdanoff2017saying}
Aja Bogdanoff. 2017.
\newblock {Saying goodbye to Civil Comments}.
\newblock
  [\href{"https://medium.com/@aja_15265/saying-goodbye-to-civil-comments-41859d3a2b1d"}{Online};
  accessed 21-July-2021].

\bibitem[{Bolukbasi et~al.(2016)Bolukbasi, Chang, Zou, Saligrama, and
  Kalai}]{Bolukbasi2016Man}
Tolga Bolukbasi, Kai-Wei Chang, James Zou, Venkatesh Saligrama, and Adam Kalai.
  2016.
\newblock Man is to computer programmer as woman is to homemaker? debiasing
  word embeddings.
\newblock In \emph{Proceedings of the 30th International Conference on Neural
  Information Processing Systems}.

\bibitem[{Borkan et~al.(2019)Borkan, Dixon, Sorensen, Thain, and
  Vasserman}]{Borkan2019Nuanced}
Daniel Borkan, Lucas Dixon, Jeffrey Sorensen, Nithum Thain, and Lucy Vasserman.
  2019.
\newblock \href {https://doi.org/10.1145/3308560.3317593} {Nuanced metrics for
  measuring unintended bias with real data for text classification}.
\newblock In \emph{Companion of The 2019 World Wide Web Conference, {WWW}}.

\bibitem[{Buolamwini and Gebru(2018)}]{buolamwini2018gender}
Joy Buolamwini and Timnit Gebru. 2018.
\newblock Gender shades: Intersectional accuracy disparities in commercial
  gender classification.
\newblock In \emph{Conference on fairness, accountability and transparency}.

\bibitem[{Chouldechova and Roth(2018)}]{chouldechova2018frontiers}
Alexandra Chouldechova and Aaron Roth. 2018.
\newblock The frontiers of fairness in machine learning.
\newblock \emph{arXiv preprint arXiv:1810.08810}.

\bibitem[{Chzhen et~al.(2019)Chzhen, Denis, Hebiri, Oneto, and
  Pontil}]{chzhen2019leveraging}
Evgenii Chzhen, Christophe Denis, Mohamed Hebiri, Luca Oneto, and Massimiliano
  Pontil. 2019.
\newblock Leveraging labeled and unlabeled data for consistent fair binary
  classification.
\newblock \emph{Advances in Neural Information Processing Systems}, 32.

\bibitem[{Clark et~al.(2020)Clark, Luong, Le, and Manning}]{Clark2020Electra}
Kevin Clark, Minh{-}Thang Luong, Quoc~V. Le, and Christopher~D. Manning. 2020.
\newblock \href {https://openreview.net/forum?id=r1xMH1BtvB} {{ELECTRA:}
  pre-training text encoders as discriminators rather than generators}.
\newblock In \emph{8th International Conference on Learning Representations,
  {ICLR} 2020, Addis Ababa, Ethiopia, April 26-30, 2020}. OpenReview.net.

\bibitem[{Crawford(2017)}]{Crawford2013}
Kate Crawford. 2017.
\newblock The trouble with bias.
\newblock \url{https://www.youtube.com/watch?v=fMym_BKWQzk}.

\bibitem[{Czarnowska et~al.(2021)Czarnowska, Vyas, and
  Shah}]{czarnowska2021quantifying}
Paula Czarnowska, Yogarshi Vyas, and Kashif Shah. 2021.
\newblock \href {https://doi.org/10.1162/tacl_a_00425} {Quantifying social
  biases in {NLP}: A generalization and empirical comparison of extrinsic
  fairness metrics}.
\newblock \emph{Transactions of the Association for Computational Linguistics}.

\bibitem[{Dai et~al.(2020)Dai, Lai, Yang, and Le}]{Dai2020Funnel}
Zihang Dai, Guokun Lai, Yiming Yang, and Quoc Le. 2020.
\newblock \href
  {https://proceedings.neurips.cc/paper/2020/hash/2cd2915e69546904e4e5d4a2ac9e1652-Abstract.html}
  {{Funnel-Transformer}: Filtering out sequential redundancy for efficient
  language processing}.
\newblock In \emph{Annual Conference on Neural Information Processing Systems
  2020}.

\bibitem[{D'Amour et~al.(2020)D'Amour, Heller, Moldovan, Adlam, Alipanahi,
  Beutel, Chen, Deaton, Eisenstein, Hoffman, Hormozdiari, Houlsby, Hou, Jerfel,
  Karthikesalingam, Lucic, Ma, McLean, Mincu, Mitani, Montanari, Nado,
  Natarajan, Nielson, Osborne, Raman, Ramasamy, Sayres, Schrouff, Seneviratne,
  Sequeira, Suresh, Veitch, Vladymyrov, Wang, Webster, Yadlowsky, Yun, Zhai,
  and Sculley}]{Amour2020Underspecification}
Alexander D'Amour, Katherine~A. Heller, Dan Moldovan, Ben Adlam, Babak
  Alipanahi, Alex Beutel, Christina Chen, Jonathan Deaton, Jacob Eisenstein,
  Matthew~D. Hoffman, Farhad Hormozdiari, Neil Houlsby, Shaobo Hou, Ghassen
  Jerfel, Alan Karthikesalingam, Mario Lucic, Yi{-}An Ma, Cory~Y. McLean, Diana
  Mincu, Akinori Mitani, Andrea Montanari, Zachary Nado, Vivek Natarajan,
  Christopher Nielson, Thomas~F. Osborne, Rajiv Raman, Kim Ramasamy, Rory
  Sayres, Jessica Schrouff, Martin Seneviratne, Shannon Sequeira, Harini
  Suresh, Victor Veitch, Max Vladymyrov, Xuezhi Wang, Kellie Webster, Steve
  Yadlowsky, Taedong Yun, Xiaohua Zhai, and D.~Sculley. 2020.
\newblock \href {http://arxiv.org/abs/2011.03395} {Underspecification presents
  challenges for credibility in modern machine learning}.
\newblock \emph{CoRR}, abs/2011.03395.

\bibitem[{de~Vassimon~Manela et~al.(2021)de~Vassimon~Manela, Errington, Fisher,
  van Breugel, and Minervini}]{DeVassimon2021Stereotype}
Daniel de~Vassimon~Manela, David Errington, Thomas Fisher, Boris van Breugel,
  and Pasquale Minervini. 2021.
\newblock \href {https://aclanthology.org/2021.eacl-main.190} {Stereotype and
  skew: Quantifying gender bias in pre-trained and fine-tuned language models}.
\newblock In \emph{Proceedings of the 16th Conference of the European Chapter
  of the Association for Computational Linguistics}. Association for
  Computational Linguistics.

\bibitem[{Devlin et~al.(2019)Devlin, Chang, Lee, and
  Toutanova}]{Devlin2019BERT}
J.~Devlin, Ming-Wei Chang, Kenton Lee, and Kristina Toutanova. 2019.
\newblock Bert: Pre-training of deep bidirectional transformers for language
  understanding.
\newblock In \emph{NAACL-HLT}.

\bibitem[{Dodge et~al.(2020)Dodge, Ilharco, Schwartz, Farhadi, Hajishirzi, and
  Smith}]{dodge2020Finetuning}
Jesse Dodge, Gabriel Ilharco, Roy Schwartz, Ali Farhadi, Hannaneh Hajishirzi,
  and Noah~A. Smith. 2020.
\newblock \href {http://arxiv.org/abs/2002.06305} {Fine-tuning pretrained
  language models: Weight initializations, data orders, and early stopping}.
\newblock \emph{CoRR}, abs/2002.06305.

\bibitem[{Dwork et~al.(2012)Dwork, Hardt, Pitassi, Reingold, and
  Zemel}]{dwork2012fairness}
Cynthia Dwork, Moritz Hardt, Toniann Pitassi, Omer Reingold, and Richard Zemel.
  2012.
\newblock Fairness through awareness.
\newblock In \emph{Proceedings of the 3rd innovations in theoretical computer
  science conference}.

\bibitem[{Fish et~al.(2016)Fish, Kun, and Lelkes}]{fish2016confidence}
Benjamin Fish, Jeremy Kun, and {\'A}d{\'a}m~D Lelkes. 2016.
\newblock A confidence-based approach for balancing fairness and accuracy.
\newblock In \emph{Proceedings of the 2016 SIAM International Conference on
  Data Mining}. SIAM.

\bibitem[{Goldfarb{-}Tarrant et~al.(2021)Goldfarb{-}Tarrant, Marchant, Sanchez,
  Pandya, and Lopez}]{GoldfarbTarrant2021Intrinsic}
Seraphina Goldfarb{-}Tarrant, Rebecca Marchant, Ricardo~Mu{\~{n}}oz Sanchez,
  Mugdha Pandya, and Adam Lopez. 2021.
\newblock \href {https://arxiv.org/abs/2012.15859} {Intrinsic bias metrics do
  not correlate with application bias}.
\newblock In \emph{Proceedings of the 59th Annual Meeting of the Association
  for Computational Linguistics}. Association for Computational Linguistics.

\bibitem[{Guo et~al.(2017)Guo, Pleiss, Sun, and
  Weinberger}]{guo2017calibration}
Chuan Guo, Geoff Pleiss, Yu~Sun, and Kilian~Q. Weinberger. 2017.
\newblock \href {http://proceedings.mlr.press/v70/guo17a.html} {On calibration
  of modern neural networks}.
\newblock In \emph{Proceedings of the 34th International Conference on Machine
  Learning, {ICML} 2017, Sydney, NSW, Australia, 6-11 August 2017}.

\bibitem[{Hardt et~al.(2016)Hardt, Price, and Srebro}]{hardt2016equality}
Moritz Hardt, Eric Price, and Nati Srebro. 2016.
\newblock Equality of opportunity in supervised learning.
\newblock \emph{Advances in neural information processing systems}.

\bibitem[{He et~al.(2021)He, Liu, Gao, and Chen}]{He2021Deberta}
Pengcheng He, Xiaodong Liu, Jianfeng Gao, and Weizhu Chen. 2021.
\newblock \href {https://openreview.net/forum?id=XPZIaotutsD} {{DeBERTa}:
  decoding-enhanced {BERT} with disentangled attention}.
\newblock In \emph{9th International Conference on Learning Representations,
  {ICLR} 2021, Virtual Event, Austria, May 3-7, 2021}. OpenReview.net.

\bibitem[{Hooker et~al.(2020)Hooker, Moorosi, Clark, Bengio, and
  Denton}]{Hooker2020Characterising}
Sara Hooker, Nyalleng Moorosi, Gregory Clark, S.~Bengio, and Emily~L. Denton.
  2020.
\newblock \href {https://arxiv.org/abs/2010.03058} {Characterising bias in
  compressed models}.
\newblock \emph{ArXiv}.

\bibitem[{Hutchinson et~al.(2020)Hutchinson, Prabhakaran, Denton, Webster,
  Zhong, and Denuyl}]{hutchinson2020social}
Ben Hutchinson, Vinodkumar Prabhakaran, Emily Denton, Kellie Webster, Yu~Zhong,
  and Stephen Denuyl. 2020.
\newblock \href {https://doi.org/10.18653/v1/2020.acl-main.487} {Social biases
  in {NLP} models as barriers for persons with disabilities}.
\newblock In \emph{Proceedings of the 58th Annual Meeting of the Association
  for Computational Linguistics}. Association for Computational Linguistics.

\bibitem[{Iandola et~al.(2020)Iandola, Shaw, Krishna, and
  Keutzer}]{iandola2020squeezebert}
Forrest Iandola, Albert Shaw, Ravi Krishna, and Kurt Keutzer. 2020.
\newblock \href {https://doi.org/10.18653/v1/2020.sustainlp-1.17}
  {{S}queeze{BERT}: What can computer vision teach {NLP} about efficient neural
  networks?}
\newblock In \emph{Proceedings of SustaiNLP: Workshop on Simple and Efficient
  Natural Language Processing}.

\bibitem[{Jiang et~al.(2020)Jiang, Pacchiano, Stepleton, Jiang, and
  Chiappa}]{jiang2020wasserstein}
Ray Jiang, Aldo Pacchiano, Tom Stepleton, Heinrich Jiang, and Silvia Chiappa.
  2020.
\newblock Wasserstein fair classification.
\newblock In \emph{Uncertainty in Artificial Intelligence}.

\bibitem[{Jigsaw(2019)}]{kaggle2019jigsaw}
Kaggle Jigsaw. 2019.
\newblock {Jigsaw Unintended Bias in Toxicity Classification}.
\newblock
  [\href{https://www.kaggle.com/c/jigsaw-unintended-bias-in-toxicity-classification}{Online};
  accessed 21-July-2021].

\bibitem[{Kamiran et~al.(2012)Kamiran, Karim, and Zhang}]{kamiran2012decision}
Faisal Kamiran, Asim Karim, and Xiangliang Zhang. 2012.
\newblock Decision theory for discrimination-aware classification.
\newblock In \emph{2012 IEEE 12th International Conference on Data Mining}.
  IEEE.

\bibitem[{Kim et~al.(2019)Kim, Ghorbani, and Zou}]{kim2019multiaccuracy}
Michael~P Kim, Amirata Ghorbani, and James Zou. 2019.
\newblock Multiaccuracy: Black-box post-processing for fairness in
  classification.
\newblock In \emph{Proceedings of the 2019 AAAI/ACM Conference on AI, Ethics,
  and Society}.

\bibitem[{Kiritchenko et~al.(2021)Kiritchenko, Nejadgholi, and
  Fraser}]{kiritchenko2021confronting}
Svetlana Kiritchenko, Isar Nejadgholi, and Kathleen~C. Fraser. 2021.
\newblock Confronting abusive language online: A survey from the ethical and
  human rights perspective.
\newblock \emph{Journal of Artificial Intelligence Research}.

\bibitem[{Lan et~al.(2020)Lan, Chen, Goodman, Gimpel, Sharma, and
  Soricut}]{Lan2020Albert}
Zhenzhong Lan, Mingda Chen, Sebastian Goodman, Kevin Gimpel, Piyush Sharma, and
  Radu Soricut. 2020.
\newblock \href {https://openreview.net/forum?id=H1eA7AEtvS} {{ALBERT:} {A}
  lite {BERT} for self-supervised learning of language representations}.
\newblock In \emph{8th International Conference on Learning Representations,
  {ICLR} 2020, Addis Ababa, Ethiopia, April 26-30, 2020}. OpenReview.net.

\bibitem[{Lehman et~al.(2019)Lehman, DeYoung, Barzilay, and
  Wallace}]{lehman2019inferring}
Eric Lehman, Jay DeYoung, Regina Barzilay, and Byron~C Wallace. 2019.
\newblock Inferring which medical treatments work from reports of clinical
  trials.
\newblock In \emph{Proceedings of the North American Chapter of the Association
  for Computational Linguistics (NAACL)}.

\bibitem[{Liu et~al.(2019)Liu, Ott, Goyal, Du, Joshi, Chen, Levy, Lewis,
  Zettlemoyer, and Stoyanov}]{liu2019roberta}
Yinhan Liu, Myle Ott, Naman Goyal, Jingfei Du, Mandar Joshi, Danqi Chen, Omer
  Levy, Mike Lewis, Luke Zettlemoyer, and Veselin Stoyanov. 2019.
\newblock \href {http://arxiv.org/abs/1907.11692} {Roberta: A robustly
  optimized bert pretraining approach}.

\bibitem[{Mathew et~al.(2021)Mathew, Saha, Yimam, Biemann, Goyal, and
  Mukherjee}]{Mathew2021Hatexplain}
Binny Mathew, Punyajoy Saha, Seid~Muhie Yimam, Chris Biemann, Pawan Goyal, and
  Animesh Mukherjee. 2021.
\newblock \href {https://ojs.aaai.org/index.php/AAAI/article/view/17745}
  {{HateXplain}: {A} benchmark dataset for explainable hate speech detection}.
\newblock In \emph{Thirty-Fifth {AAAI} Conference on Artificial Intelligence,
  {AAAI} 2021}.

\bibitem[{McCoy et~al.(2019)McCoy, Pavlick, and Linzen}]{mccoy2019right}
Tom McCoy, Ellie Pavlick, and Tal Linzen. 2019.
\newblock \href {https://doi.org/10.18653/v1/P19-1334} {Right for the wrong
  reasons: Diagnosing syntactic heuristics in natural language inference}.
\newblock In \emph{Proceedings of the 57th Annual Meeting of the Association
  for Computational Linguistics}, Florence, Italy. Association for
  Computational Linguistics.

\bibitem[{Nayak(2019)}]{Nayak2019Understanding}
Pandu Nayak. 2019.
\newblock \href
  {https://blog.google/products/search/search-language-understanding-bert/}
  {Understanding searches better than ever before}.

\bibitem[{Nguyen et~al.(2020)Nguyen, Vu, and Nguyen}]{Nguyen2020BERTweet}
Dat~Quoc Nguyen, Thanh Vu, and Anh~Tuan Nguyen. 2020.
\newblock \href {https://doi.org/10.18653/v1/2020.emnlp-demos.2} {{BERTweet}:
  {A} pre-trained language model for english tweets}.
\newblock In \emph{Proceedings of the 2020 Conference on Empirical Methods in
  Natural Language Processing: System Demonstrations, {EMNLP} 2020 - Demos}.

\bibitem[{Olteanu et~al.(2017)Olteanu, Talamadupula, and
  Varshney}]{olteanu2017limits}
Alexandra Olteanu, Kartik Talamadupula, and Kush~R. Varshney. 2017.
\newblock \href {https://doi.org/10.1145/3091478.3098871} {The limits of
  abstract evaluation metrics: The case of hate speech detection}.
\newblock In \emph{Proceedings of the 2017 {ACM} on Web Science Conference,
  WebSci 2017, Troy, NY, USA, June 25 - 28, 2017}.

\bibitem[{Park et~al.(2021)Park, Hu, Singh, Sylla, Dankwa-Mullan, Koski, and
  Das}]{park2021comparison}
Yoonyoung Park, Jianying Hu, Moninder Singh, Issa Sylla, Irene Dankwa-Mullan,
  Eileen Koski, and Amar~K. Das. 2021.
\newblock \href {https://doi.org/10.1001/jamanetworkopen.2021.3909}
  {{Comparison of Methods to Reduce Bias From Clinical Prediction Models of
  Postpartum Depression}}.
\newblock \emph{JAMA Network Open}.

\bibitem[{Paszke et~al.(2019)Paszke, Gross, Massa, Lerer, Bradbury, Chanan,
  Killeen, Lin, Gimelshein, Antiga, Desmaison, Kopf, Yang, DeVito, Raison,
  Tejani, Chilamkurthy, Steiner, Fang, Bai, and Chintala}]{NEURIPS2019_9015}
Adam Paszke, Sam Gross, Francisco Massa, Adam Lerer, James Bradbury, Gregory
  Chanan, Trevor Killeen, Zeming Lin, Natalia Gimelshein, Luca Antiga, Alban
  Desmaison, Andreas Kopf, Edward Yang, Zachary DeVito, Martin Raison, Alykhan
  Tejani, Sasank Chilamkurthy, Benoit Steiner, Lu~Fang, Junjie Bai, and Soumith
  Chintala. 2019.
\newblock \href
  {http://papers.neurips.cc/paper/9015-pytorch-an-imperative-style-high-performance-deep-learning-library.pdf}
  {Pytorch: An imperative style, high-performance deep learning library}.
\newblock In \emph{Advances in Neural Information Processing Systems 32}.

\bibitem[{Patterson et~al.(2021)Patterson, Gonzalez, Le, Liang, Munguia,
  Rothchild, So, Texier, and Dean}]{patterson2021carbon}
David Patterson, Joseph Gonzalez, Quoc Le, Chen Liang, Lluis-Miquel Munguia,
  Daniel Rothchild, David So, Maud Texier, and Jeff Dean. 2021.
\newblock Carbon emissions and large neural network training.
\newblock \emph{arXiv preprint arXiv:2104.10350}.

\bibitem[{{Perspective API}(2021)}]{Perspective2021}
{Perspective API}. 2021.
\newblock {Using Machine Learning to Reduce Toxicity Online}.
\newblock [\href{https://perspectiveapi.com/how-it-works/}{Online}; accessed
  21-July-2021].

\bibitem[{Pleiss et~al.(2017)Pleiss, Raghavan, Wu, Kleinberg, and
  Weinberger}]{pleiss2017fairness}
Geoff Pleiss, Manish Raghavan, Felix Wu, Jon Kleinberg, and Kilian~Q
  Weinberger. 2017.
\newblock On fairness and calibration.
\newblock \emph{arXiv preprint arXiv:1709.02012}.

\bibitem[{Radford et~al.(2019)Radford, Wu, Child, Luan, Amodei, and
  Sutskever}]{Radford2019Language}
Alec Radford, Jeffrey Wu, Rewon Child, David Luan, Dario Amodei, and Ilya
  Sutskever. 2019.
\newblock \href {https://openai.com/blog/better-language-models/} {{Language
  Models are Unsupervised Multitask Learners}}.

\bibitem[{Rajpurkar et~al.(2018)Rajpurkar, Jia, and Liang}]{Rajpurkar2018Know}
Pranav Rajpurkar, Robin Jia, and Percy Liang. 2018.
\newblock \href {https://doi.org/10.18653/v1/P18-2124} {Know what you don{'}t
  know: Unanswerable questions for {SQ}u{AD}}.
\newblock In \emph{Proceedings of the 56th Annual Meeting of the Association
  for Computational Linguistics (Volume 2: Short Papers)}, Melbourne,
  Australia. Association for Computational Linguistics.

\bibitem[{Rogers et~al.(2021)Rogers, Kovaleva, and
  Rumshisky}]{Rogers2021Primer}
Anna Rogers, Olga Kovaleva, and Anna Rumshisky. 2021.
\newblock A primer in bertology: What we know about how bert works.
\newblock \emph{Transactions of the Association for Computational Linguistics}.

\bibitem[{Sanh et~al.(2020)Sanh, Debut, Chaumond, and
  Wolf}]{sanh2020distilbert}
Victor Sanh, Lysandre Debut, Julien Chaumond, and Thomas Wolf. 2020.
\newblock \href {http://arxiv.org/abs/1910.01108} {Distilbert, a distilled
  version of bert: smaller, faster, cheaper and lighter}.

\bibitem[{Schwaller et~al.(2021)Schwaller, Probst, Vaucher, Nair, Kreutter,
  Laino, and Reymond}]{Schwaller2021Mapping}
Philippe Schwaller, Daniel Probst, Alain~C. Vaucher, Vishnu~H. Nair, David
  Kreutter, Teodoro Laino, and Jean-Louis Reymond. 2021.
\newblock \href {https://doi.org/10.1038/s42256-020-00284-w} {Mapping the space
  of chemical reactions using attention-based neural networks}.
\newblock \emph{Nature Machine Intelligence}.

\bibitem[{Solaiman and Dennison(2021)}]{Solaiman2021Process}
Irene Solaiman and Christy Dennison. 2021.
\newblock \href {https://arxiv.org/abs/2106.10328} {Process for adapting
  language models to society {(PALMS)} with values-targeted datasets}.
\newblock In \emph{Annual Conference on Neural Information Processing Systems}.

\bibitem[{Strubell et~al.(2019)Strubell, Ganesh, and
  McCallum}]{Strubell2019Energy}
Emma Strubell, Ananya Ganesh, and Andrew McCallum. 2019.
\newblock \href {https://doi.org/10.18653/v1/p19-1355} {Energy and policy
  considerations for deep learning in {NLP}}.
\newblock In \emph{Proceedings of the 57th Conference of the Association for
  Computational Linguistics, {ACL}}.

\bibitem[{Subramanian et~al.(2020)Subramanian, Baldini, Ravichandran,
  Katz-Rogozhnikov, Ramamurthy, Sattigeri, Kush~R, Wang, Mangalath, and
  Kleiman}]{Subramanian2020Natural}
Shivashankar Subramanian, Ioana Baldini, Sushma Ravichandran, Dmitriy
  Katz-Rogozhnikov, Karthikeyan~Natesan Ramamurthy, Prasanna Sattigeri,
  Varshney Kush~R, Annmarie Wang, Pradeep Mangalath, and Laura Kleiman. 2020.
\newblock A natural language processing system for extracting evidence of drug
  repurposing from scientific publications.
\newblock \emph{Proceedings of the AAAI Conference on Innovative Applications
  of Artificial Intelligence}.

\bibitem[{Sun et~al.(2019)Sun, Gaut, Tang, Huang, ElSherief, Zhao, Mirza,
  Belding, Chang, and Wang}]{sun2019mitigating}
Tony Sun, Andrew Gaut, Shirlyn Tang, Yuxin Huang, Mai ElSherief, Jieyu Zhao,
  Diba Mirza, Elizabeth Belding, Kai-Wei Chang, and William~Yang Wang. 2019.
\newblock Mitigating gender bias in natural language processing: Literature
  review.
\newblock In \emph{Proceedings of the 57th Annual Meeting of the Association
  for Computational Linguistics}.

\bibitem[{Sun et~al.(2020)Sun, Yu, Song, Liu, Yang, and
  Zhou}]{Sun2020MobileBERT}
Zhiqing Sun, Hongkun Yu, Xiaodan Song, Renjie Liu, Yiming Yang, and Denny Zhou.
  2020.
\newblock \href {https://doi.org/10.18653/v1/2020.acl-main.195} {{MobileBERT}:
  a compact task-agnostic {BERT} for resource-limited devices}.
\newblock In \emph{Proceedings of the 58th Annual Meeting of the Association
  for Computational Linguistics, {ACL} 2020}.

\bibitem[{Verma and Rubin(2018)}]{Verma2018Fairness}
Sahil Verma and Julia Rubin. 2018.
\newblock \href {https://doi.org/10.1145/3194770.3194776} {Fairness definitions
  explained}.
\newblock In \emph{Proceedings of the International Workshop on Software
  Fairness}, New York, NY, USA. Association for Computing Machinery.

\bibitem[{Vogels(2021)}]{vogels2021state}
Emily~A. Vogels. 2021.
\newblock {The State of Online Harassment}.
\newblock
  [\href{https://www.pewresearch.org/internet/2021/01/13/the-state-of-online-harassment/}{Online};
  accessed 21-July-2021].

\bibitem[{Wang et~al.(2019{\natexlab{a}})Wang, Pruksachatkun, Nangia, Singh,
  Michael, Hill, Levy, and Bowman}]{wang2019superglue}
Alex Wang, Yada Pruksachatkun, Nikita Nangia, Amanpreet Singh, Julian Michael,
  Felix Hill, Omer Levy, and Samuel~R. Bowman. 2019{\natexlab{a}}.
\newblock Super{GLUE}: A stickier benchmark for general-purpose language
  understanding systems.
\newblock \emph{arXiv preprint 1905.00537}.

\bibitem[{Wang et~al.(2019{\natexlab{b}})Wang, Singh, Michael, Hill, Levy, and
  Bowman}]{wang2019glue}
Alex Wang, Amanpreet Singh, Julian Michael, Felix Hill, Omer Levy, and
  Samuel~R. Bowman. 2019{\natexlab{b}}.
\newblock {GLUE}: A multi-task benchmark and analysis platform for natural
  language understanding.
\newblock In \emph{Proceedings of ICLR.}

\bibitem[{Webster et~al.(2020)Webster, Wang, Tenney, Beutel, Pitler, Pavlick,
  Chen, and Petrov}]{Webster2020Measuring}
Kellie Webster, Xuezhi Wang, Ian Tenney, Alex Beutel, Emily Pitler, Ellie
  Pavlick, Jilin Chen, and Slav Petrov. 2020.
\newblock \href {http://arxiv.org/abs/2010.06032} {Measuring and reducing
  gendered correlations in pre-trained models}.
\newblock \emph{CoRR}, abs/2010.06032.

\bibitem[{Wei et~al.(2021)Wei, Ramamurthy, and Calmon}]{Wei2021Optimized}
Dennis Wei, Karthikeyan~Natesan Ramamurthy, and Flavio~P. Calmon. 2021.
\newblock \href {http://jmlr.org/papers/v22/20-1143.html} {Optimized score
  transformation for consistent fair classification}.
\newblock \emph{Journal of Machine Learning Research}.

\bibitem[{Wei et~al.(2020)Wei, Ramamurthy, and
  du~Pin~Calmon}]{Wei2020Optimized}
Dennis Wei, Karthikeyan~Natesan Ramamurthy, and Fl{\'{a}}vio du~Pin~Calmon.
  2020.
\newblock \href {http://proceedings.mlr.press/v108/wei20a.html} {Optimized
  score transformation for fair classification}.
\newblock In \emph{The 23rd International Conference on Artificial Intelligence
  and Statistics, {AISTATS} 2020, 26-28 August 2020}.

\bibitem[{Wolf et~al.(2020)Wolf, Debut, Sanh, Chaumond, Delangue, Moi, Cistac,
  Rault, Louf, Funtowicz, Davison, Shleifer, von Platen, Ma, Jernite, Plu, Xu,
  Scao, Gugger, Drame, Lhoest, and Rush}]{wolf2020transformers}
Thomas Wolf, Lysandre Debut, Victor Sanh, Julien Chaumond, Clement Delangue,
  Anthony Moi, Pierric Cistac, Tim Rault, Rémi Louf, Morgan Funtowicz, Joe
  Davison, Sam Shleifer, Patrick von Platen, Clara Ma, Yacine Jernite, Julien
  Plu, Canwen Xu, Teven~Le Scao, Sylvain Gugger, Mariama Drame, Quentin Lhoest,
  and Alexander~M. Rush. 2020.
\newblock \href {https://www.aclweb.org/anthology/2020.emnlp-demos.6}
  {Transformers: State-of-the-art natural language processing}.
\newblock In \emph{Proceedings of the 2020 Conference on Empirical Methods in
  Natural Language Processing: System Demonstrations}. Association for
  Computational Linguistics.

\bibitem[{Woodworth et~al.(2017)Woodworth, Gunasekar, Ohannessian, and
  Srebro}]{woodworth2017learning}
Blake Woodworth, Suriya Gunasekar, Mesrob~I Ohannessian, and Nathan Srebro.
  2017.
\newblock Learning non-discriminatory predictors.
\newblock In \emph{Conference on Learning Theory}, pages 1920--1953. PMLR.

\bibitem[{Yang et~al.(2020)Yang, Cisse, and Koyejo}]{yang2020fairness}
Forest Yang, Mouhamadou Cisse, and Oluwasanmi~O Koyejo. 2020.
\newblock Fairness with overlapping groups; a probabilistic perspective.
\newblock \emph{Advances in Neural Information Processing Systems}, 33.

\end{thebibliography}
\bibliographystyle{acl_natbib}

\appendix
\section{Appendix}

In this appendix, we discuss the datasets we used in our experiments, include additional experimental results and provide more details on post-processing methods for bias mitigation. We conclude with remarks on the reproducibility of this study.

\subsection{Datasets}
\label{appendix:datasets}
\subsubsection{Jigsaw Unintended Bias in Toxicity Classification}

In 2019, Jigsaw released a large dataset as part of the ``Unintended Bias in Toxicity Classification'' Kaggle competition~\citep{kaggle2019jigsaw}. The dataset is a collection of roughly two million samples of text from online discussions~\citep{bogdanoff2017saying}. The samples are rated for toxicity and annotated with attributes for sensitive groups. Table~\ref{table:jigsaw-groups} shows the groups we considered in our analysis and the available fine-grained group annotations. Note that we considered the coarser groups; a sample text belongs to a sensitive (coarse) group if any (fine-grained) annotation for the sample text exists. We used the original training dataset split in a 80/20 ratio for training and development (dev) tuning, respectively. For reporting test results, we used the private test split released on Kaggle. Statistics for the dataset splits are shown in Table~\ref{table:jigsaw-stats}. Each sample in the dataset (see Table~\ref{table:jigsaw-samples} for a few samples from the dataset) has a toxicity score and we consider anything higher than 0.5 to be toxic.

For the Jigsaw dataset, a combination of automation and crowdsourcing was used to ensure that identity (i.e., sensitive group) labels are a reasonable approximation of true identity-related content (see \href{https://www.kaggle.com/c/jigsaw-unintended-bias-in-toxicity-classification/overview/faq}{Jigsaw FAQ}). Not all the dataset was labeled for identity terms. While these labels are imperfect, we do not believe that the degree of imperfection invalidates our study. We note that the problem of protected attribute labels being imperfect is well-accepted and studied~\citep{Awasthi2020Equalized}.

Noisy and incomplete sensitive group labels are another reason why we chose equalized odds as the fairness measure. EO is a valid fairness measure even when there is overlap between the protected groups (e.g., the group labeled ``non-religion'' still has samples mentioning religion). %In fact, the EO measure is valid no matter how the sensitive group is defined. 
To see this, recall that EO requires that the prediction conditioned on the true label be independent of the protected attribute and its violation can be measured by the difference $\lvert \mathbb{E}[\hat{Y}|Y=1,A=1] - \mathbb{E}[\hat{Y}|Y=1] \rvert$ (similarly for $Y=0$). The first term in the difference is measured on a subset of comments ($A = 1$) that contain identity information. This is a good estimate if a sufficient number of samples were annotated, regardless of the potentially missing identity annotations on the remaining samples. The second term does not depend on annotations at all. Thus, the estimate of EO is not affected by the lack of annotations on some of the comments.

\begin{table}[htb]
\caption{The sensitive groups for Jigsaw dataset with their corresponding fine-grained annotations.}
\label{table:jigsaw-groups}
\small
\begin{center}
\begin{tabular}{|p{0.3\linewidth} | p{0.6\linewidth}|}
\hline
Group & Fine-grained annotation\\
\hline
religion &  atheist, buddhist, christian, hindu, jewish, other religion\\
race & white, asian, black, latino, other race or ethnicity\\
gender and sexual orientation$^*$ & bisexual, female, male, heterosexual, homosexual gay or lesbian, transgender, other gender, other sexual orientation\\
\hline
\multicolumn{2}{l}{$^*$Throughout the paper, we use ``gender'' for short.}
\end{tabular}
\end{center}
\end{table}

\begin{table*}[htb]
\caption{Jigsaw dataset samples.}
\label{table:jigsaw-samples}
\small
\begin{center}
\begin{tabular}{|p{0.7\linewidth} | p{0.1\linewidth} | p{0.1\linewidth}|}
\hline
Comment text & Toxicity & Group\\
\hline
The Atwood fable is Donald, is it? My impression of this noise (over Atwood) is that it’s a gimmick by Atwood and her publisher to cash in on the Donald effect. As if we needed slaves in bonnets to remind us that Donald is a jerk (and where was Atwood’s novel when Monica was being pawed over?). A word to defenders of women: don’t spend your political capital on stupid analogies. & Toxic & Gender\\
\hline
I got a question for you, dear, and it is a fair question: We all know what is happening in Syria; where are all the women’s marches over the slaughter in that country?. And, why has Trudeau been silent, like his pal Barry Obama, on taking effective military action against Syria? All you lefties are the same: you have no side vision. & Normal & Gender\\
\hline
\end{tabular}
\end{center}
\end{table*}

\begin{table}[htb]
\caption{Jigsaw dataset statistics: sample counts per dataset split and sensitive group.}
\label{table:jigsaw-stats}
\small
\begin{center}
\begin{tabular}{c c c c c}%{|p{0.7\linewidth} | p{0.1\linewidth} | p{0.1\linewidth}|}
\hline
Split & Total & Religion & Race & Gender\\
\hline
Train & 1443899 & 50748 & 31022 & 70703\\
Dev & 360975 & 12769 & 7999 & 17869 \\
Test & 97320 & 3316 & 1911 & 4367\\
\hline
\end{tabular}
\end{center}
\end{table}

\subsubsection{HateXplain: Toxic text in Twitter and Twitter-like text}

HateXplain~\citep{Mathew2021Hatexplain} was recently introduced with the intent of studying explanations in offensive and hate speech in Twitter and Twitter like data (i.e., \url{gab.com}). For the purposes of our study, we collapse the annotations for offensive and hate speech into one class of toxic text. Similar to the Jigsaw dataset, HateXplain samples have fine-grained annotations for sensitive groups. We use as groups the coarse-level annotations, as we did for the Jigsaw dataset. The groups that we consider are presented in Table~\ref{table:hatex-groups} and a few examples from the dataset are shown in Table~\ref{table:hatex-samples}. Note the text in each sample is represented in the dataset as a list of tokens; in the table, we concatenated them with spaces and this is the way we use them as inputs for the classifiers as well. We used the splits as provided in the dataset; dataset statistics are shown in Table~\ref{table:hatex-stats}.

\begin{table}[htb]
\caption{The sensitive groups for HateXplain dataset with their corresponding fine-grained annotations.}
\small
\label{table:hatex-groups}
\begin{center}
\begin{tabular}{|p{0.3\linewidth} | p{0.45\linewidth}|}
\hline
Group & Fine-grained annotation\\
\hline
race & African, Arab, Asian, Caucasian, Hispanic\\
religion & Islam, Buddhism, Jewish, Hindu, Christian\\
gender & Men, Women\\
\hline
\end{tabular}
\end{center}
\end{table}

\begin{table*}[htb]
\caption{HateXplain dataset samples.}
\label{table:hatex-samples}
\small
\begin{center}
\begin{tabular}{|p{0.7\linewidth} | p{0.08\linewidth} | p{0.08\linewidth}|}
\hline
Text & Toxicity & Group\\
\hline
why is anna acting like she black & Normal & Race\\
\hline
like brazil many are descendants of west africa social constructs labels classifications have ruined a many people who falsely self identify as white or just do not even know abt their native land all b c of them bum ass caucazoids aka caucasians & Toxic & Race\\
\hline
\end{tabular}
\end{center}
\end{table*}

\begin{table}[htb]
\caption{HateXplain dataset statistics: sample counts per dataset split and sensitive group.}
\label{table:hatex-stats}
\small
\begin{center}
\begin{tabular}{c c c c c}
\hline
Split & Total & Religion & Race & Gender\\
\hline
Train & 15383 & 3924 & 5418 & 3102\\
Dev & 1922 & 481 & 672 & 396 \\
Test & 1924 & 468 & 685 & 375\\
\hline
\end{tabular}
\end{center}
\end{table}

\subsection{The influence of random seeds on accuracy and equalized odds}
\label{appendix:perf_plots}
\begin{figure*}[h]
\begin{center}
\begin{tabular}{ccc}
& Jigsaw Dataset &\\
%  \includegraphics[scale=0.175]{img/jigsaw_seeds_accuracy_religion.png}&
%  \hspace{-.4cm}\includegraphics[scale=0.175]{img/jigsaw_seeds_accuracy_race.png}&
%  \hspace{-.4cm}\includegraphics[scale=0.175]{img/jigsaw_seeds_accuracy_gender_and_sexual_orientation_legend.png}

 \includegraphics[scale=0.175]{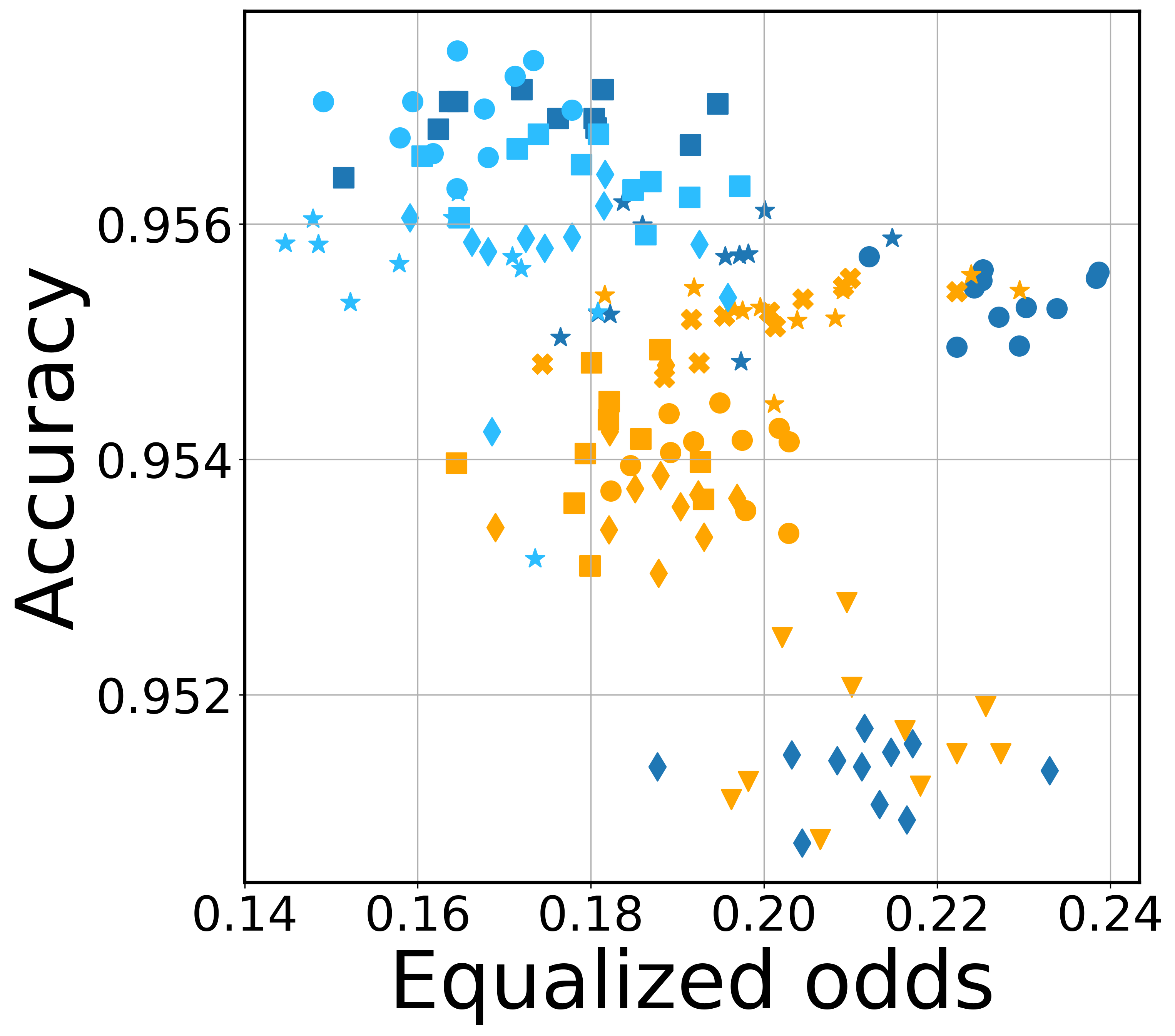}&
 \hspace{-.4cm}\includegraphics[scale=0.175]{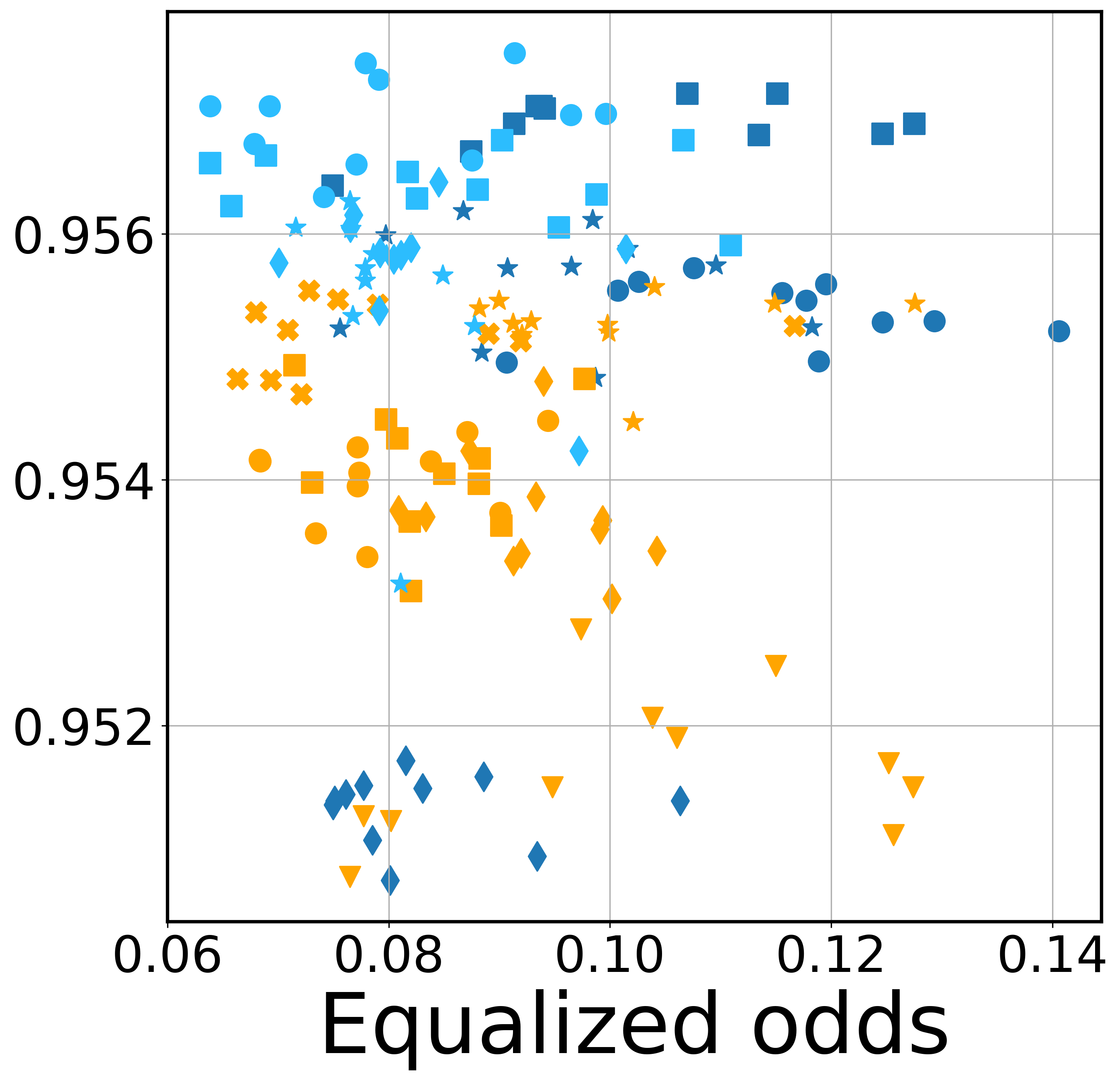}&
 \hspace{-.4cm}\includegraphics[scale=0.175]{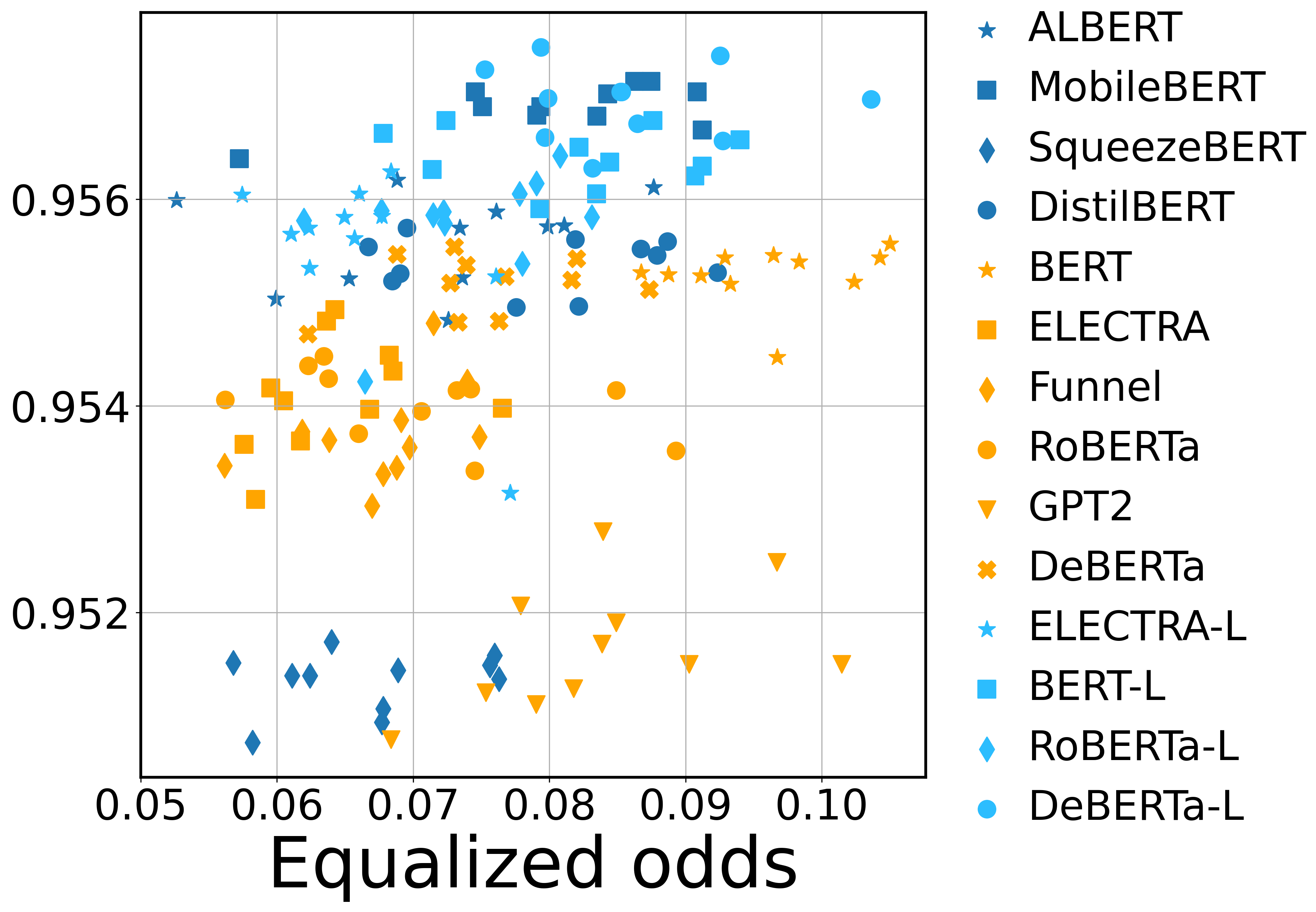}
 \\
 & HateXplain Dataset &\\
%  \includegraphics[scale=0.175]{img/hatex_random_seeds_accuracy_religion.png}&
%  \hspace{-.4cm}\includegraphics[scale=0.175]{img/hatex_random_seeds_accuracy_race.png}&
%  \hspace{-.4cm}\includegraphics[scale=0.175]{img/hatex_random_seeds_accuracy_gender_legend.png}
 \includegraphics[scale=0.175]{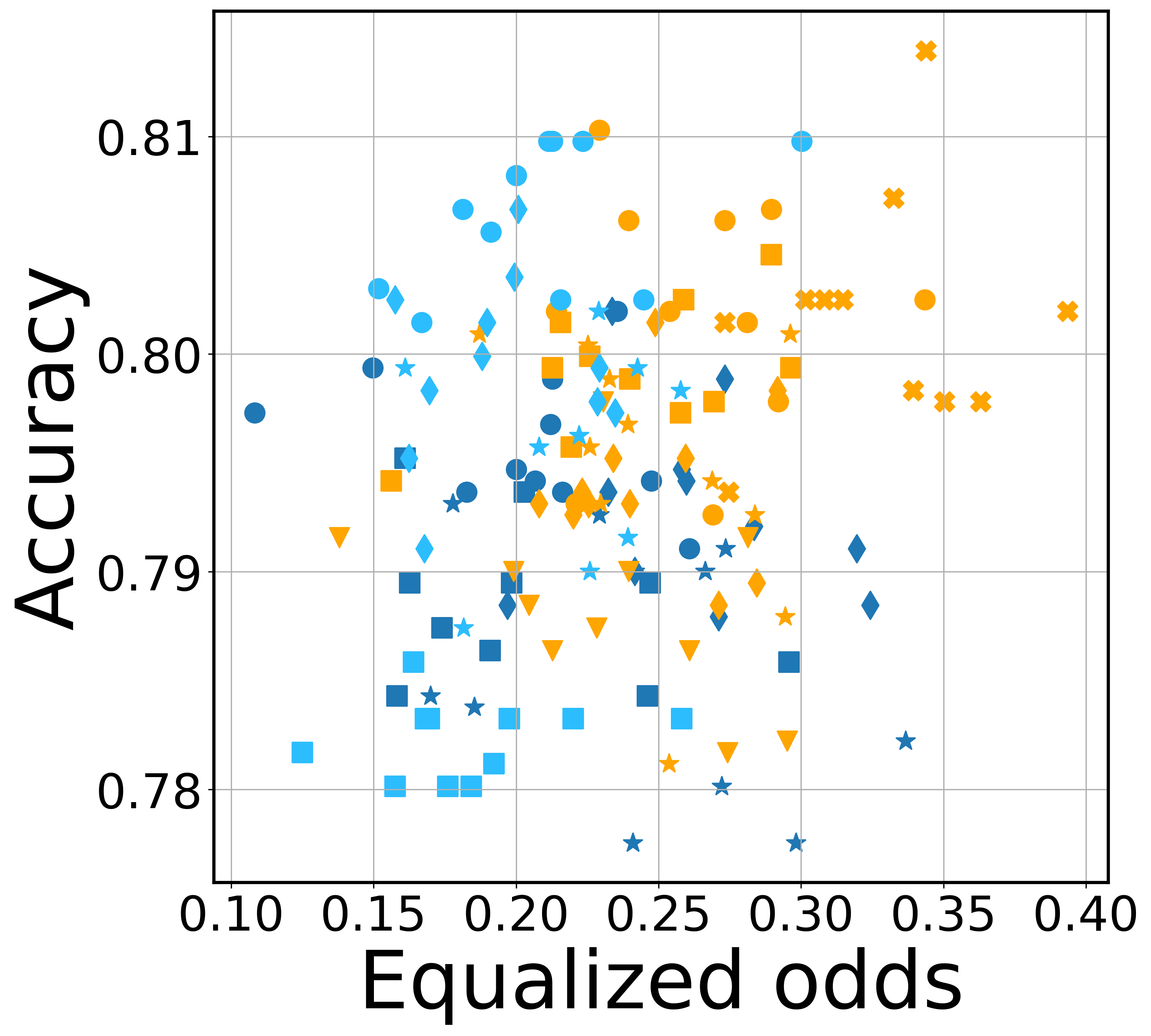}&
 \hspace{-.4cm}\includegraphics[scale=0.175]{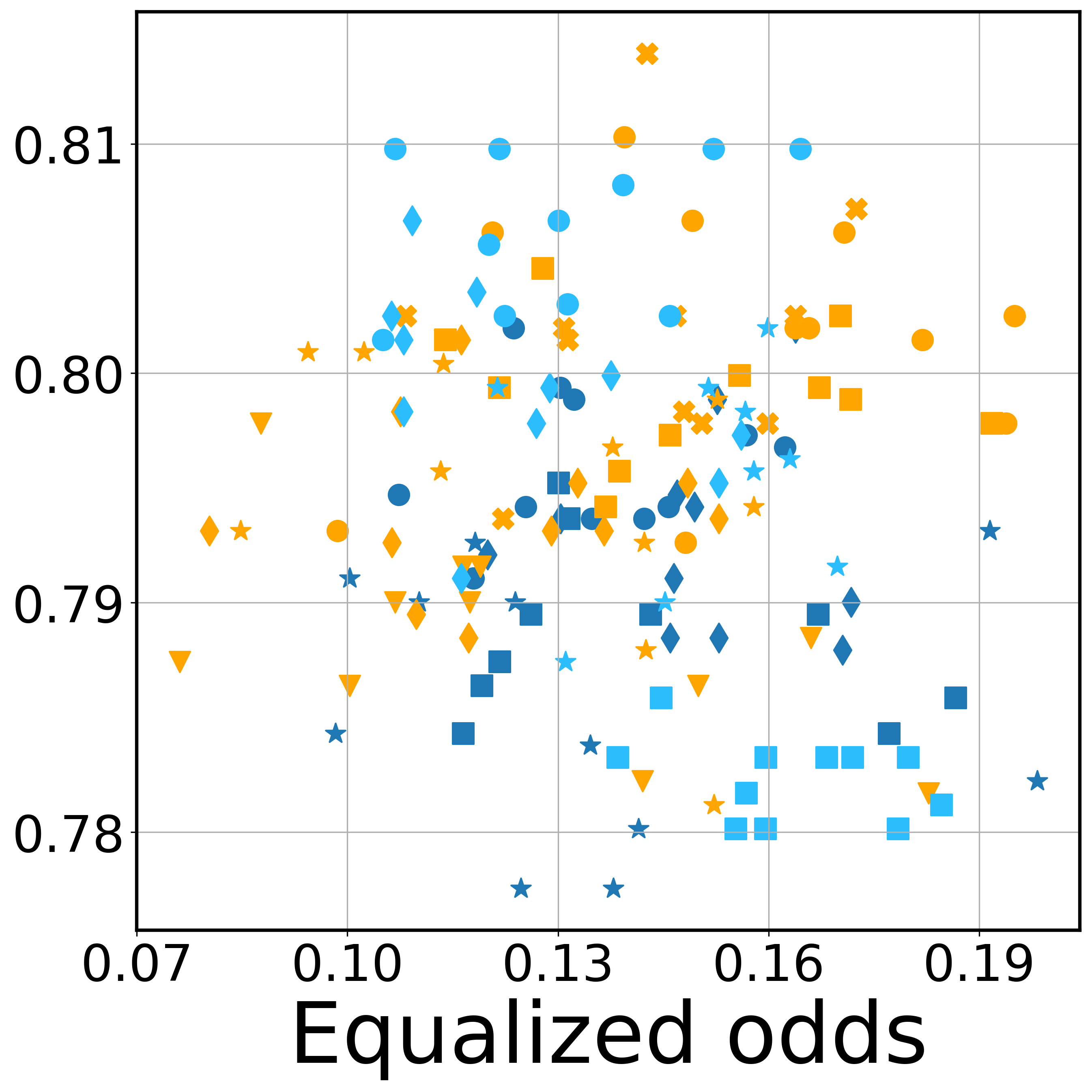}&
 \hspace{-.4cm}\includegraphics[scale=0.175]{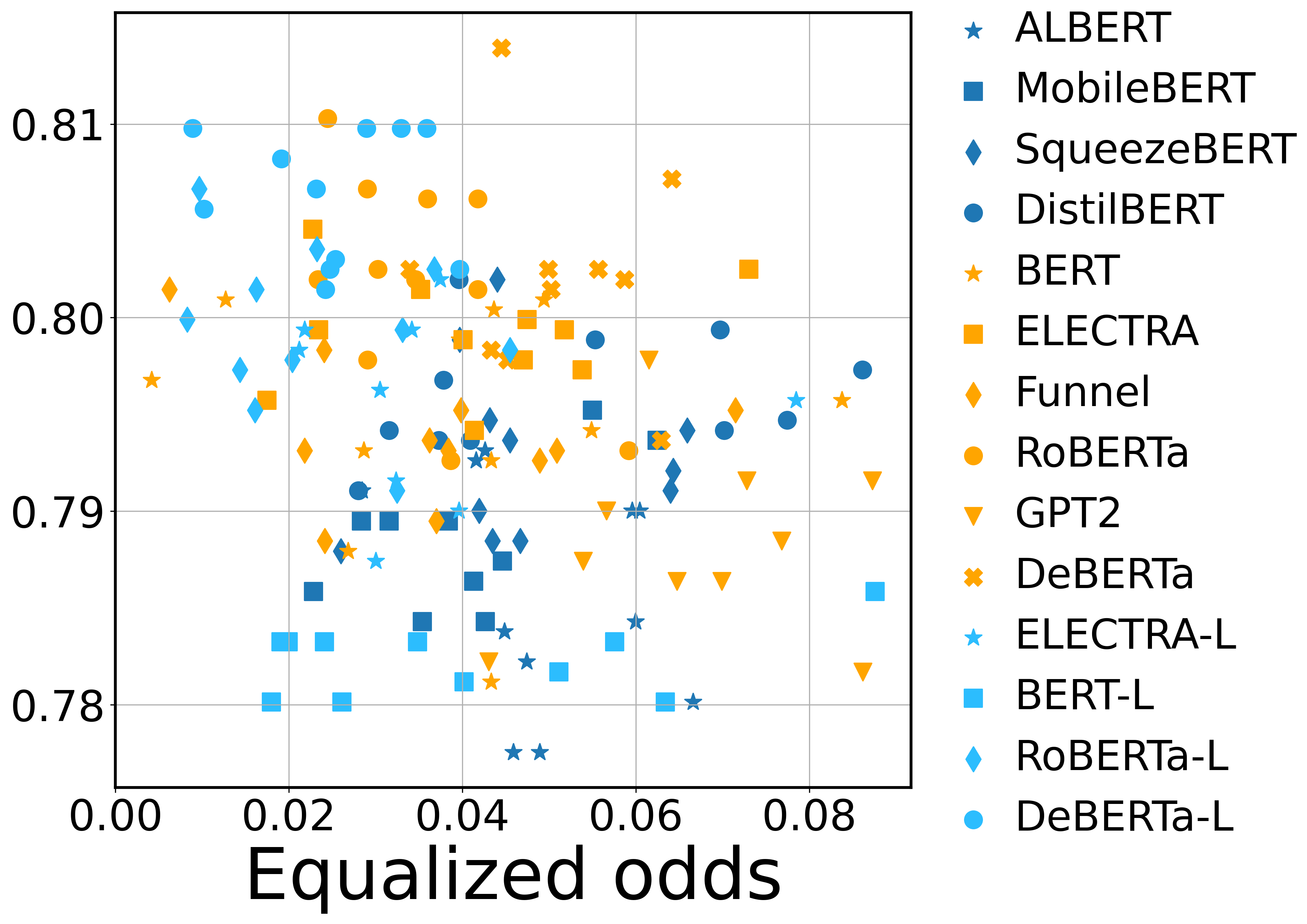}
 \\
 religion & race & gender\\
\end{tabular}
\end{center}
\caption{Accuracy versus equalized odds for fine-tuned LMs when varying the random seed used in fine-tuning.}
\label{fig:rand_seeds_accuracy}
\end{figure*}

In this section we present graphs similar to the ones in Section~\ref{section:seeds} using accuracy as a measure of performance instead of balanced accuracy. These plots makes it obvious how close in performance all models are and emphasize the gap in fairness measure observed across different random seeds for each fine-tuned model. The results are shown in Figure~\ref{fig:rand_seeds_accuracy}. Note that all Jigsaw models get an accuracy in performance of approximately 95\% with a gap of approximately .05 for equalized odds. HateXplain models exhibit a higher variance in accuracy (4-5\%) across all models with an even larger gap of 0.15 for equalized odds for most models. Note that each LM has a modest variation in accuracy that spans approximately 1\%.

For HateXplain, we also experimented with BERTweet~\citep{Nguyen2020BERTweet}, a BERT-base sized model following the RoBERTa pretraining procedure that is further trained on Twitter data, using the checkpoint available in the Hugging Face model hub. In our experiments, BERTweet presented the largest variation for accuracy (results not shown), achieving both the best and the worst accuracy across all models (across the 11 random seeds we used), spanning a spread of 4.5\%. The EO measure for BERTweet exhibited a variation of 0.12 for religion. We acknowledge that a more thorough analysis is required to better understand the effects of in-domain pretraining (in this case on tweets) for both accuracy and fairness. For example, recent work showed that model behavior can be adjusted to a set of ``target values'' if the model is trained on a small, well-behaved dataset~\cite{Solaiman2021Process}.

\subsection{Fair Score Transformer (FST)}
\label{appendix:fst}

In this section, we expand on our discussion of the application of FST in this work.

The generalized equalized odds (GEO) criterion targeted by FST is computed as the maximum of the between-group absolute differences in average scores for positively-labeled and negatively-labeled instances \citep{Wei2020Optimized}. It is analogous to EO where instead of the predicted label, the corresponding probability for the label is used instead. 

Regarding issue 1) mentioned in Section~\ref{sec:method:post} (calibration of input scores), we found that the distributions of softmax outputs of the tested LMs are bimodal and highly concentrated near values of $0$ and $1$ (as commonly observed with deep neural networks). Such skewed distributions appear to violate FST's expectation of probabilities as input and are typically not encountered on tabular datasets on which FST was previously tested. Thus we experimented with calibrating the LM outputs. We considered both logistic regression of the class label on the logit outputs of the LMs (a generalization of temperature scaling~\citep{guo2017calibration}), as well as linear regression on the logit outputs followed by clipping of the resulting values to the interval $[0, 1]$. In general, logistic regression proved somewhat beneficial for the Jigsaw dataset and we included it in our results.

\begin{figure*}[htb]
\begin{center}
\begin{tabular}{ccc}
 \includegraphics[scale=0.16]{img/jigsaw_bert_fst_nocal_disco_rain_religion.png}&
 \hspace{-.4cm}\includegraphics[scale=0.16]{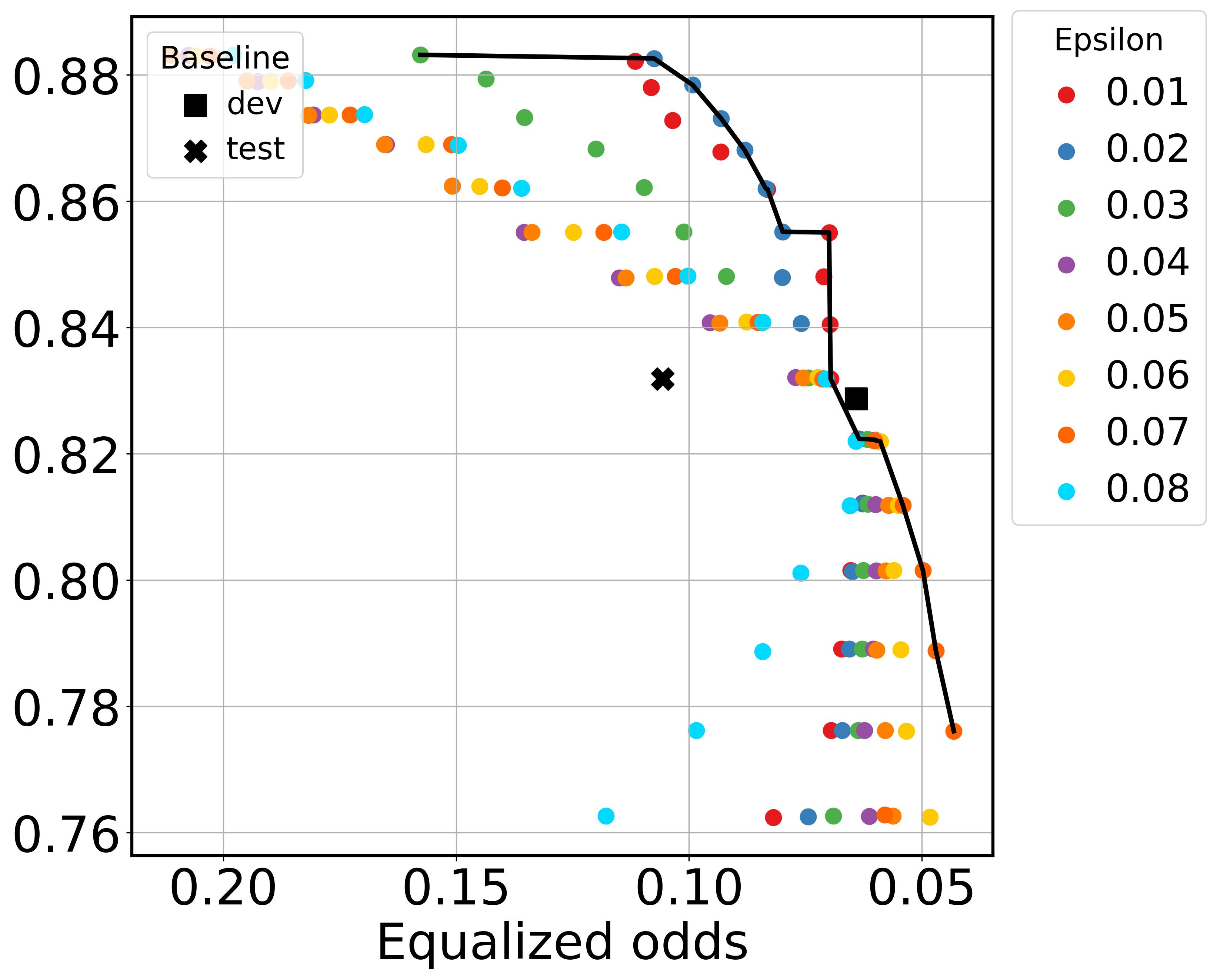}&
 \hspace{-.4cm}\includegraphics[scale=0.16]{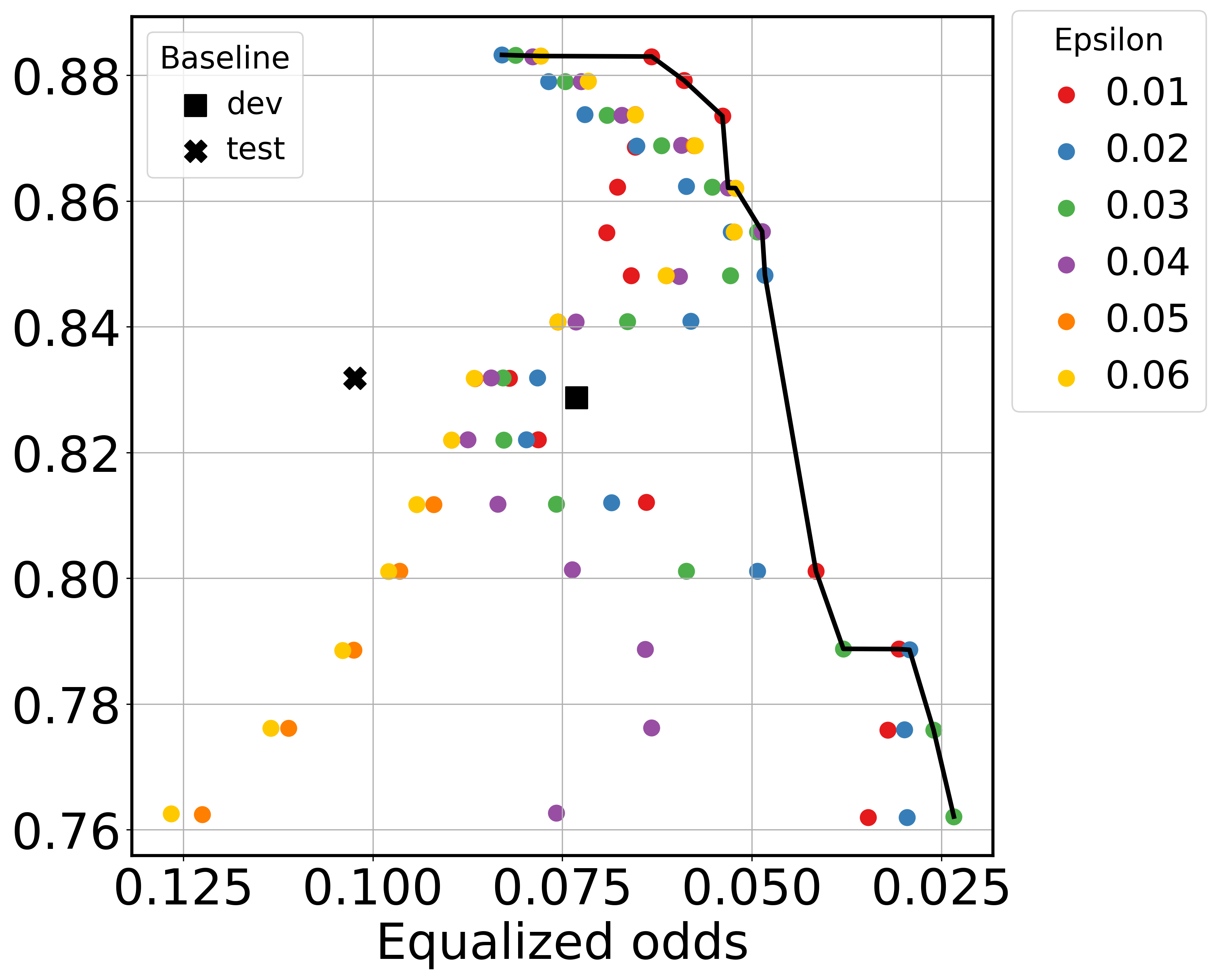}
 \\
 religion & race & gender\\
\end{tabular}
\end{center}
\caption{FST tuning for BERT: Balanced accuracy versus equalized odds on the Jigsaw dataset when varying fairness parameter $\epsilon$ and binary classification threshold $t$ after applying the FST method for group bias mitigation.}
\label{fig:jigsaw_disco}
\end{figure*}

\begin{figure*}[tb]
\begin{center}
\begin{tabular}{ccc}
 \includegraphics[scale=0.29]{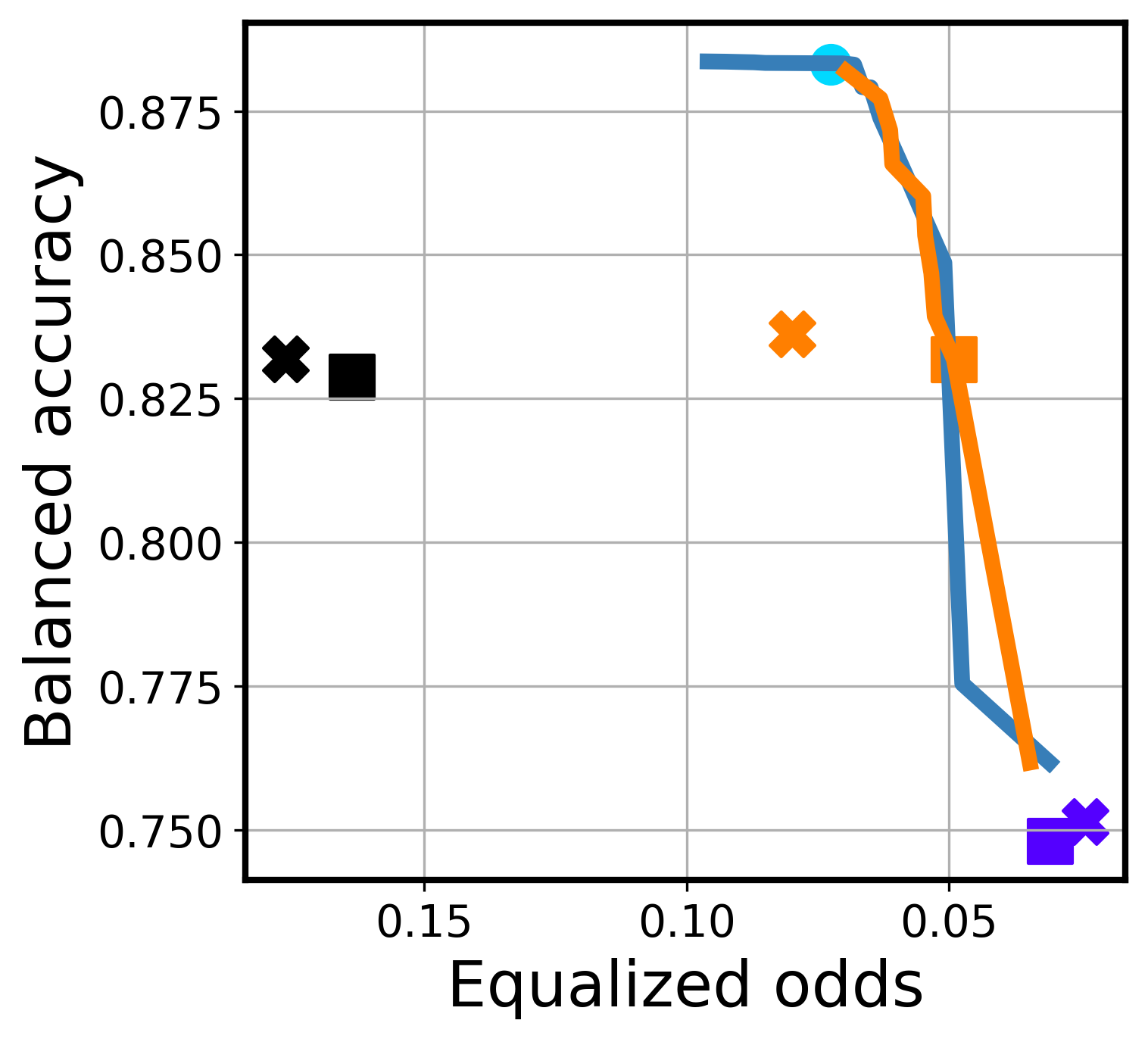}&
 \hspace{-.4cm}\includegraphics[scale=0.29]{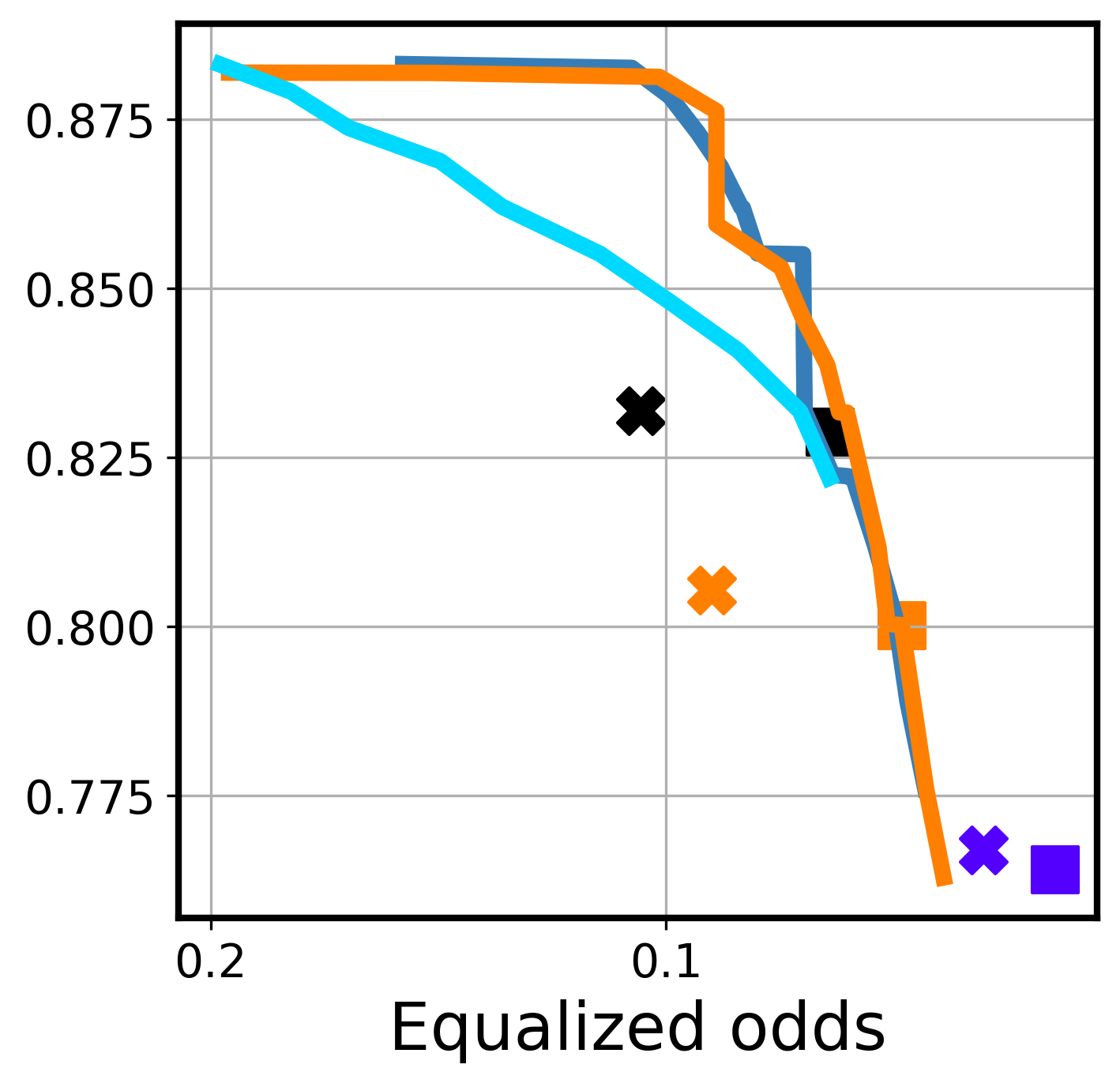}&
 \hspace{-.4cm}\includegraphics[scale=0.29]{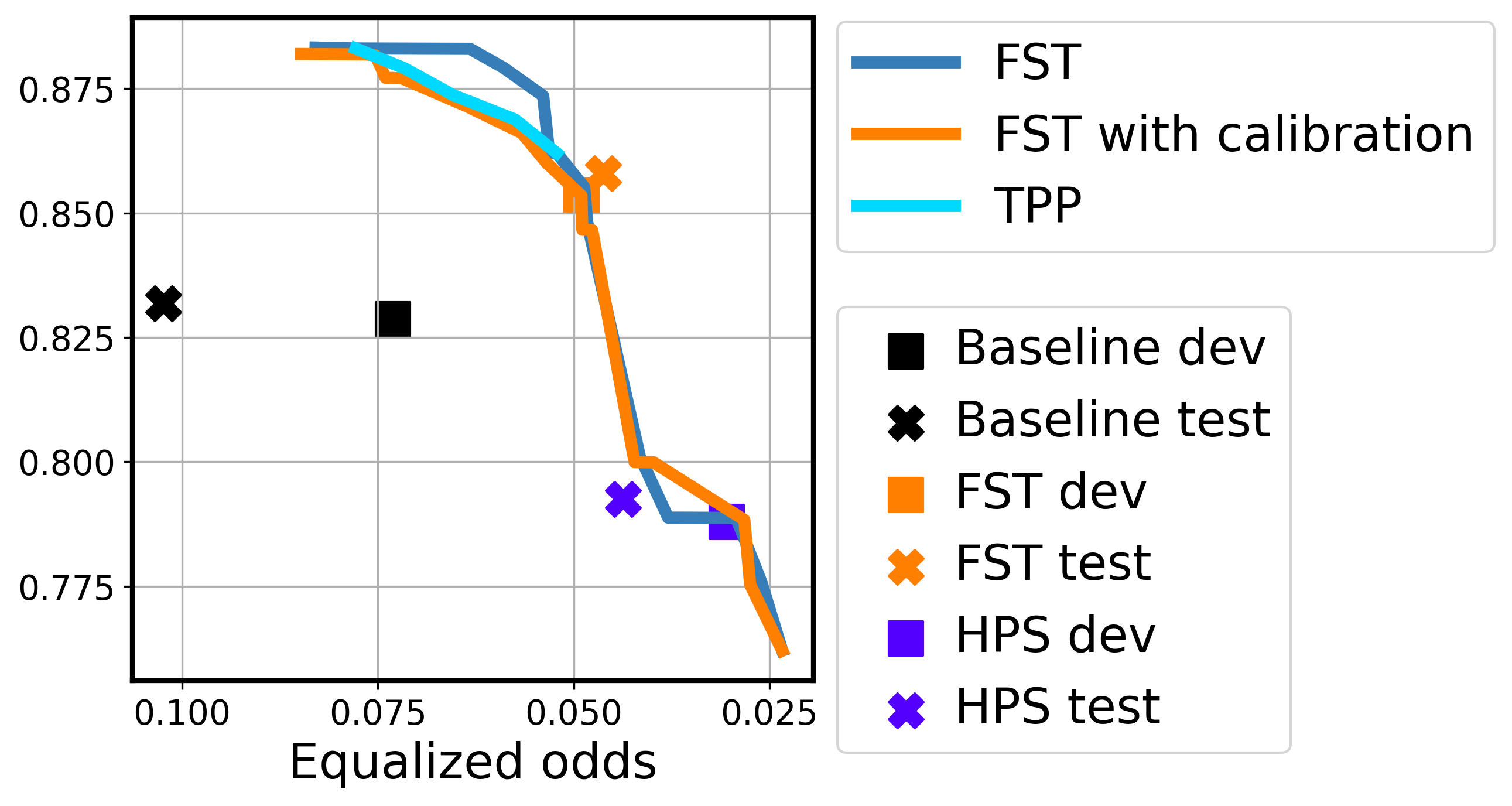}
 \\
 religion & race & gender\\
\end{tabular}
\end{center}
\caption{BERT: Balanced accuracy versus equalized odds on the Jigsaw dataset when applying the FST and HPS methods for group bias mitigation and threshold post-processing (TPP) alone.}
\label{fig:jigsaw_bias_mitigation}
\end{figure*}

Regarding issue 2) (choice of fairness parameter), we found, as noted by \citet{Wei2020Optimized}, that while the parameter $\epsilon$ controls the deviation from GEO (i.e.~the ``GEO difference''), this is not always correlated with the EO difference, which is a function of the output after thresholding. Regarding 3) (classification threshold), we found that varying the threshold $t$ can significantly affect equalized odds as well as accuracy and balanced accuracy, and can sometimes even produce a reasonable trade-off between them. For this reason, we included a version of post-processing (see ``Threshold post-processing (TPP)'' in Section~\ref{sec:method:post}. This effect of the prediction threshold on fairness has not been explored in previous work to our knowledge. 

As a result of our observations regarding 2) and 3), we used the following procedure to select a \emph{set} of $(\epsilon, t)$ pairs to map out a trade-off between fairness and accuracy. The training set used to fine-tune the LMs is never seen by FST. The development dataset (``dev'') is used to both tune the FST parameters and evaluate the resulting transformation. As such, the dev dataset was further split into a dev-train set and a dev-eval set. Given an $\epsilon$ value, FST was fit on the dev-train set to ensure a GEO difference of at most $\epsilon$. Then on the dev-eval set, given $\epsilon$ and $t$, scores were transformed by FST with parameter $\epsilon$, thresholded at level $t$ to produce a binary label, and finally evaluated for both fairness and accuracy. Each $(\epsilon, t)$ pair thus yields one point in the equalized odds-accuracy plane, as seen in Figure~\ref{fig:jigsaw_disco}. We selected $(\epsilon, t)$ pairs that are Pareto-efficient on the dev-eval set, to ensure the best fairness-accuracy trade-off.

This is the first time FST is used with unstructured, text data and with large datasets in the order of millions of samples. First, we implemented FST following the proposed implementation in~\citet{Wei2020Optimized}. This first implementation ended up with numerical instabilities that lead to either slow running times (in the order of hours) or even situations when the method did not converge. We managed to improve upon the computational cost of FST, which was instrumental in scaling to the large Jigsaw dataset and allowing rapid experimentation. Specifically, in the dual ADMM algorithm of~\citet{Wei2020Optimized}, the first step (eq.~(14) therein) consists of $n$ parallel optimizations, each involving a single variable. We observed that these optimizations can be done in closed form by solving a cubic equation. We refer to \citet[Appendix~B.1]{Wei2021Optimized} for details of the closed-form solution as it is not the focus of the present paper. The replacement of an iterative optimization with a closed-form solution greatly reduces the computational cost of FST. The improved FST runs in the order of 1-2 minutes for the Jigsaw dataset and in seconds for HateXplain. Equally important, it also eliminates instances of the iterative optimization failing to converge.

\subsection{Bias mitigation through post-processing methods}
\label{appendix:bias_mitigation}

In this section we present additional results on applying post-processing methods for group bias mitigation. We first discuss the results of parameter tuning for Fair Score Transformer (FST)~\citep{Wei2020Optimized}. More details about FST itself can be found in the Appendix~\ref{appendix:fst}. The FST method has one parameter, $\epsilon$, that can be fine-tuned. 
Using the transformed scores from the FST, we also investigate tuning the threshold used in the binary classifier, instead of using the default value of 0.5, as explained in Section~\ref{sec:method:post}. Figure~\ref{fig:jigsaw_disco} depicts the data points obtained by varying epsilon and for each epsilon value, varying the classification threshold.~\footnote{All points are shown for the dev set as this plot corresponds to tuning FST parameters.} When choosing an operating point, the points on the black Pareto frontier are the most interesting points: highest balanced accuracy and lowest equalized odds. For reference, we also show the baseline points without bias mitigation for the dev and test sets. All data points are plotted for fine-tuned BERT. Similar trends are observed for the rest of the models considered in this study and for the HateXplain dataset.

\begin{figure*}[tb]
\begin{center}
\begin{tabular}{ccc}
\includegraphics[scale=0.29]{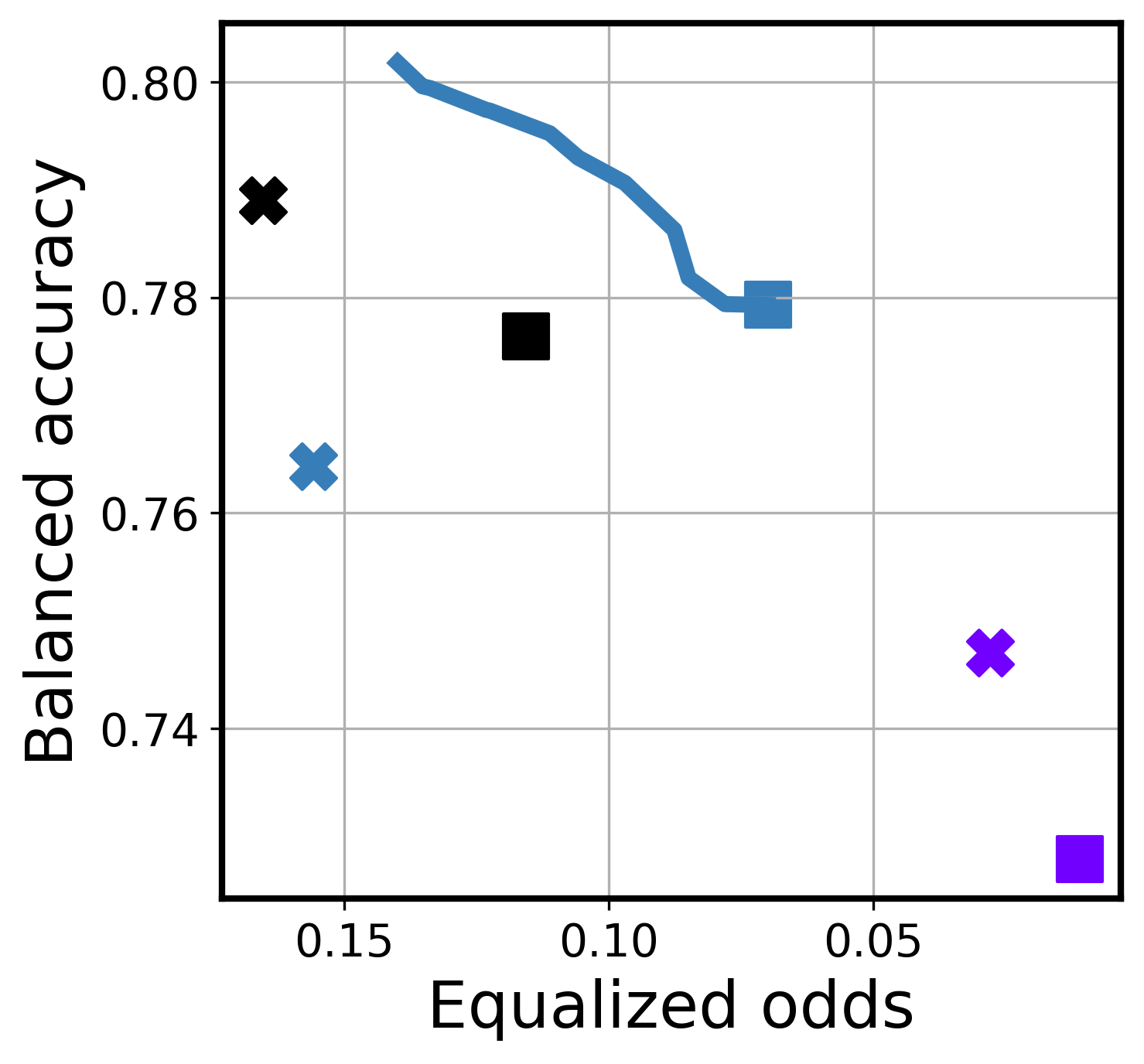} &
\includegraphics[scale=0.29]{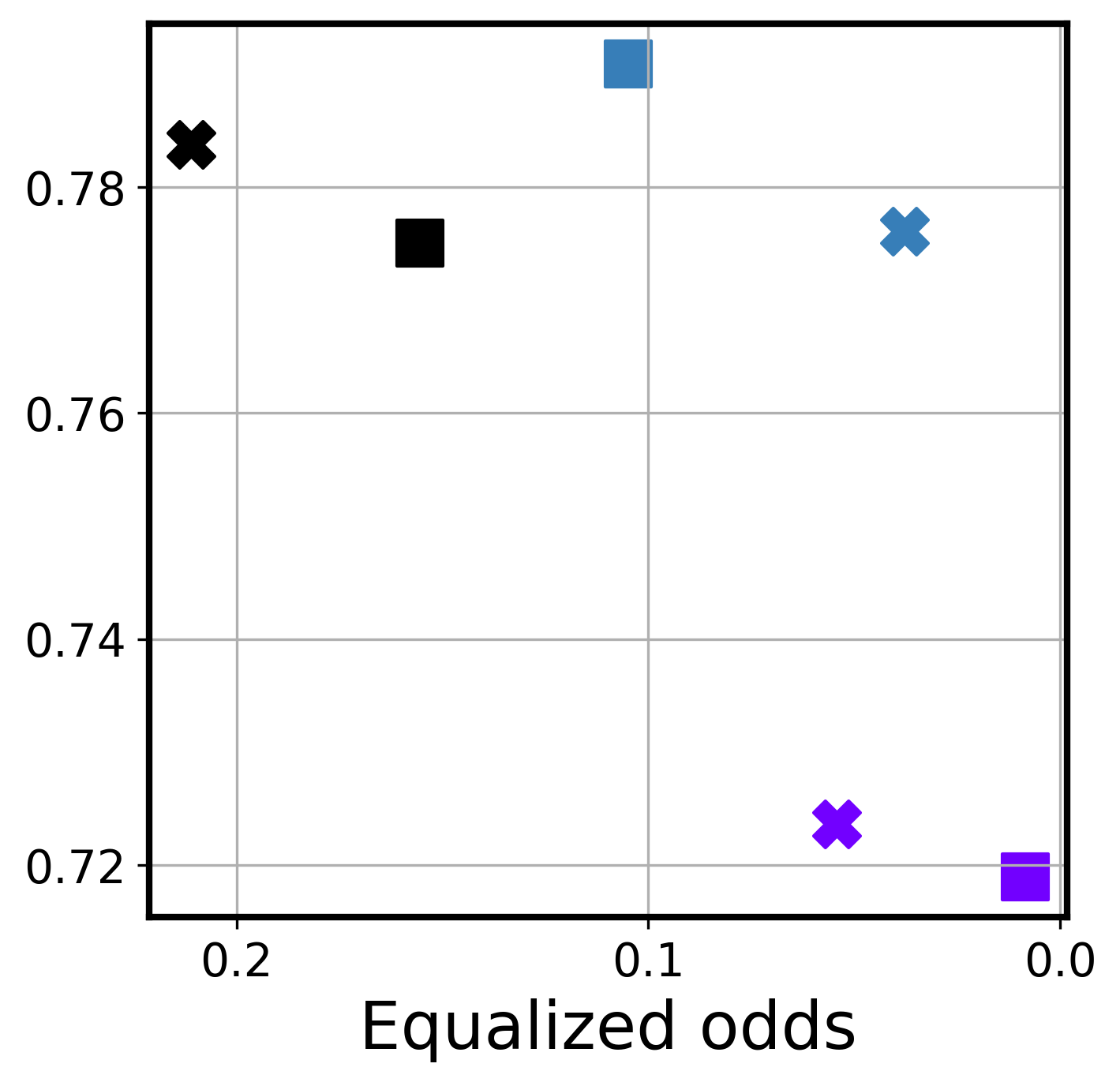} &
\includegraphics[scale=0.29]{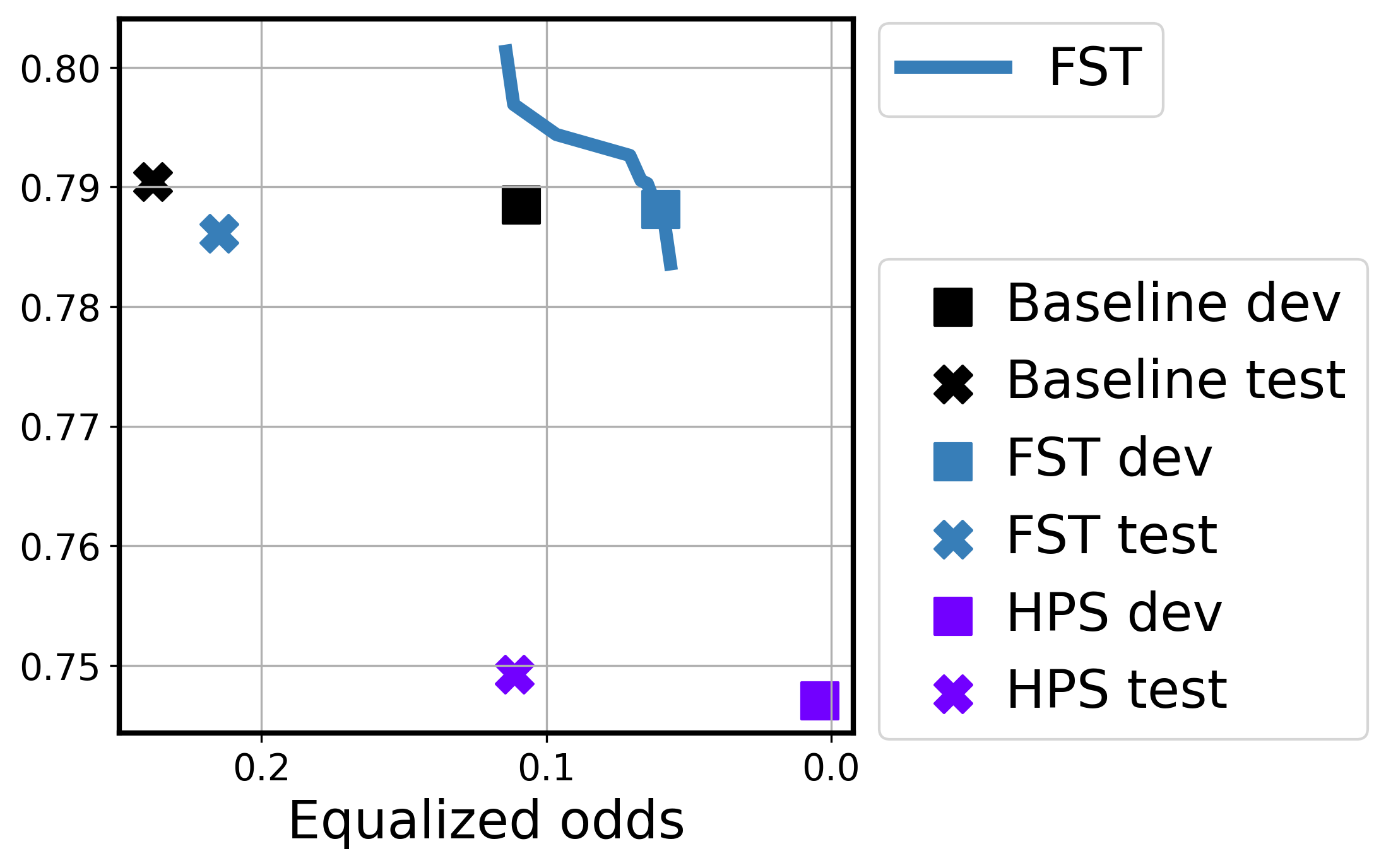}
\\
DistilBERT & BERT & DEBERTA large
\\
\end{tabular}
\end{center}
\caption{Balanced accuracy versus equalized odds for fine-tuned LMs (religion) on the HateXplain dataset when applying the FST and HPS methods for group bias mitigation and threshold post-processing (TPP) alone.}
\label{fig:hatex_bias_mitigation}
\end{figure*}

We also experimented with calibrating the scores using logistic regression before post-processing. In Figure~\ref{fig:jigsaw_bias_mitigation}, we plot the Pareto frontiers of bias mitigation when applying FST, with and without calibration, along with the threshold post-processing (TPP) method. We also show the result of HPS, which yields a single operating point, as well as the baselines without bias mitigation. In general, on the Jigsaw dataset, FST is successful in reducing EO with different degrees of success depending on the model/group. It thus offers an interesting set of points with different accuracy-EO trade-offs. For reference, we show the equivalent point for the test set (orange $x$) for the operating point in dev that achieves an equalized odds of at most 0.05 (orange square). In certain cases, FST manages to lower the equalized odds with minimal or no decrease in accuracy, as seen in the religion and gender columns in Figure~\ref{fig:jigsaw_bias_mitigation}. %Remarkably, for the Jigsaw dataset, FST manages to achieve considerably improved fairness with minimal or no degradation in accuracy.
Note that all points in the plots except for the $x$ points are plotted using the dev dataset split, the $x$ points are test points corresponding to dev points that obtain an EO of at most 0.05.

In comparison, HPS seems particularly effective in lowering the equalized odds and thus improving the fairness of the model, with some penalty on the accuracy. For Jigsaw, applying only TPP (i.e., tuning the threshold used in the binary classification) also offers some interesting operating points. TPP has a small search space compared to FST and sometimes the Pareto frontier is reduced to one point, as is the case for the religion group. In general, FST has superior Pareto frontiers compared to TPP alone. In addition, as we will discuss shortly, TPP proved inefficient for the HateXplain dataset. Last, using score calibration before feeding the scores to FST does not seem to offer significant improvements. Similar trends can be observed for the rest of the models.

In Figure~\ref{fig:hatex_bias_mitigation}, we show the results of applying bias mitigation techniques for a few LMs, one for each size category, on the HateXplain dataset with religion as the sensitive group. Unlike Jigsaw, the results of the bias mitigation techniques follow different trends. HPS still manages to substantially reduce the EO for all models, but with a considerable decrease in balanced accuracy (in some cases, more than six percentage points). For FST, the fine-tuning for epsilon and classification threshold does not lead to a large search space as observed in the Jigsaw case. Moreover, the reduction in EO is more limited and sometimes the improvement observed for the dev set disappears in test. There are cases, though, such as BERT, where FST successfully reduces EO and the reduction is maintained or even improved in test. Across the board, tuning only the threshold used in classification (TPP) did not lead to improved results and we omit showing them in the plots.

Overall, we find the post-processing methods for bias mitigation worth considering. They are straightforward to apply, run in the order of seconds or minutes on the CPU of a laptop and they offer interesting operating points when other methods for bias elimination would incur a significant computational cost, such as pre-processing or in-processing techniques. Obtaining the Pareto frontiers is instantaneous as the search space for FST is not that large.

\balance
\subsection{Other post-processing methods for bias mitigation}
In addition to the two post-processing methods that we considered in our study, other post-processing methods for bias mitigation include assigning favorable labels to unprivileged groups in regions of high classifier uncertainty \citep{kamiran2012decision}, minimizing error disparity while maintaining classifier calibration \citep{pleiss2017fairness}, a relaxed nearly-optimal procedure for optimizing equalized odds \citep{woodworth2017learning}, shifting the decision boundary for the protected
group \citep{fish2016confidence}, iterative post-processing to achieve unbiased predictions on every identifiable subpopulation \citep{kim2019multiaccuracy}, recalibrating a classifier using a group-dependent threshold to optimize equality of opportunity (defined as the difference between the group-wise true positive rates)~\citep{chzhen2019leveraging}, using optimal transport to ensure similarity in group-wise predicted score distributions \citep{jiang2020wasserstein}, and a plug-in approach for transforming the predicted probabilities to satisfy fairness constraints \citep{yang2020fairness}.

\subsection{Reproducibility statement}
The data processing we performed for the datasets we used is briefly explained in Appendix~\ref{appendix:datasets}. In all our experiments we used unmodified versions of the model implementations from the Hugging Face transformers library (version 4.3.3) and the main scripts to tune the models are modified versions of the sequence text classification examples accompanying the library. The hyper-parameter tuning we performed was minimal (varying the number of epochs from 1-3, two values for learning rates $2e-6$ and $2e-5$, 11 values for random seeds). More details on the experimental infrastructure can be found in Section~\ref{section:lms}. The main limiting factor in reproducing the results presented in this study is having access to GPUs such as the NVIDIA V100 and A100 and generous, parallel compute time. 
%We could not find an open-source implementation for FST. 
At the time of this writing, the implementation of FST that we used is evolving proprietary code that may become available for external consumption. More details are provided in Appendix~\ref{appendix:fst}. For HPS, we used the open-source implementation that can be found as part of the \href{https://github.com/Trusted-AI/AIF360/blob/master/aif360/algorithms/postprocessing/eq_odds_postprocessing.py}{AIF360 toolkit, ``equalized odds post-processing'' method}.

% \clearpage
% \newpage
% \begin{minipage}{\textwidth}

% \subsection{Mathematical definitions for the measures used}

% \paragraph{Accuracy}

% \[
% accuracy = \frac{\textit{number of samples predicted correctly}}{\textit{total number of samples}}
% \]

% \paragraph{Balanced accuracy}
% \[
% \textit{balanced accuracy} = \frac{\frac{\textit{true negatives}}{\textit{total number of negative samples}} + \frac{\textit{true positives}}{\textit{total number of positive samples}}}{2}
% \]

% \paragraph{Equalized odds}
% \[
% \textit{TPR} = \frac{\textit{number of samples predicted correctly as positive}}{\textit{total number of positive samples}}
% \]

% \[
% \textit{FPR} = \frac{\textit{number of samples predicted incorrectly as positive}}{\textit{total number of negative samples}}
% \]

% \[
% \textit{EO} = \max(abs(\textit{TPR}_{\textit{protected group}} - \textit{TPR}_{\textit{unprotected group})},
% abs(\textit{FPR}_{\textit{protected group}} - \textit{FPR}_{\textit{unprotected group}}))
% \]

% \paragraph{Generalized equalized odds}
% \[
% \textit{generalized TPR} = \textit{genTPR} = \frac{\textit{sum of predicted probabilities for positive samples}}{\textit{total number of positive samples}}
% \]

% \[
% \textit{generalized TNR} = \textit{genTNR} =  \frac{\textit{sum of predicted probabilities for negative samples}}{\textit{total number of negative samples}}
% \]

% \[
% \textit{generalized EO} = \max(abs(\textit{genTPR}_{\textit{protected}} - \textit{genTPR}_{\textit{unprotected}}),abs(\textit{genTNR}_{\textit{protected}} - \textit{genTNR}_{\textit{unprotected})})
% \]

% \end{minipage}

\end{document}